%% file: Main_Paper/arxiv_2026_new.tex
\documentclass[11pt]{article}

\def\paperformat{arxiv}

\usepackage{etoolbox}
\providecommand{\paperformat}{aistats}% default if not set elsewhere

\ifdefstring{\paperformat}{neurips}{%
  \typeout{Using NeurIPS 2026 style}%
  \usepackage[preprint]{neurips_2026}%
}{%
\ifdefstring{\paperformat}{iclr}{%
  \typeout{Using ICLR 2027 style}%
  \usepackage{iclr2027/iclr2027_conference}%
  \usepackage{times}%
}{%
\ifdefstring{\paperformat}{aistats}{%
  \typeout{Using AISTATS 2027 style (submission/under review)}%
  \usepackage{AISTATS2027PaperPack/aistats2027}%
}{%
  \typeout{Using plain arXiv/article style}%
  \usepackage[margin=0.8in]{geometry}%
  \frenchspacing
}}}

\usepackage{natbib}
\usepackage{enumerate}
\usepackage{graphicx} % Required for inserting images
\usepackage{svg}
\usepackage{tikz}
\usepackage{stmaryrd}
\definecolor{grpR}{HTML}{EAF1FB}\definecolor{grpRd}{HTML}{CFDFF5} % Reasoning
\definecolor{grpK}{HTML}{EAF6EC}\definecolor{grpKd}{HTML}{CDE9D2} % Knowledge
\definecolor{grpM}{HTML}{FDF1E3}\definecolor{grpMd}{HTML}{F9DDBB} % Math
\definecolor{grpI}{HTML}{F3EDF8}\definecolor{grpId}{HTML}{E1D3EE} % Instruction following

\usetikzlibrary{patterns,matrix,positioning}
\usepackage{wrapfig} % allows text to wrap around figures/tables
\usepackage{subcaption}
\usepackage{tocloft} % compact, configurable ToC (used in supplement)
\usepackage{mathrsfs}
\usepackage{amsmath}
\usepackage{amssymb}
\usepackage{bm} % provides \bm for bold math symbols
\usepackage{colortbl} % in preamble
\usepackage[most]{tcolorbox}

\usepackage{algorithm}
\usepackage{algorithmicx}
\usepackage{algpseudocode}
\usepackage{float} % provides the [H] specifier (non-floating) so algorithms can live inside minipage/wrapfigure

\usepackage[table,dvipsnames]{xcolor}
\usepackage[normalem]{ulem}
\usepackage{fontawesome5}

\newcommand{\todonotesopts}{textsize=footnotesize}
\ifdefstring{\paperformat}{arxiv}{\renewcommand{\todonotesopts}{disable}}{}
\usepackage[\todonotesopts]{todonotes}

\usepackage{pgfplots}
\usepackage{pgfplotstable}
\pgfplotsset{compat=1.17}

\usepackage{adjustbox}
\usepackage{longtable}
\usepackage{caption}
\usepackage[toc, page]{appendix}

\usepackage[utf8]{inputenc} % allow utf-8 input
\usepackage[T1]{fontenc} % use 8-bit T1 fonts

\usepackage{hyperref}

\definecolor{mahogany}{HTML}{C04000}
\definecolor{forestgreenweb}{HTML}{228B22}
\hypersetup{%
  colorlinks = true,
  citecolor  = mahogany,
  linkcolor  = forestgreenweb,
  urlcolor   = forestgreenweb,
}

\usepackage[nameinlink,noabbrev]{cleveref}

\usepackage{url} % simple URL typesetting
\usepackage{booktabs} % professional-quality tables
\usepackage{amsfonts} % blackboard math symbols
\usepackage{nicefrac} % compact symbols for 1/2, etc.
\usepackage{microtype} % microtypography

\ifdefstring{\paperformat}{arxiv}{%
  \usepackage[numbered]{bookmark}

  \usepackage{placeins} % provides \FloatBarrier
  \floatplacement{figure}{!t}
  \floatplacement{table}{!t}

  \usepackage{titlesec}
  \titlespacing*{\section}{0pt}{1.0ex plus 0.3ex minus 0.2ex}{0.6ex}
  \titlespacing*{\subsection}{0pt}{0.9ex plus 0.2ex minus 0.2ex}{0.5ex}
  \titlespacing*{\subsubsection}{0pt}{0.7ex plus 0.2ex minus 0.2ex}{0.4ex}
  \titlespacing*{\paragraph}{0pt}{0.25ex plus 0.1ex minus 0.1ex}{0.6em}
  \titlespacing*{\subparagraph}{0pt}{0.25ex plus 0.1ex minus 0.1ex}{0.6em}

  \setlength{\abovedisplayskip}{2pt plus 1pt minus 1pt}
  \setlength{\belowdisplayskip}{2pt plus 1pt minus 1pt}
  \setlength{\abovedisplayshortskip}{1pt plus 1pt minus 1pt}
  \setlength{\belowdisplayshortskip}{1pt plus 1pt minus 1pt}
  \setlength{\jot}{1pt}
}{}

\newcommand{\one}{\mathbf 1}
\newcommand{\bnu}{\boldsymbol{\nu}}

\usepackage{multirow}

\usepackage{amsthm}

\newtheorem{lemma}{Lemma}
\newtheorem{corollary}{Corollary}

\newenvironment{proofsketch}{%
  \begin{proof}[\textbf{\textcolor{black!70}{Proof sketch}}]%
  \color{black!70}%
}{\color{black}\end{proof}}

\makeatletter
\AtBeginDocument{%
  \@ifundefined{remark}{}{%
    \let\remarkorig\remark
    \let\endremarkorig\endremark
    \renewenvironment{remark}{\color{black!70}\remarkorig}{\endremarkorig\color{black}}%
  }%
}
\makeatother

\input{macros}

\input{settings}

\let\remorig\rem
\let\endremorig\endrem
\renewenvironment{rem}{\color{black!70}\remorig}{\endremorig\color{black}}

\ifdefstring{\paperformat}{arxiv}{%
  \let\oldbibliographystyle\bibliographystyle
  \renewcommand{\bibliographystyle}[1]{\oldbibliographystyle{alpha}}%
}{}

\begin{document}

% -------- Title block (format-dependent) --------
\ifdefstring{\paperformat}{aistats}{%
  \twocolumn[
  \aistatstitle{\papertitle}

  % In submission/under-review mode, the style file will automatically typeset
  % an anonymous author block (regardless of the contents here).
  \aistatsauthor{%
    Prateek Chanda$^{1}$ \And Saral Sureka$^{1}$ \And Krishnateja Killamsetty$^{2}$ \And Ganesh Ramakrishnan$^{1}$
  }

  \aistatsaddress{%
    $^{1}$ Indian Institute of Technology Bombay (cse.iitb.ac.in) \And
    $^{2}$ Amazon (amazon.com)
  }
  ]
}{%
  % Plain article/arXiv title block (like the body-only file's tabular style).
  \title{\papertitle}
  \author{%
  \begin{tabular}{c}
    Prateek Chanda\textsuperscript{1} \quad
    Saral Sureka\textsuperscript{1} \quad
    Krishnateja Killamsetty\textsuperscript{2} \quad
    Ganesh Ramakrishnan\textsuperscript{1}\\[0.5ex]
    \textcolor{magenta}{\textsuperscript{1}Indian Institute of Technology Bombay (cse.iitb.ac.in)}\\
    \textcolor{magenta}{\textsuperscript{2}Amazon (amazon.com)}
  \end{tabular}
  }
  \date{\today}
  \maketitle
}
% ----------------------------------------------

\begin{abstract}
Supervised fine-tuning over a large pool of heterogeneous tasks is often computationally expensive, leading practitioners to train language models on a fixed training budget of carefully chosen SFT tasks. Thus, achieving strong downstream performance hinges on selecting an effective subset of tasks, or equivalently, identifying an optimal task mixture under a fixed budget constraint. In practice, supervised finetuning mixtures are often determined using simple heuristics (e.g., uniform or size-proportional sampling) that ignore task interactions, which can hurt transfer and waste budget on redundant task sources. We introduce \methodprop, a framework for learning continuous task mixtures through energy-based modelling over SFT tasks. Each task forms a node in a Markov random field, with unary potentials capturing task utility and pairwise potentials modelling inter-task relationships via behavioral divergences between predictive distributions of single-task fine-tuned models. Optimizing this objective yields task mixtures that balance utility, coverage, and redundancy. We further show that the induced set function defined over the set of such tasks is weakly submodular under budget constraints, enabling approximation guarantees for discrete task selection. Empirically, we provide a fixed, benchmark-independent mixture that improves over heuristic mixing across multiple model families and different training budgets. \methodprop{} also uncovers interpretable structure in task interactions, revealing complementarities and redundancies that inform effective SFT mixtures.
\end{abstract}

\addtocontents{toc}{\protect
  \setcounter{tocdepth}{-1}}
  \section{Introduction}

   Supervised fine-tuning (SFT) on a mixture of instruction datasets shapes much of a language model's reasoning and problem-solving ability. With a fixed training budget and pools of hundreds of tasks, the central design choice is \emph{which tasks to include and how much of the budget each receives}. Common practice relies on uniform sampling, dataset-size weighting \citep{chung2022scaling}, or hand-crafted rules, which can let dominant sources take over the mixture and spend budget on redundant tasks. Automated methods such as submodular selection (SMART; \citealp{renduchintala-etal-2024-smart}), influence-based retrieval (LESS, \citealp{xia2024less}; BIDS, \citealp{dai2025improving}) and proxy-based mixture prediction (RegMix, \citealp{liu2024regmix}; Data Mixing Laws, \citealp{li2025data}) typically select discrete subsets rather than continuous proportions, rely on semantic or per-example scores that do not capture cross-task transfer, or require repeated training runs.

  \begin{center}
\textcolor{blue!70}{\textit{How can we automatically choose an SFT task mixture that is simultaneously representative and diverse, and that reliably improves downstream performance under a fixed budget?}}
\end{center}

  We introduce \textbf{\methodprop{}} (\textbf{Task Mixtures via Probabilistic Graphical Modelling}), which models the task pool as a dense Markov random field: \emph{unary potentials} reward representative tasks and \emph{pairwise potentials} penalize redundant ones. Both are computed from \emph{behavioral} similarity, by comparing the predictions of single-task fine-tuned models, rather than from metadata or embeddings. The learned weights also show which tasks act as broad ``attractors'' and which are specialized, and for budgeted selection we show that the induced set function is weakly submodular, so greedy construction comes with approximation guarantees.

  % Research-question helper (ensures \rq{Q?}{...} renders in light gray)
  \renewcommand{\rq}[2]{%
    \par\smallskip
    \noindent\hangindent=1.6em\hangafter=1%
    \textbf{#1.}\hspace{0.6em}{\color{black!65}\itshape #2}\par
  }

  \rq{Q1}{Can functional, behavior-based task affinities be used to compute continuous mixture proportions that
  outperform strong heuristic and selection baselines?}

  \rq{Q2}{Does the resulting mixture optimization yield interpretable structure (task clusters, redundancy, and
  influential ``attractor'' tasks) and support efficient budgeted construction?}

\textbf{Contributions.} Our main contributions are as follows:
\begin{itemize}
  \item \textbf{An energy model over behavioral task affinities.} We cast SFT mixture design as energy minimization over tasks, with affinities measured from how single-task models predict each other's data.

  \item \textbf{Guarantees for budgeted task selection.} We show that the induced set function is monotone and weakly submodular, so greedy task selection has an approximation guarantee.

  \item \textbf{Insight into what drives the mixture.} We characterize how much spectral shift convexity actually requires, and show that the mixture favors broadly transferable tasks.

  \item \textbf{A single, benchmark-independent mixture.} Computed once, the same mixture outperforms heuristic and selection baselines across three model families across different budgets on state-of-the-art benchmarks.
\end{itemize}

\section{Preliminaries and Background work}
\textbf{Notation.}
We write $[n]:=\{1,\ldots,n\}$ and $\mathbb R^n_+$ for the non-negative
orthant. Vectors are lower-case bold ($\vx$) and matrices upper-case bold
($\mX$), with individual entries denoted as $\vx_{[i]}$ and $\mX_{[ij]}$. $\vx_{[i:j]}$ denotes the
subvector with indices $i,\ldots,j-1$, and $\vx_{[:j]}:=\vx_{[0:j]}$.
$\mathbf 1_n$, $\mathbf 1_{m\times n}$ and $\mathbf I$ are the all-ones
vector, the all-ones matrix and the identity. $\bDelta_n:=\{\vx\in\mathbb R^n_+:\sum_i\vx_{[i]}=1\}$
denotes the probability simplex, and $(a)_+:=\max\{a,0\}$. For
$\bmu,\bnu\in\bDelta_n$,
$\mathrm{KL}(\bmu\,\|\,\bnu):=\sum_i\bmu_{[i]}\log(\bmu_{[i]}/\bnu_{[i]})$ denotes the Kullback--Leibler divergence,
assuming $\bnu_{[i]}>0$ whenever $\bmu_{[i]}>0$. For symmetric matrix $\mX$,
$\lambda_{\min}(\mX)$ and $\lambda_{\max}(\mX)$ are its extreme eigenvalues,
and for $\mX\succ0$, $\kappa(\mX):=\lambda_{\max}(\mX)/\lambda_{\min}(\mX)$ is
its condition number. We write $\mX^{\succeq0}$ for a positive semi-definite (PSD)
matrix and $\mX^{\nsucceq0}$ else.

Further, let $\one=(1,\dots,1)^\top\in\mathbb R^n$ denote the all-ones vector, and $\one^\perp=\{\bm h\in\mathbb R^n:\one^\top\bm h=0\}$
be its orthogonal complement: the vectors whose entries sum to zero. If
$\bm p,\bm p'\in\bDelta_n$, then $\one^\top(\bm p'-\bm p)=1-1=0$, so
$\bm p'-\bm p\in\one^\perp$.

\paragraph{Submodularity ratio.}
Let $f : 2^{\V} \to \RR$ be a set function over ground set $\V$. For any subset
$\calS \subseteq \V$ and element $e \in \V \setminus \calS$, the marginal gain is
$\Delta f(e \mid \calS) = f(\calS \cup \{e\}) - f(\calS)$. The function $f$ is
\emph{submodular} if $\Delta f(e \mid \calA) \geq \Delta f(e \mid \calB)$ for all
$\calA \subseteq \calB \subseteq \V$ and $e \notin \calB$. The
\emph{submodularity ratio}~\citep{das2018approximate} captures how close a
non-submodular function is to being submodular:
\begin{equation}
\label{eq:gamma_L_kappa}
  \gamma_{\L,\kappa}(f) = \min_{\calS \subseteq \L,\;
    \calA:\,|\calA|\leq\kappa,\; \calA \cap \calS = \emptyset}
  \frac{\sum_{u \in \calA} \Delta f(u \mid \calS)}{f(\calS \cup \calA) - f(\calS)}.
\end{equation}
with $0/0 := 1$. If $\gamma > 0$, the greedy algorithm for
$\max_{|\calS| \leq \kappa} f(\calS)$ achieves a $(1 - e^{-\gamma})$ approximation to the
optimum~\citep{nemhauser1978analysis,das2018approximate}.

\noindent
\textbf{RSC and RSM}\label{sec:RSCRSM}. A differentiable function $l: \RR^{d} \rightarrow \RR$ is \textit{restricted strong concave} (RSC) with parameter $c_{\Omega}$ and \textit{restricted smooth} (RSM) with parameter $C_{\Omega}$ on a domain $\Omega\subset\RR^d \times \RR^d$ if $\forall\, (\bx, \by) \in \Omega$,
\begin{equation*}
 l(\bx)-l(\by)+\inner{\nabla l(\bx)}{\by-\bx} \in [\tfrac{c_{\Omega}}{2}\|\by-\bx\|_2^2,\tfrac{C_{\Omega}}{2}\|\by-\bx\|_2^2].
\end{equation*}
RSC and RSM have been studied in the context of weakly submodular functions \citep{elenberg2018restricted,gurumoorthy2019efficient}.
Let $\Omega_{\kappa} := \{ (\mathbf x, \mathbf y) : \|\mathbf x\|_0 \le \kappa,\; \|\mathbf y\|_0 \le \kappa,\; \mathbf x, \mathbf y \ge 0 \}$, and let $c_{\kappa}$ and $C_{\kappa}$ denote the RSC and RSM parameters of $l$ on $\Omega_{\kappa}$. Since $\Omega_k \subseteq \Omega_{\kappa}$ whenever $k \le \kappa$, it is easily verified that $c_k \ge c_{\kappa}$ and $C_k \le C_{\kappa}$. Finally, define $\widetilde{\Omega} := \{ (\mathbf x,\mathbf y) : \|\mathbf x - \mathbf y\|_0 \le 1 \}$, with associated smoothness parameter $\widetilde{C}_1$.

\textbf{Related Work.} Choosing the right mixture of training data is central
to efficient LLM fine-tuning, whether the goal is a specific task or broad
generalization. Early work measured task similarity with Maximum Mean
Discrepancy or reconstruction error to retrieve examples
\citep{achille2019task2vec, hwang2020data, alvarez2020geometric}; more recent
work represents tasks by lightweight LoRA adapters and uses low-rank
similarity to guide transfer \citep{kim-etal-2024-single}.

A second line of work estimates the influence of individual examples on a target
task. Influence functions \citep{koh2017understanding} are accurate but
expensive, so cheaper proxies track forgotten examples
\citep{toneva2018an} or gradient norms \citep{paul2021deep}. For
instruction tuning, LESS \citep{xia2024less} retrieves examples whose
low-rank gradient features match the target, and BIDS
\citep{dai2025improving} normalizes these scores per task to avoid favoring
high-gradient tasks.

Other work targets diversity and coverage, combining difficulty scores
with clustering \citep{zheng2023coveragecentric, maharana2024mathbbd} or
selecting coresets that approximate full training dynamics
\citep{sener2018active, killamsetty2021retrieve}. For multi-task instruction
tuning, SMART \citep{renduchintala-etal-2024-smart} allocates the budget
across tasks with a submodular objective and selects non-redundant examples.
Because scoring every example on a large model is costly, STAFF
\citep{zhang2025staff} scores with a smaller sibling model first, TSDS
\citep{liu2024tsds} uses optimal transport for scalable distribution
matching, and DELIFT \citep{agarwal2025delift} scores examples by their
usefulness as in-context prompts for one another. See \Cref{app:additional-related-work} for additional related work on
\emph{data mixture optimisation} and
\emph{task-affinity-based} reweighting.

\textbf{Contextualizing our framework.} These methods rank data by scalar
relevance scores, either per instance \citep{xia2024less, dai2025improving}
or per dataset \citep{achille2019task2vec, kim-etal-2024-single}, and do not
model redundancy between tasks. We instead pose mixture selection as
optimization over an energy landscape of pairwise task interactions,
encoded in a symmetric affinity matrix $\bS$ estimated from the predictive
behavior of single-task models.

% (Drop-in revision block) Abstract moved to Appendix for now.

% ---------------------------------------------------------------------
% NOTE: Additional affinity-geometry/projection details moved to the Appendix
% (see Supplementary Material).
% ---------------------------------------------------------------------
\section{Proposed Approach}

\paragraph{Problem setup.}\label{Sec:ProblemSetup}
We consider a set of $n$ instruction tuning tasks $\V :=\{\T_1,\dots,\T_n\}$, where each task $\T_i$ is associated with a corresponding dataset $\mathcal{D}_i$. Typically we consider supervised finetuning instruction dataset of the form  $\mathcal{D}_i = \{\bx_i, y_i\}_{i=1}^{n}$
where each $\bx_i$
is an input sequence and $y_i$
is the corresponding
output.

Our goal is to learn a sampling distribution $\vp^\ast \in \bDelta^n$ over n such tasks, which determines how training effort can be distributed across the collection of tasks.

\textbf{Redundancy and structure across tasks.}
Tasks in a training pool are neither independent nor equally informative. Redundancy can arise from overlapping datasets across collections, such as \textit{\textsc{BoolQ}}, \textit{\textsc{RTE}}, and \textit{\textsc{COPA}} appearing in both \textit{\textsc{FLAN}}~\citep{longpre2023flan} and \textit{\textsc{SuperGLUE}}~\citep{wang2019superglue}, from multiple prompt variants of the same dataset in \textit{\textsc{T0}}~\citep{sanh2021multitask}, or from tasks targeting similar capabilities, such as \textit{\textsc{MNLI}}, \textit{\textsc{SNLI}}, and \textit{\textsc{ANLI}} for natural language inference. In contrast, tasks such as \textit{\textsc{Evol-CodeAlpaca}} for code generation, \textit{\textsc{NuminaMath-TIR}} for mathematical reasoning, and \textit{\textsc{Aya}}~\citep{singh2024aya} for multilingual instruction following provide complementary coverage. Existing mixture methods often optimize tasks within predefined groups or relative to a fixed target \citep{renduchintala-etal-2024-smart,liu2024tsds}, which can miss these interactions. A task that is redundant in one mixture may be useful in another because of the other tasks selected. This motivates modeling the task pool jointly to account for redundancy, complementarity, and cross-task effects when allocating the training budget.

%\paragraph{Global Cross Task Similarity.} To capture this cross-task structure, we construct a symmetric similarity matrix $\bS \in \mathbb{R}^{n\times n}$, where $\bS_{[ij]}$ measures the affinity between tasks $\T_i$ and $\T_j$. Beyond pairwise similarities, we characterize the global relatedness of each task by its total similarity mass, $\bS_{[i]} := \sum_{j=1}^n \bS_{[ij]}$ which measures how strongly $\T_i$ is associated with the overall task mixture. Tasks with larger $\bS_{[i]}$ are broadly related to other tasks and thus capture more shared structure, whereas tasks with smaller $\bS_{[i]}$ exhibit more specialized behavior.

\paragraph{Cross-task Similarity.}
We collect pairwise task affinities in a symmetric matrix
$\bS\in\mathbb R^{n\times n}$, where $\bS_{[ij]}$ measures how related
$\T_i$ and $\T_j$ are (defined in \Cref{eq:PMI_matrix}). The \emph{degree}
$\bS_{[i]}:=\sum_{j}\bS_{[ij]}$ summarizes how strongly $\T_i$ relates to the
pool as a whole: tasks with larger degree share more structure with other
tasks, while tasks with smaller degree are more specialized.

\textbf{Energy-based modelling.}
We model the task pool as a dense Markov random field
(MRF)~\citep{kindermann1980markov,wainwright2008graphical}: each node is a
task, and the edge weight $\bS_{ij}$ is the affinity between tasks $\T_i$ and
$\T_j$. Modelling all tasks jointly lets the mixture account for coverage and
redundancy at once: broadly connected tasks gain weight, while highly similar
tasks compete for probability mass. We write
$\boldsymbol{\xi}_n=\{\T_i\}_{i=1}^n$ for the resulting
assignment of a sampling weight to each task.

\begin{figure}[t]
\centering
% Optimal mixture for the similarities drawn below (rho = 0.5, tau just above tau_c); sums to 1.
\pgfmathsetmacro{\pOne}{0.18}\pgfmathsetmacro{\pTwo}{0.07}\pgfmathsetmacro{\pThree}{0.21}
\pgfmathsetmacro{\pFour}{0.13}\pgfmathsetmacro{\pFive}{0.08}\pgfmathsetmacro{\pSix}{0.33}
% colour intensity of a task, proportional to its weight (light = small, dark = large)
\colorlet{mixbase}{blue!70!black}
\newcommand{\shadeof}[1]{\pgfmathsetmacro{\shade}{int(12+78*(#1)/0.33)}}
\begin{minipage}[c]{0.50\linewidth}
  \centering
  \begin{tikzpicture}[
      scale=0.78, transform shape,
      node/.style={circle, draw=blue!60!black, line width=0.5pt, minimum size=6.4mm, inner sep=0pt, font=\scriptsize},
      edge/.style={line width=0.8pt, draw=blue!55},
      strong/.style={line width=1.2pt, draw=blue!80!black},
      weak/.style={line width=0.55pt, draw=blue!35, dashed},
      edgelabel/.style={midway, fill=white, inner sep=0.6pt, font=\scriptsize, text=black!80}
  ]
  % --- Nodes (regular hexagon layout) ---
  \foreach \i/\ang/\p in {1/90/\pOne,2/150/\pTwo,3/30/\pThree,4/210/\pFour,5/330/\pFive,6/270/\pSix} {
    \shadeof{\p}
    \pgfmathsetmacro{\txt}{\shade>55 ? 1 : 0}
    \ifnum\txt=1 \def\tcol{white}\else\def\tcol{black!90}\fi
    \node[node, fill=mixbase!\shade!white, text=\tcol] (T\i) at (\ang:2.1) {$\T_{\i}$};
  }

  % --- Edges with weights (similarities S_{ij}); all pairs i<j ---
  \draw[strong] (T1) -- node[edgelabel] {$0.8$} (T2);
  \draw[strong] (T1) -- node[edgelabel] {$0.6$} (T3);
  \draw[weak]   (T1) -- node[edgelabel] {$0.2$} (T4);
  \draw[weak]   (T1) to[bend left=12] node[edgelabel] {$0.3$} (T5);
  \draw[edge]   (T1) to[bend right=12] node[edgelabel] {$0.5$} (T6);

  \draw[edge]   (T2) -- node[edgelabel] {$0.4$} (T3);
  \draw[edge]   (T2) -- node[edgelabel] {$0.3$} (T4);
  \draw[weak]   (T2) to[bend left=10] node[edgelabel] {$0.2$} (T5);
  \draw[strong] (T2) -- node[edgelabel] {$0.6$} (T6);

  \draw[edge]   (T3) to[bend left=10] node[edgelabel] {$0.5$} (T4);
  \draw[edge]   (T3) -- node[edgelabel] {$0.4$} (T5);
  \draw[strong] (T3) -- node[edgelabel] {$0.7$} (T6);

  \draw[strong] (T4) -- node[edgelabel] {$0.8$} (T5);
  \draw[edge]   (T4) to[bend left=10] node[edgelabel] {$0.4$} (T6);

  \draw[edge]   (T5) -- node[edgelabel] {$0.5$} (T6);
  \end{tikzpicture}
\end{minipage}%
\begin{minipage}[c]{0.50\linewidth}
  \centering
  \begin{tikzpicture}[x=7.6cm, y=0.52cm, font=\scriptsize]
    \def\xmax{0.36}
    % axis with ticks, title above
    \draw[black!70] (0,0.35) -- (\xmax,0.35);
    \foreach \xt in {0,0.1,0.2,0.3} {
      \draw[black!70] (\xt,0.42) -- (\xt,0.28);
      \node[above, inner sep=1pt, black!70] at (\xt,0.42) {\pgfmathprintnumber[fixed,precision=1]{\xt}};
    }
    \node[font=\bfseries\scriptsize, black!70] at (\xmax/2,1.75) {Optimal mixture $\vp^\ast$};
    % bars: row y = 1..6; shade proportional to weight, value printed after the bar
    \foreach \y/\p/\lab in {1/\pOne/$\T_1$,2/\pTwo/$\T_2$,3/\pThree/$\T_3$,4/\pFour/$\T_4$,5/\pFive/$\T_5$,6/\pSix/$\T_6$} {
      \shadeof{\p}
      \draw[fill=mixbase!\shade!white, draw=blue!60!black, line width=0.4pt] (0,-\y+0.15) rectangle (\p,-\y+0.85);
      \node[anchor=east] at (-0.008,-\y+0.5) {\lab};
      \node[anchor=west, text=black!80, inner sep=1.5pt] at (\p,-\y+0.5) {\pgfmathprintnumber[fixed,precision=2]{\p}};
    }
  \end{tikzpicture}
\end{minipage}
\caption{\textit{Illustrative example.} Left: a six-task MRF whose edges are labelled with similarities $\bS_{ij}$ (thicker edges for high $\bS_{ij}$, dashed for low). Right: the optimal mixture $\vp^\ast$ from Eq.~(\ref{Eq: original opt}) for these similarities; darker shading marks larger weight in both panels. The most connected task $\T_6$ receives the most mass, while the strongly redundant pairs ($\T_1,\T_2$ and $\T_4,\T_5$, both with $\bS_{ij}=0.8$) split theirs.}
\end{figure}
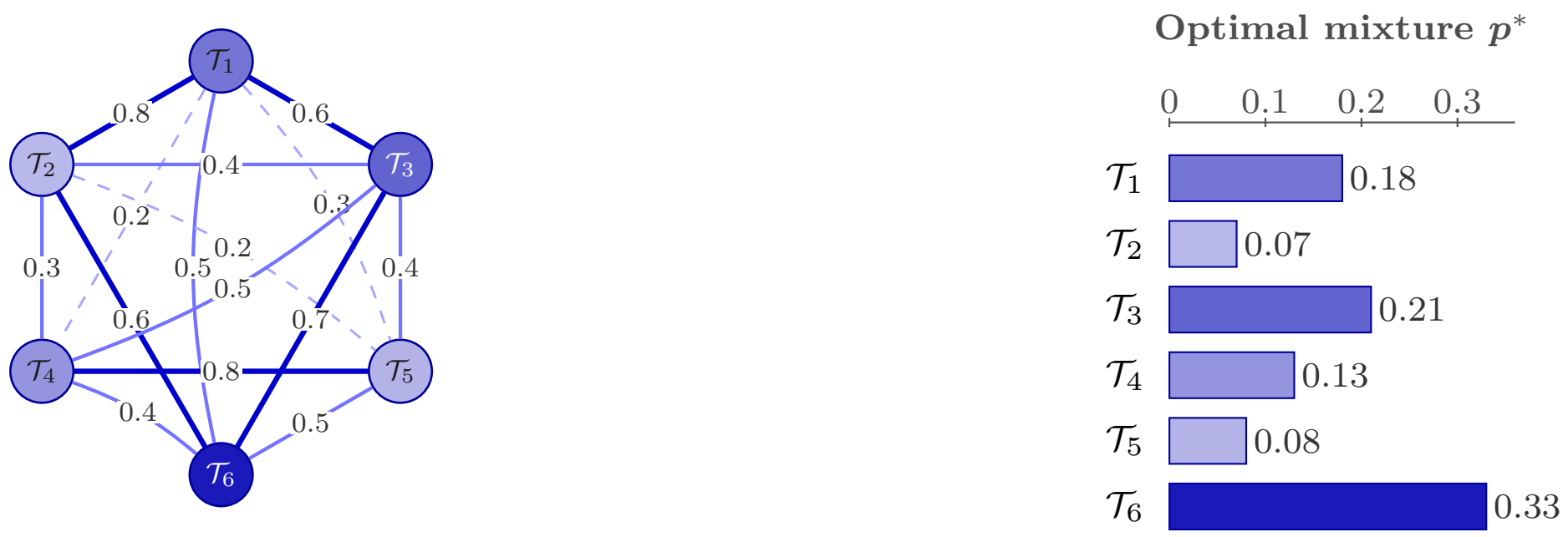

\subsection{Task Selection via Energy-Based Model}

Each task receives a \emph{unary potential} proportional to its degree,
$\bPsi_{\texttt{un}}:=\beta\,\bS\one$ (so ${\bPsi_{\texttt{un}}}_{[i]}=\beta\bS_{[i]}$),
and each pair of tasks a \emph{pairwise potential} proportional to their
affinity, $\bPsi_{\texttt{pair}}:=\lambda\bS$, where $\beta,\lambda>0$ set the
strength of each term. The energy of a mixture $\vp\in\bDelta_n$ is
\begin{equation}
  \begin{aligned}
    \maxop_{\vp\in\bDelta_n}\ \mathbb{E}(\vp)
    &= \sum_{i=1}^n {\bPsi_{\texttt{un}}}_{[i]} \vp_{[i]}
       - \frac{1}{2}\sum_{i=1}^{n}\sum_{j=1}^{n}{\bPsi_{\texttt{pair}}}_{[ij]}\vp_{[i]} \vp_{[j]} \\
    &= \mathcolorbox{teal!12}{\bPsi_{\texttt{un}}^{\top}\vp}\; -\;
       \mathcolorbox{violet!12}{\tfrac{1}{2}\vp^{\top}\bPsi_{\texttt{pair}}\vp}.
  \end{aligned}
  \label{Eq: original opt max}
\end{equation}
The first term rewards $\mathcolorbox{teal!12}{\text{representative}}$ tasks,
those with high total affinity to the pool; the second penalizes placing mass
on $\mathcolorbox{violet!12}{\text{\itshape similar pairs}}$, which promotes
diversity. Equivalently, we minimize $\hat{\mathbb E}(\vp):=-\mathbb E(\vp)$:
\begin{equation}
  \minop_{\vp\in\bDelta_n}\ \hat{\mathbb{E}}(\vp)
  = -\bPsi_{\texttt{un}}^{\top}\vp + \tfrac{1}{2}\vp^{\top}\bPsi_{\texttt{pair}}\vp .
  \label{Eq: original opt}
\end{equation}
Dividing by $\lambda$ does not change the minimizer, so it depends only on
the ratio $\rho\coloneqq\beta/\lambda$: larger $\rho$ favors high-degree tasks, smaller $\rho$
weights redundancy more heavily. \Cref{sec:mixture} shows that the minimizer is the projection of the scaled uniform allocation $\rho\one$ onto the simplex in the geometry of task affinities, and \Cref{sec:b2} explains how $\rho$ controls coverage.

\begin{comment}
\citep{yarotsky2026corner} actually discusses the same quadratic formulation [\textcolor{red}{\textbf{More to elaborate}}]
\end{comment}

\setlength{\parskip}{0pt}%
\setlength{\parindent}{0pt}%

\paragraph{When is the objective convex?}
If $\bPsi_{\texttt{pair}}\succeq0$, \Cref{Eq: original opt} is a convex
quadratic program over the simplex and can be solved with standard methods
such as projected gradient descent (PGD;
\citealp[Chapter~7]{wright2022optimization}) or entropic mirror descent
\citep{beck2003mirror}. PGD then converges to the global optimum at rate
$O(1/t)$ with constant step size $\eta<1/L$, where
$L=\lambda_{\max}(\bPsi_{\texttt{pair}})$ is the Lipschitz constant of the
gradient; pseudocode is given in \Cref{alg:mixture}.

\paragraph{Non-convexity of the objective.}
Strictly, PGD needs $\bPsi_{\texttt{pair}}$ to be positive semidefinite only
on the directions the simplex allows. As we show next, the affinity matrix
in our setting does not satisfy even this weaker condition, so we modify the
objective to restore convexity.

\paragraph{Non-PSD Correction via Spectrum Shifting.}
When the pairwise similarity matrix $\bPsi_{\texttt{pair}}$ is not positive semidefinite (PSD), i.e., $\bPsi_{\texttt{pair}}\nsucceq 0$, the resulting optimization problem is no longer convex. To restore convexity, we apply a spectrum-shifting correction \citep{chen2009learning,wu2005analysis} to obtain a positive semidefinite matrix.

A common way to restore convexity while maintaining a positive spectral margin is to shift the spectrum of $\bPsi_{\texttt{pair}}$ so that its smallest eigenvalue is at least a small constant $\epsilon>0$, i.e., $\bPsi_{\texttt{pair}}^{\succeq\epsilon\mathbf I}:=\bPsi_{\texttt{pair}}+\bigl(\epsilon-\lambda_{\min}(\bPsi_{\texttt{pair}})\bigr)_{+}\mathbf I$. This correction ensures $\lambda_{\min}(\bPsi_{\texttt{pair}}^{\succeq\epsilon\mathbf I})\geq\epsilon$, thereby yielding a positive-definite quadratic objective. However, the shift also modifies the original objective by introducing the additional isotropic regularization term $\bigl(\epsilon-\lambda_{\min}(\bPsi_{\texttt{pair}})\bigr)_{+}\norm{\vp}_{2}^{2}$. We read this shift as restoring the self-affinity that the hollow matrix $\bS$ leaves out, and discuss how large it needs to be in \Cref{app:shift}.

%When the required shift is large, indicating substantial negative curvature in $\bPsi_{\texttt{pair}}$, this induced $\ell_2$ regularization can dominate the original objective and bias the optimum toward the uniform distribution, which minimizes $\norm{\vp}_{2}^{2}$ over the simplex.

\paragraph{Critical shift $\tau_c$.}\label{para:critical-shift} %Positive definiteness on all of $\mathbb R^n$ is more than convexity on the simplex requires.
Any two learnt mixtures $\vp,\vp'\in\bDelta_n$ differ by $\bm h=\vp'-\vp\in\one^\perp$, so
$\hat{\mathbb E}(\cdot)$ is strictly convex on $\bDelta_n$ if and only if
$\bm h^\top(\bS+\tau\mathbf I)\bm h>0$ for all nonzero $\bm h\in\one^\perp$.
Let $\bm U\in\mathbb R^{n\times(n-1)}$ have orthonormal columns spanning
$\one^\perp$. We define the \emph{critical shift}
\begin{equation}
  \tau_c:=-\lambda_{\min}\!\big(\bm U^\top\bS\bm U\big)
  \label{eq:tauc}
\end{equation}
which is the negative of smallest eigenvalue of $\bS$ restricted to zero-sum directions.
Then $\hat{\mathbb E}(\cdot)$ is strictly convex on $\bDelta_n$ exactly when
$\tau>\tau_c$, whereas $\bS+\tau\mathbf I\succ0$ requires $\tau>\tau_0$, where $\tau_0 = -\lambda_{\min}\!\big(\bS)$ .
Since $\bm U^\top\bS\bm U$ is a compression of $\bS$, Cauchy interlacing \citep{horn2012matrix} gives
$\tau_c\le\tau_0$ (Lemma~\ref{lem:cpsd}). For our affinity $\tau_c=17.09$ and $\tau_0=186.30$; \Cref{app:spectrum} traces this gap to a single outlier eigenvalue of $\bS$ whose eigenvector is close to $\one$.

\textbf{Invariance to a constant offset.}\label{para:offset-invariance}
Adding the same constant $c$ to every affinity does not change the optimal
mixture. On $\bDelta_n$ we have $\one^\top\vp=1$, so replacing $\bS$ by
$\bS+c\one\one^\top$ changes the pairwise term by
$\tfrac{\lambda}{2}c(\one^\top\vp)^2=\tfrac{\lambda}{2}c$ and the unary term
$\beta\,\one^\top\bS\vp$ by $\beta cn$; both are constants. The two shifts
respond differently to such an offset. Because $\bm U^\top\one=\bm 0$, we have
$\bm U^\top(\bS+c\one\one^\top)\bm U=\bm U^\top\bS\bm U$, so $\tau_c$ is
unchanged, whereas $\tau_0=-\lambda_{\min}(\bS+c\one\one^\top)$ increases
monotonically towards $\tau_c$ as $c\to\infty$. Hence $\tau_c$ is intrinsic to
the mixture problem, and $\tau_0$ depends on an offset the objective ignores
(\Cref{app:shift}).

\begin{figure*}[t]
  \centering
  \begin{subfigure}[t]{0.32\textwidth}\centering
    \includegraphics[width=\linewidth]{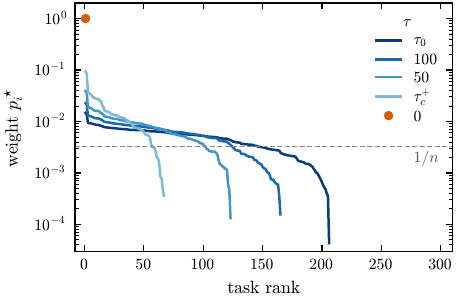}
    \caption{Sorted weights $\vp^\ast_{[i]}(\tau)$.}\label{fig:shift-rank}
  \end{subfigure}\hfill
  \begin{subfigure}[t]{0.32\textwidth}\centering
    \includegraphics[width=\linewidth]{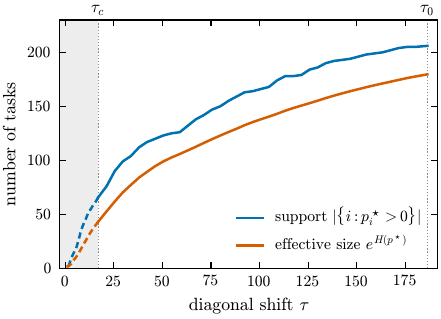}
    \caption{Support and effective size.}\label{fig:shift-breadth}
  \end{subfigure}\hfill
  \begin{subfigure}[t]{0.32\textwidth}\centering
    \includegraphics[width=\linewidth]{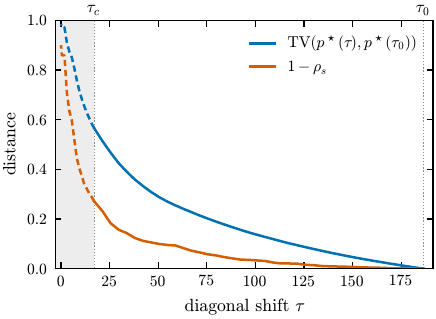}
    \caption{Distance to $\vp^\ast(\tau_0)$.}\label{fig:shift-distance}
  \end{subfigure}
  \caption{Effect of the spectrum shift $\tau$ in $\bS+\tau\mathbf I$ on the
  optimal mixture ($n=309$, $\rho=0.02424$). Shaded: $\tau\le\tau_c$, where the
  objective is not convex on $\bDelta_n$. Dashed lines show the local minimum
  reached from $\vp^\ast(\tau_0)$. At $\tau=0$ the solution is a single task.
  Per-collection mass is in \Cref{fig:shift-mass}; discussion in \Cref{app:shift}.}
  \label{fig:shift}
\end{figure*}

%\begin{rem}
%\textbf{Representative/Diversity Tradeoff } For large values of $\rho = \frac{\beta}{\lambda}\uparrow$, the learned optimal task probabilities $\vp^{\ast}$ is pulled towards more higher unary-mass regions, favoring tasks that are individually most representative. Conversely, for small values $\rho =\frac{\beta}{\lambda}\downarrow$ , the solution promotes spread-outmass allocation, encouraging diversity by penalizing co-occurrence in the similarity space.\end{rem}

\subsection{Design
  Choices: Pairwise Potentials}

  Our objective function in ~\Cref{Eq: original opt} depends critically on
  modeling pairwise interactions between tasks. To capture how task pairs correlate,
  it is essential to define a similarity metric that robustly encodes these
  relationships.

\textbf{Pitfalls of Semantic Similarity} Prior work \citep{renduchintala-etal-2024-smart} often relies
  on semantic similarity measures between tasks; however, these approaches are
  restrictive and agnostic to downstream model behavior. \Cref{fig:cosine} in \Cref{app:similarity-notions} shows that cosine similarity between task vectors carries little information about task structure.

  We thereby move towards a more grounded similarity measure.

\textbf{Pointwise Mutual Information} Given two tasks $\textcolor{AccentBlue}{\T_{i}}, \textcolor{AccentOrange}{\T_{j}}$ and corresponding datasets associated with it $\D_{\textcolor{AccentBlue}{\T_i}}=\{\bx^{\textcolor{AccentBlue}{\T_i}}_{q}, y^{\textcolor{AccentBlue}{\T_i}}_{q}\}_{q=1}^{m_1}$ and
  $\D_{\textcolor{AccentOrange}{\T_j}}= \{\bx^{\textcolor{AccentOrange}{\T_j}}_{r}, y^{\textcolor{AccentOrange}{\T_j}}_{r}\}_{r=1}^{m_2}$ respectively, we define the similarity
  score across two tasks $\textcolor{AccentBlue}{\T_{i}}$ and $\textcolor{AccentOrange}{\T_{j}}$ denoted as $\bS_{\mathrm{PMI}}(\textcolor{AccentBlue}{\T_{i}}
  ; \textcolor{AccentOrange}{\T_{j}})$
\begin{equation}
\begin{aligned}
\bS_{\mathrm{PMI}}(\textcolor{AccentBlue}{\T_{i}}; \textcolor{AccentOrange}{\T_{j}})
:= \frac{1}{2}\Bigg[
&\frac{1}{m_2}\sum_{r=1}^{m_2}\log\frac{\PP_{\btheta^{\ast}(\textcolor{AccentBlue}{\T_i})}\!\left(y^{\textcolor{AccentOrange}{\T_j}}_{r}\mid\bx^{\textcolor{AccentOrange}{\T_j}}_{r}\right)}{\PP_{\btheta^{\ast}(\textcolor{AccentOrange}{\T_j})}\!\left(y^{\textcolor{AccentOrange}{\T_j}}_{r}\mid\bx^{\textcolor{AccentOrange}{\T_j}}_{r}\right)} \\
&\quad+ \frac{1}{m_1}\sum_{q=1}^{m_1}\log\frac{\PP_{\btheta^{\ast}(\textcolor{AccentOrange}{\T_j})}\!\left(y^{\textcolor{AccentBlue}{\T_i}}_{q}\mid\bx^{\textcolor{AccentBlue}{\T_i}}_{q}\right)}{\PP_{\btheta^{\ast}(\textcolor{AccentBlue}{\T_i})}\!\left(y^{\textcolor{AccentBlue}{\T_i}}_{q}\mid\bx^{\textcolor{AccentBlue}{\T_i}}_{q}\right)}
\Bigg]
\end{aligned}
\label{eq:PMI_matrix}
\end{equation}

  where $\btheta^{\ast}(\T_{i}) := \btheta_{0}+ \boldsymbol\tau({\T_i})$ , $\boldsymbol
  \tau({\T_i})$ indicating the task vector for task $\T_{i}$ \citep{ortizjimenez2023taskarithmetic_tangent}
  and $\PP_{\btheta^{\ast}(\T_{i})}$
  indicates the next token probability scores under converged finetuned
  model parameter $\btheta^{\ast}(\T_{i})$. How the scores are computed (models, held-out examples, cost) is described in \Cref{app:affinity-impl}, and how to read their sign in \Cref{app:affinity-geometry-details}.

%\textbf{Reading the affinity.} $\bS_{ij}$ is the symmetrised average log-likelihood ratio between the single-task models of $\T_i$ and $\T_j$, each evaluated on the other's data. It is pointwise mutual information in the sense of \citet{church1990word}, between the task identity and the observed response, and not mutual information between two random variables. Because each model fits its own task best, $\bS_{ij}\le0$ for most pairs ($95\%$ of off-diagonal entries) and every degree $\bS_{[i]}=\sum_j\bS_{[ij]}$ is negative, ranging from $-1241$ to $-3.4$. A task is therefore \emph{representative} when its degree is \emph{least negative}: other tasks' models predict it almost as well as its own model does. By the invariance above, only the differences between degrees matter.
% [CONFIRM] if you prefer a new name, e.g. "cross-task log-likelihood affinity (CLA)", replace \bS_{\mathrm{PMI}} throughout.

\begin{figure}[t]
  \centering
  \includegraphics[width=0.6\linewidth]{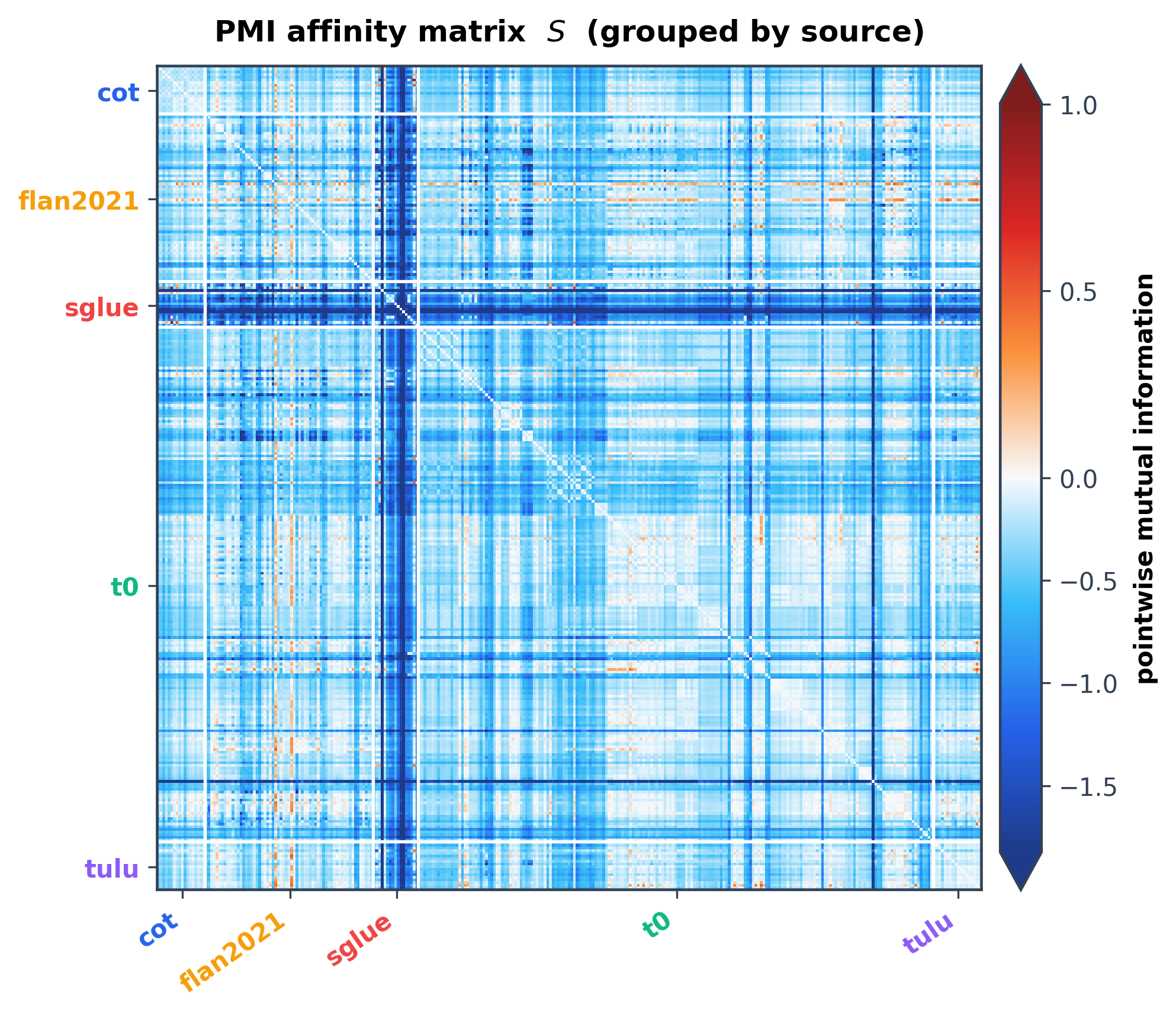}
  \vspace{-0.5em}
  \caption{HeatMap of PMI Affinity Matrix across all 309 tasks as specified in \Cref{sec:datasets_tasks}}
  \vspace{-0.75em}
  \label{fig:a1-matrix}
\end{figure}

  %\textbf{Incremental Composition of adding new tasks}: To model incremental composition, we define the independent set family $\mathcal{I}= \{\S \subseteq \V \mid |\S| \leq k\}$, and pose the top-$k$ task selection problem as $\max_{A \in \mathcal{I}}f(\tilde{A}),$ which lifts the continuous optimization (in Eq: \eqref{Eq: original opt max}) to a discrete set function defined over subsets of tasks. This formulation encourages incremental construction of $\boldsymbol{\xi}_{k}$ by choosing the set  $\Bar{\A}$ that supports the highest relaxed utility score under $\mathbb{E}(\cdot)$.

\subsection{Task Discovery}

\paragraph{From continuous mixtures to a set function over tasks.}

Given a current task mixture $\boldsymbol{\xi}_{k}$ composed of $k$ tasks, our goal is to
  evaluate the marginal utility of introducing a new candidate task $\T_{k+1}$ to form
  an augmented mixture $\boldsymbol{\xi}_{k+1}$. Considering $\V$ as defined earlier (in \emph{Problem Setup}: ~\Cref{Sec:ProblemSetup}) as the set of $n$ tasks, for
$\A\subseteq\V$ with index set $\Bar{\A}\subseteq[n]$, we define
\begin{equation}
  f(\Bar{\A}) := \max_{\vp\in\bDelta_n,\ \operatorname{supp}(\vp)\subseteq\Bar{\A}}\mathbb{E}(\vp),
  \label{eq:weak-submodularity}
\end{equation}
with maximizer $\zeta^{(\Bar{\A})}$, so that
$f(\Bar{\A})=\mathbb{E}(\zeta^{(\Bar{\A})})$.

\textbf{Incremental composition.}
Selecting at most $k$ tasks therefore reduces to the problem
$\max_{\A\in\mathcal I} f(\Bar{\A})$ with
$\mathcal{I}=\{\A\subseteq\V : |\A|\le k\}$. This turns the continuous
problem ~\Cref{Eq: original opt max} into a set function over tasks, and
lets us build $\boldsymbol{\xi}_k$ one task at a time by adding the task that
most increases $f(\cdot)$.

\noindent\textbf{Task affinity.}
To measure how much adding a new task changes the overall mixture, we compare the
learnt probability mixtures of $\boldsymbol{\xi}_k$ and $\boldsymbol{\xi}_{k+1}$, given by
$\vp^\ast(k)$ and $\vp^\ast(k+1)$ (with a slight abuse of notation,
$\vp^\ast(k):= \zeta^{\Bar{(\boldsymbol{\xi}_k)}}$ denotes the optimal mixture computed over $k$
tasks). We compare $\vp^\ast(k)$ with the restriction of $\vp^\ast(k+1)$ to
the first $k$ tasks as
$\mathrm{TV}\big(\vp^\ast(k),{\vp^\ast(k+1)}_{[:k]}\big)
:=\tfrac{1}{2}\sum_{i=0}^{k-1}\big|{\vp^\ast(k)}_{[i]}-{\vp^\ast(k+1)}_{[i]}\big|$.

\begin{rem}
Intuitively, this metric tells us how much the existing mixture shifts when a new task joins it. A small value means the weights of the first $k$ tasks barely move, which typically happens when the new task is largely redundant
with those already selected.
\end{rem}

\begin{figure*}[t]
  \centering
  \begin{subfigure}[t]{0.32\textwidth}\centering
    \includegraphics[width=\linewidth]{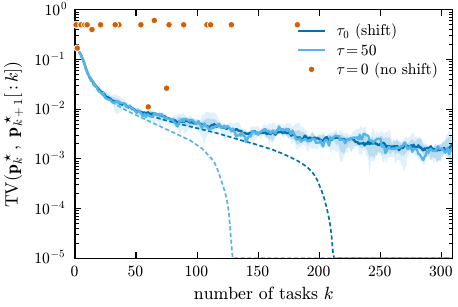}
    \caption{Affinity as defined.}\label{fig:incr-tv}
  \end{subfigure}\hfill
  \begin{subfigure}[t]{0.32\textwidth}\centering
    \includegraphics[width=\linewidth]{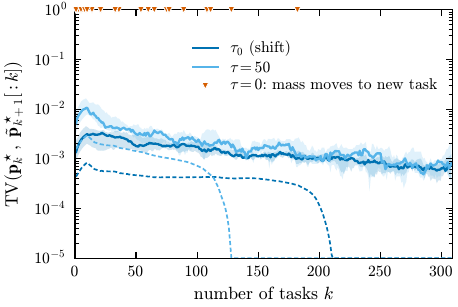}
    \caption{Renormalised affinity $\widetilde{\mathrm{TV}}_k$.}\label{fig:incr-tvr}
  \end{subfigure}\hfill
  \begin{subfigure}[t]{0.32\textwidth}\centering
    \includegraphics[width=\linewidth]{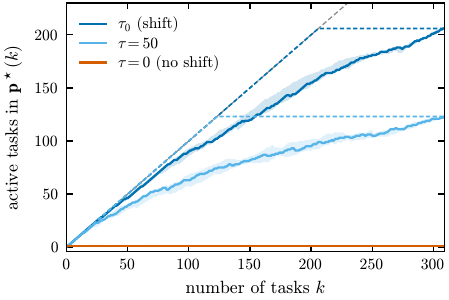}
    \caption{Active tasks in $\vp^\ast(k)$.}\label{fig:incr-active}
  \end{subfigure}
  \caption{Incremental composition. Solid: median over five random insertion
  orders (band: range). Dashed: greedy order. TV curves are a 15-step moving
  average. Markers: steps at which the unshifted ($\tau=0$) mixture changes,
  each moving all mass to the new task. Grey dashed line in (c): all $k$
  tasks active.}
  \label{fig:incr}
\end{figure*}

\textbf{Renormalised affinity.} Since
$\sum_{i<k}\big(\vp^\ast(k)_{[i]}-\vp^\ast(k{+}1)_{[i]}\big)=\vp^\ast(k{+}1)_{[k]}$,
the affinity above equals $\tfrac12\vp^\ast(k{+}1)_{[k]}$ whenever no
existing task gains mass. In our runs this holds at every step where any
existing weight changes, so the measure records the weight given to the new
task rather than any reshuffling of the existing ones. We therefore also
report
$\widetilde{\mathrm{TV}}_k:=\mathrm{TV}\big(\vp^\ast(k),\,\vp^\ast(k{+}1)_{[:k]}/(1-\vp^\ast(k{+}1)_{[k]})\big)$,
which compares the two mixtures on the first $k$ tasks after renormalisation. Both are reported along greedy and random insertion orders in \Cref{fig:incr} and discussed in \Cref{app:incremental}.

%\paragraph{Solver.Problem~\eqref{eq:obj} is a strongly convex QP over the simplex. We solve it by accelerated projected gradient (FISTA; \citealp{beck2009fast}) with step$1/L$,  $L=\lambda_{\max}(\widetilde{\bm S})$ (the objective divided by $\lambda$, which leaves the minimiser unchanged), and the exact Euclidean simplex projection \citep{duchi2008efficient,condat2016fast}, stopping when $\lVert\bm p_{k+1}-\bm p_k\rVert_1<10^{-12}$. The result is the exact minimiser of \eqref{eq:obj}, so $\bm p^\star$ satisfies the full KKT conditions, including complementary slackness on the inactive tasks.

\begin{algorithm}[t]
\caption{Algorithm for \methodprop{}}
\label{alg:mixture}
\small
\begin{algorithmic}[1]
\Require affinity $\bS$, ratio $\rho>0$, shift $\tau>\tau_c$, tolerance $\epsilon$
\Ensure $\vp^\star=\argmin_{\vp\in\bDelta_n}\tfrac12\vp^\top\widetilde{\bS}\vp-\rho\,\one^\top\widetilde{\bS}\vp$
\State $\widetilde{\bS}\gets\bS+\tau\mathbf I$,\; $L\gets\lambda_{\max}(\widetilde{\bS})$,\; $\vp_0=\bm y_0\gets\tfrac1n\one$,\; $t_0\gets1$
\For{$j=0,1,2,\dots$}
  \State $\vp_{j+1}\gets\Pi_{\bDelta_n}\!\big(\bm y_j-\widetilde{\bS}(\bm y_j-\rho\one)/L\big)$
  \State $t_{j+1}\gets\tfrac12\big(1+\sqrt{1+4t_j^2}\big)$
  \State $\bm y_{j+1}\gets\vp_{j+1}+\tfrac{t_j-1}{t_{j+1}}(\vp_{j+1}-\vp_j)$
  \State \textbf{if} $\lVert\vp_{j+1}-\vp_j\rVert_1<\epsilon$ \textbf{then return} $\vp_{j+1}$
\EndFor
\end{algorithmic}
\end{algorithm}

\textbf{Solver.} \Cref{alg:mixture} is accelerated projected gradient (FISTA) on \Cref{Eq: original opt} divided by $\lambda$, with the exact Euclidean projection onto $\bDelta_n$. For $\tau>\tau_c$ the problem is strictly convex on the simplex, so the algorithm returns the unique minimizer. With the settings used in all experiments ($\rho=0.02424$, $\tau=\tau_0+10^{-3}$, $\epsilon=10^{-14}$) it converges in $38$ iterations to a point that satisfies the full KKT conditions; details and verification are in \Cref{app:algorithm}.

\textbf{Instance-level sampling.} Given the learnt mixture $\vp^\ast\in\bDelta_n$, we
allocate a budget of $\mathcal{B}$ training instances across the $n$ tasks by drawing
$\mathcal{B}$ i.i.d.\ task indices from $\vp^\ast$. The resulting counts
$\bm k=(k_1,\dots,k_n)\in\mathbb{Z}_{\ge0}^n$, with $\sum_{i=1}^n k_i=\mathcal{B}$,
then follow $\bm k\sim\mathrm{Mult}(\mathcal{B},\vp^\ast)$:
\[
  \mathbb{P}(\bm k\mid \mathcal{B},\vp^\ast)=\frac{\mathcal{B}!}{\prod_{i=1}^n k_i!}\prod_{i=1}^n\big(\vp^\ast_{[i]}\big)^{k_i},
  \qquad \mathbb{E}[k_i]=\mathcal{B}\,\vp^\ast_{[i]}.
\]
We then sample $k_i$ instances uniformly from task $\T_i$, so tasks outside the support of $\vp^\ast$ contribute nothing. In practice we average five such draws, round the result to integers that sum to $\mathcal B$, and cap each count at the size of its task (\Cref{app:sampling}).

\section{Theoretical Results}\label{sec:theory}

  The continuous view of \Cref{Eq: original opt} yields mixture weights, but practical
  pipelines often require \emph{budgeted} variants (e.g., selecting a subset of tasks and then optimizing
  proportions on that subset). To justify greedy construction in this discrete setting, we analyze
  the set function $f(\cdot)$ of Eq.~\eqref{eq:weak-submodularity}, defined for nonempty supports.
  Any two points of $\bDelta_n$ always differ by a zero-sum vector, so the relevant curvature constants are the
  \emph{zero-sum restricted eigenvalues} of $\bPsi_{\texttt{pair}}$:
  for $k\ge2$, $m_k$ and $M_k$ are the minimum and maximum of $\bm h^\top\bPsi_{\texttt{pair}}\bm h/\|\bm h\|_2^2$
  over nonzero $\bm h\in\one^\perp$ with at most $k$ nonzero entries.
  For $\bPsi_{\texttt{pair}}=\lambda(\bS+\tau\mathbf I)$ these satisfy
  $m_k\ge\lambda(\tau-\tau_c)$ and $M_k\le\lambda(\tau+\lambda_{\max}(\bm U^\top\bS\bm U))$ for every $k$,
  with $\tau_c$ the critical shift of \eqref{eq:tauc}.
  Full proofs appear in Appendix~\ref{app:submod-proofs}.

  \begin{restatable}[Monotonicity]{lemm}{lemmMonotonicity}
  \label{lem:monotonicity}
    The set function $f(\cdot)$ defined in Eq \eqref{eq:weak-submodularity} is monotone: for any nonempty $\mathcal{P} \subseteq \mathcal{Q}\subseteq\V$, $f(\mathcal{P}) \leq f(\mathcal{Q}).$
  \end{restatable}

\begin{proofsketch}
If $\mathcal{P}\subseteq\mathcal{Q}$, then any mixture supported on $\mathcal{P}$ is also feasible for $\mathcal{Q}$, so the feasible set only grows.
Maximizing the same objective over a larger feasible set cannot reduce the optimum, hence $f(\mathcal{P})\le f(\mathcal{Q})$.
\end{proofsketch}

  \begin{restatable}[Restricted strong concavity and smoothness]{lemm}{lemmFiniteRscRsm}
  \label{lem:finite-rsc-rsm}
   Let $\bPsi_{\texttt{pair}}$ be symmetric. For all $\vp,\vq\in\bDelta_n$ with $|\supp(\vp)\cup\supp(\vq)|\le k$,
    the energy $\mathbb{E}(\vp) = \bPsi_{\texttt{un}}^{\top}\vp - \frac{1}{2}\vp^{\top}\bPsi_{\texttt{pair}}\vp$ of \eqref{Eq: original opt max} satisfies
    {\setlength{\abovedisplayskip}{2pt}%
     \setlength{\belowdisplayskip}{2pt}%
     \begin{equation*}
       \begin{aligned}
         -\frac{M_k}{2}\|\vp-\vq\|_{2}^{2}
         &\leq \mathbb{E}(\vp)-\mathbb{E}(\vq)-\nabla\mathbb{E}(\vq)^{\top}(\vp-\vq)\\
         &\leq -\frac{m_k}{2}\|\vp-\vq\|_{2}^{2},
       \end{aligned}
     \end{equation*}
    }
    with finite constants, and $m_k>0$ for every $k$ whenever $\tau>\tau_c$.
  \end{restatable}

\begin{proofsketch}
The Bregman remainder of the quadratic equals $-\tfrac12\bm h^{\top}\bPsi_{\texttt{pair}}\bm h$ with $\bm h=\vp-\vq$. This vector lies in $\one^\perp$ and has at most $k$ nonzero entries. Apply the definition of $m_k$ and $M_k$.
\end{proofsketch}

  \begin{restatable}[Weak submodularity]{thm}{weaksubmodularity}
  \label{thm:weak-submodularity}
    Let $\mathcal{L},\S\subseteq\V$ be disjoint with $\mathcal{L}\neq\emptyset$, and suppose $m_{|\mathcal{L}|+|\S|}>0$
    (in particular, whenever $\tau>\tau_c$). Then the submodularity ratio of $f$ satisfies
    \begin{equation*}
      \begin{aligned}
        \gamma_{\mathcal{L},\S}
        &:=\frac{\sum_{j\in\S}\big[f(\mathcal{L}\cup\{j\})-f(\mathcal{L})\big]}{f(\mathcal{L}\cup\S)-f(\mathcal{L})} \ge\frac{m_{|\mathcal{L}|+|\S|}}{2\,M_{|\mathcal{L}|+1}}\\
        &\ge\frac{\tau-\tau_c}{2\big(\tau+\lambda_{\max}(\bm U^\top\bS\bm U)\big)}\;>\;0 .
      \end{aligned}
    \end{equation*}
  \end{restatable}

  The proof (Appendix~\ref{app:weak-submod-proof}) adapts the restricted-strong-concavity argument of
  ProtoDash \citep[Thm.~4.3]{gurumoorthy2019efficient} and \citep{elenberg2018restricted} from the nonnegative orthant to the simplex.
  A new task can only enter by taking mass from the current support, along $\bm e_j-\zeta$ with step at most $1$, which costs a factor $2$.
  On a constrained face, the first-order condition is centered at the multiplier $\nu$, so the bound is naturally expressed in terms of $\nabla_j\mathbb{E}-\nu$.
  Combined with monotonicity, the greedy algorithm started from the best single task attains
  $f(\A_k)-f(\A_1)\ge(1-e^{-\gamma_k})(\mathrm{OPT}_{k-1}-f(\A_1))$ with $\gamma_k=m_{2k-1}/(2M_{k+1})$
  (Corollary~\ref{cor:greedy-app}, via \citealp{das2011submodular}).

Together, these properties support greedy budgeted task discovery with approximation guarantees. We instantiate the bound numerically (at $\tau=\tau_0$) and empirically verify weak submodularity in Appendix~\ref{app:weak-submod-exp} (\Cref{fig:submod}). \Cref{app:incremental} studies how the mixture evolves when tasks are added one at a time.

\begin{comment}
  \textbf{3D Surface Plot (\texttt{3d\_surface\_stacked\_all\_in\_one.pdf})}
  To visualize how the mixture changes as we grow it, we generate a 3D surface
  plot over the greedy construction trajectory. For each step $k\in\{1,\ldots,K\}$
  and each candidate next dataset $i$, we compute the affinity score between the
  current $k$-dataset mixture and the hypothetical $(k+1)$-dataset mixture obtained
  by adding $i$ (using the total-variation-based affinity defined above). The
  surface height corresponds to this affinity (lower values indicate that adding
  $i$ perturbs the existing mixture less), and we stack surfaces for different
  similarity constructions used to build the underlying similarity matrix.
\end{comment}

\section{Experimental Setup}
To evaluate \methodprop{}, we fine-tune recent 7B--9B models on mixtures drawn from 309 instruction-tuning tasks under varying training budgets. We evaluate on benchmarks spanning reasoning, knowledge, mathematics, and instruction following. We ask two questions: (1) does a single, benchmark-independent \methodprop{} mixture outperform heuristic and existing selection methods (LESS and SMART)? and (2) what structure does the learned mixture exhibit, including task selection, pairwise effects, and stability as tasks are added?

  \textbf{Models} We evaluate \methodprop{} on three decoder language models:
\textit{Qwen-2-7B}~\citep{yang2024qwen2technicalreport}, \textit{Llama-2-7B}~\citep{touvron2023llama} and \textit{Qwen 3.5 9B Base} \citep{qwen3.5}.
Fine-tuning hyperparameters are reported in \Cref{app:finetune}.
\textbf{Run variability.} For each model and budget $\mathcal{B}$, we fine-tune three times with different random seeds and report the mean.

\textbf{Instruction Corpora.} We evaluate our framework on a diverse set of
  instruction tuning datasets spanning language understanding, reasoning, and
knowledge. These include: \textit{\textsc{Flan~2021}}~\citep{longpre2023flan,chung2022scaling} (${\sim}$17.5M examples), \textit{\textsc{T0}}~\citep{sanh2021multitask},
\textit{\textsc{Chain-of-Thought (CoT)}}~\citep{wei2022chain},
\textit{\textsc{Tulu~V3}}~\citep{lambert2024t,wang2023super}, and
\textit{\textsc{GLUE/SuperGLUE}}~\citep{wang2018glue,wang2019superglue}.
Together these span 309 tasks (listed in \Cref{sec:datasets_tasks}), providing a rich and heterogeneous
testbed for data mixture selection.

  \textbf{Mixture hyperparameters.} We compute a single mixture once from Qwen2-7B single-task models with $\rho=0.02424$ ($n\rho=7.49$) and $\tau=\tau_0+10^{-3}$, and reuse it for all models and budgets; thus Llama-2-7B and Qwen3.5-9B results test transfer across families/sizes (\Cref{app:affinity-impl}). This mixture has $206$ active tasks and effective size $e^{H(\vp^\ast)}=179.6$ (\Cref{tab:pairwise}). We fix $\rho$ a priori to target $\sim\!2/3$ coverage; \Cref{sec:b2} gives the full $\rho$ path, and \Cref{app:algorithm,app:sampling} describe mixture computation and sampling.

\paragraph{Ablation: spectrum shift and offset effects.}
Motivated by our convexity-on-the-simplex analysis (\Cref{para:critical-shift}), we ablate the spectrum shift $\tau$ in $\bS+\tau\mathbf I$ (\Cref{fig:shift}). For PMI affinity, $\tau_c=17.09$ whereas $\tau_0=186.30$, an order-of-magnitude gap between simplex convexity and global PSD. As $\tau$ decreases from $\tau_0$, the learned mixture contracts smoothly (support $206\to67$ and effective size $e^{H(\vp^\ast)}$ drops $179.6\to44.6$ by $\tau_c^{+}$), while at $\tau=0$ projected gradient collapses to a single-task vertex, showing the shift is essential (\Cref{app:shift}).

\textbf{Baselines for Comparison.} To evaluate the efficacy of our framework, we compare against several standard heuristic and SoTA data selection strategies.

We compare our methodology against the following baselines, where $\mathcal{B}$ is the
total instance budget, $|\mathcal D_i|$ the size of task $\T_i$, and
$N=\sum_{j=1}^n|\mathcal D_j|$ the size of the pool (implementation details in \Cref{app:baselines}):
\begin{enumerate}
  \item \textit{\textsc{Uniform}}\label{itm:uniform}: equal weight per task,
        $\vp_{[i]}=1/n$, so each task contributes $\mathcal{B}/n$ instances, sampled
        uniformly within the task.
  \item \textit{\textsc{EPM}}\label{itm:epm}: weight proportional to task size,
        $\vp_{[i]}=|\mathcal D_i|/N$, so task $\T_i$ contributes
        $\mathcal{B}\,|\mathcal D_i|/N$ instances, sampled uniformly within the task.
  \item \textit{\textsc{Random}}\label{itm:random}: $\mathcal{B}$ instances drawn uniformly without
        replacement from the pooled set $\bigcup\mathcal D_i$; in
        expectation each task contributes $\mathcal{B}\,|\mathcal D_i|/N$ instances.
  \item \textit{\textsc{LESS}} \citep{xia2024less}: an influence-function-based
        approach that selects training instances by gradient similarity to a
        target benchmark.
  \item \textit{\textsc{SMarT}} \citep{renduchintala-etal-2024-smart}: uses
        submodular optimization to allocate the task budget and select
        diverse, non-redundant instances.
\end{enumerate}

\textbf{Evaluation Metrics.} For the performance of our models, we evaluate on
\emph{eleutherai/lm-evaluation-harness} \citep{sutawika2024eleutherai} on six Open LLM Leaderboard benchmarks: \textit{BBH},  \textit{GPQA}, \textit{IFEval}, \textit{Math}, \textit{MMLU-Pro} and \textit{MUSR}. The evaluation protocol is detailed in \Cref{app:evaluation}.

% ===== Table: Qwen-2-7b =====

\begin{table*}[t]
\centering
\tiny
\setlength{\tabcolsep}{1pt}
\renewcommand{\arraystretch}{1.05}
\caption{Instruct-tuning performance on Leaderboard benchmarks (25K and 50K samples).}

\begin{subtable}[t]{0.485\textwidth}
\centering
\caption{Qwen2-7B}
\label{tab:qwen-25k-50k}
\resizebox{\linewidth}{!}{%
\begin{tabular}{@{}l l >{\columncolor{grpR}}c >{\columncolor{grpR}}c >{\columncolor{grpK}}c >{\columncolor{grpK}}c >{\columncolor{grpM}}c >{\columncolor{grpI}}c c@{}}
\toprule
\multicolumn{1}{c}{\shortstack{Size\\($\mathcal{B}$)}} & \multicolumn{1}{c}{Method}
  & \multicolumn{2}{>{\columncolor{grpR}}c}{Reasoning} & \multicolumn{2}{>{\columncolor{grpK}}c}{Knowledge} & \multicolumn{1}{>{\columncolor{grpM}}c}{Math} & \multicolumn{1}{>{\columncolor{grpI}}c}{Instr.} & \\
\cmidrule(lr){3-4}\cmidrule(lr){5-6}\cmidrule(lr){7-7}\cmidrule(lr){8-8}
 &  & BBH & MuSR & GPQA & MMLU-Pro & MATH & IFEval & Avg. \\
\midrule

\multirow{6}{*}{25K}  & Random (\ref{itm:random})      & $49.09$ & $46.03$ & $31.88$ & $41.57$ & $18.81$ & $32.85$ & $36.71$ \\
  & Uniform (\ref{itm:uniform})     & $50.13$ & $46.83$ & $33.06$ & $41.61$ & $20.85$ & $29.26$ & $36.96$ \\
  & EPM (\ref{itm:epm})         & $49.70$ & $45.37$ & $33.14$ & $41.46$ & $20.68$ & $30.94$ & $36.88$ \\
  & LESS   \citep{xia2024less}     & $51.73$ & $35.98$ & $30.20$ & $42.30$ & $20.02$ & $32.97$ & $35.53$ \\
  & SMART \citep{renduchintala-etal-2024-smart}   & $50.92$ & $44.91$ & $31.96$ & $41.37$ & $17.82$ & $\mathbf{40.29}$ & $37.88$ \\
  & \cellcolor{black!6}\bfseries \methodprop{} (\ref{eq:PMI_matrix})  & \cellcolor{grpRd}$\mathbf{52.02}$ & \cellcolor{grpRd}$\mathbf{47.01}$ & \cellcolor{grpKd}$\mathbf{33.41}$ & \cellcolor{grpKd}$\mathbf{42.51}$ & \cellcolor{grpMd}$\mathbf{20.96}$ & \cellcolor{grpId}$\secondbest{39.09}$ & \cellcolor{black!6}$\mathbf{39.17}$ \\
 %  & \cellcolor{black!6}JSD (\ref{eq:JSD_matrix})  & \cellcolor{black!6}$50.96$ & \cellcolor{black!6}$45.94$ & \cellcolor{black!6}$31.80$ & \cellcolor{black!6}$41.98$ & \cellcolor{black!6}$\mathbf{21.15}$ & \cellcolor{black!6}$\mathbf{41.97}$ & \cellcolor{black!6} \\

\midrule

\multirow{6}{*}{50K}  & Random (\ref{itm:random})      & $51.28$ & $46.03$ & $32.63$ & $41.98$ & $19.18$ & $37.41$ & $38.09$ \\
  & Uniform (\ref{itm:uniform})     & $51.33$ & $46.16$ & $32.30$ & $42.00$ & $20.83$ & $34.29$ & $37.82$ \\
  & EPM (\ref{itm:epm})         & $51.17$ & $45.37$ & $32.72$ & $41.95$ & $20.02$ & $34.17$ & $37.57$ \\
  & LESS \citep{xia2024less}  & $51.36$ & $35.85$ & $29.70$ & $41.81$ & $19.26$ & $35.49$ & $35.58$ \\
   & SMART \citep{renduchintala-etal-2024-smart}  & $51.41$ & $43.78$ & $31.46$ & $41.89$ & $18.58$ & $\mathbf{39.87}$ & $37.83$ \\
  & \cellcolor{black!6}\bfseries \methodprop{} (\ref{eq:PMI_matrix})   & \cellcolor{grpRd}$\mathbf{52.17}$ & \cellcolor{grpRd}$\mathbf{47.30}$ & \cellcolor{grpKd}$\mathbf{32.80}$ & \cellcolor{grpKd}$\mathbf{42.17}$ & \cellcolor{grpMd}$\mathbf{21.83}$ & \cellcolor{grpId}$\secondbest{39.73}$ & \cellcolor{black!6}$\mathbf{39.33}$ \\
 %  & \cellcolor{black!6}JSD (\ref{eq:JSD_matrix})  & \cellcolor{black!6}$52.16$ & \cellcolor{black!6}$46.09$ & \cellcolor{black!6}$\mathbf{33.14}$ & \cellcolor{black!6}$41.91$ & \cellcolor{black!6}$19.96$ & \cellcolor{black!6}$\mathbf{47.00}$ & \cellcolor{black!6} \\
\bottomrule
\end{tabular}%
}
\end{subtable}
\hfill
\begin{subtable}[t]{0.485\textwidth}
\centering
\caption{Llama2-7B}
\label{tab:llama-25k-50k}
\resizebox{\linewidth}{!}{%
\begin{tabular}{@{}l l >{\columncolor{grpR}}c >{\columncolor{grpR}}c >{\columncolor{grpK}}c >{\columncolor{grpK}}c >{\columncolor{grpM}}c >{\columncolor{grpI}}c c@{}}
\toprule
\multicolumn{1}{c}{\shortstack{Size\\($\mathcal{B}$)}} & \multicolumn{1}{c}{Method}
  & \multicolumn{2}{>{\columncolor{grpR}}c}{Reasoning} & \multicolumn{2}{>{\columncolor{grpK}}c}{Knowledge} & \multicolumn{1}{>{\columncolor{grpM}}c}{Math} & \multicolumn{1}{>{\columncolor{grpI}}c}{Instr.} & \\
\cmidrule(lr){3-4}\cmidrule(lr){5-6}\cmidrule(lr){7-7}\cmidrule(lr){8-8}
 &  & BBH & MuSR & GPQA & MMLU-Pro & MATH & IFEval & Avg. \\
\midrule
\multirow{6}{*}{25K}  & Random (\ref{itm:random})      & $34.82$ & $36.77$ & $26.26$ & $18.77$ & $0.98$ & $34.65$ & $25.38$ \\
  & Uniform (\ref{itm:uniform})     & $35.01$ & $40.27$ & $27.01$ & $17.68$ & $1.51$ & $35.01$ & $26.08$ \\
  & EPM (\ref{itm:epm})         & $35.93$ & $42.86$ & $26.01$ & $18.36$ & $1.51$ & $34.05$ & $26.45$ \\
  & LESS \citep{xia2024less}       & $40.59$ & $41.57$ & $24.12$ & $18.16$ & $0.01$ & $34.17$ & $26.44$ \\
  & SMART \citep{renduchintala-etal-2024-smart}  & $36.30$ & $40.48$ & $25.34$ & $19.02$ & $1.35$ & $\mathbf{37.81}$ & $26.72$ \\
  & \cellcolor{black!6}\bfseries \methodprop{} (\ref{eq:PMI_matrix})  & \cellcolor{grpRd}$\mathbf{40.95}$ & \cellcolor{grpRd}$\mathbf{42.98}$ & \cellcolor{grpKd}$\mathbf{27.18}$ & \cellcolor{grpKd}$\mathbf{19.24}$ & \cellcolor{grpMd}$\mathbf{1.59}$ & \cellcolor{grpId}$\secondbest{35.61}$ & \cellcolor{black!6}$\mathbf{27.93}$ \\
 %  & \cellcolor{black!6}JSD (\ref{eq:JSD_matrix})  & \cellcolor{black!6}$34.54$ & \cellcolor{black!6}$40.21$ & \cellcolor{black!6}$\mathbf{27.85}$ & \cellcolor{black!6}$17.90$ & \cellcolor{black!6}$1.51$ & \cellcolor{black!6}$\mathbf{37.29}$ & \cellcolor{black!6} \\

\midrule
\multirow{6}{*}{50K}  & Random (\ref{itm:random})      & $35.65$ & $37.70$ & $26.68$ & $18.11$ & $1.34$ & $35.81$ & $25.88$ \\
  & Uniform (\ref{itm:uniform})     & $34.80$ & $40.57$ & $27.85$ & $18.95$ & $1.61$ & $\mathbf{39.01}$ & $27.13$ \\
  & EPM (\ref{itm:epm})         & $35.32$ & $42.06$ & $26.34$ & $17.99$ & $1.28$ & $35.07$ & $26.34$ \\
  & LESS \citep{xia2024less}       & $41.33$ & $42.26$ & $24.75$ & $19.06$ & $1.36$ & $34.07$ & $27.14$ \\
  & SMART \citep{renduchintala-etal-2024-smart}  & $36.71$ & $37.83$ & $25.92$ & $19.24$ & $1.36$ & $34.77$ & $25.97$ \\
  & \cellcolor{black!6}\bfseries \methodprop{} (\ref{eq:PMI_matrix}) & \cellcolor{grpRd}$\mathbf{41.59}$ & \cellcolor{grpRd}$\mathbf{45.11}$ & \cellcolor{grpKd}$\mathbf{27.94}$ & \cellcolor{grpKd}$\mathbf{19.67}$ & \cellcolor{grpMd}$\mathbf{1.89}$ & \cellcolor{grpId}$\secondbest{35.77}$ & \cellcolor{black!6}$\mathbf{28.66}$ \\
 %  & \cellcolor{black!6}JSD (\ref{eq:JSD_matrix})  & \cellcolor{black!6}$36.24$ & \cellcolor{black!6}$42.46$ & \cellcolor{black!6}$\mathbf{28.02}$ & \cellcolor{black!6}$19.27$ & \cellcolor{black!6}$0.98$ & \cellcolor{black!6}$\mathbf{39.25}$ & \cellcolor{black!6} \\
\bottomrule
\end{tabular}%
}
\end{subtable}
\end{table*}

% ===== Table: Qwen-3.5-9b =====
\begin{table*}[!t]
    \centering
    \tiny
    \setlength{\tabcolsep}{1pt}
    \renewcommand{\arraystretch}{1.05}
    \caption{Qwen3.5-9B}
    \label{tab:qwen3_9B-25k-50k}
    \begin{minipage}[t]{0.485\textwidth}
        \centering
        \resizebox{\linewidth}{!}{%
        \begin{tabular}{@{}l l >{\columncolor{grpR}}c >{\columncolor{grpR}}c >{\columncolor{grpK}}c >{\columncolor{grpK}}c >{\columncolor{grpM}}c >{\columncolor{grpI}}c c@{}}
            \toprule
            \multicolumn{1}{c}{\shortstack{Size\\($\mathcal{B}$)}} & \multicolumn{1}{c}{Method}
              & \multicolumn{2}{>{\columncolor{grpR}}c}{Reasoning} & \multicolumn{2}{>{\columncolor{grpK}}c}{Knowledge} & \multicolumn{1}{>{\columncolor{grpM}}c}{Math} & \multicolumn{1}{>{\columncolor{grpI}}c}{Instr.} & \\
            \cmidrule(lr){3-4}\cmidrule(lr){5-6}\cmidrule(lr){7-7}\cmidrule(lr){8-8}
            &  & BBH & MuSR & GPQA & MMLU-Pro & MATH & IFEval & Avg. \\
            \midrule

            \multirow{6}{*}{25K}
            & Random (\ref{itm:random})
            & $57.89$ & $38.76$ & $36.07$ & $31.10$ & $27.02$ & $46.16$ & $39.50$ \\

            & Uniform (\ref{itm:uniform})
            & $53.08$ & $39.02$ & $31.29$ & $23.74$ & $28.25$ & $45.08$ & $36.74$ \\

            & EPM (\ref{itm:epm})
            & $55.86$ & $39.42$ & $30.62$ & $25.65$ & $25.73$ & $44.48$ & $36.96$ \\

            & LESS \citep{xia2024less}
            & $61.62$ & $44.90$ & $36.58$ & $51.78$ & $29.68$ & $45.29$ & $44.98$ \\

            & SMART \citep{renduchintala-etal-2024-smart}
            & $40.37$ & $39.02$ & $29.11$ & $12.77$ & $27.04$ & $44.12$ & $32.07$ \\

            & \cellcolor{black!6}\bfseries \methodprop{} (\ref{eq:PMI_matrix})
            & \cellcolor{grpRd}$\mathbf{62.58}$
            & \cellcolor{grpRd}$\mathbf{46.43}$
            & \cellcolor{grpKd}$\mathbf{38.93}$
            & \cellcolor{grpKd}$\mathbf{54.91}$
            & \cellcolor{grpMd}$\mathbf{30.06}$
            & \cellcolor{grpId}$\mathbf{56.12}$
            & \cellcolor{black!6}$\mathbf{48.17}$ \\

            \midrule

            \multirow{6}{*}{50K}
            & Random (\ref{itm:random})
            & $59.75$ & $36.29$ & $37.02$ & $32.45$ & $26.66$ & $46.36$ & $39.76$ \\

            & Uniform (\ref{itm:uniform})
            & $60.33$ & $38.62$ & $32.80$ & $40.92$ & $27.49$ & $46.76$ & $41.15$ \\

            & EPM (\ref{itm:epm})
            & $62.00$ & $39.15$ & $36.33$ & $35.84$ & $26.36$ & $48.20$ & $41.31$ \\

            & LESS \citep{xia2024less}
            & $65.94$ & $47.75$ & $34.65$ & $52.83$ & $29.76$ & $45.47$ & $46.07$ \\

            & SMART \citep{renduchintala-etal-2024-smart}
            & $38.47$ & $38.23$ & $28.69$ & $12.72$ & $28.56$ & $46.64$ & $32.22$ \\

            & \cellcolor{black!6}\bfseries \methodprop{} (\ref{eq:PMI_matrix})
            & \cellcolor{grpRd}$\mathbf{67.36}$
            & \cellcolor{grpRd}$\mathbf{51.31}$
            & \cellcolor{grpKd}$\mathbf{41.55}$
            & \cellcolor{grpKd}$\mathbf{56.84}$
            & \cellcolor{grpMd}$\mathbf{32.74}$
            & \cellcolor{grpId}$\mathbf{60.16}$
            & \cellcolor{black!6}$\mathbf{51.66}$ \\

            \bottomrule
        \end{tabular}%
        }
    \end{minipage}
\end{table*}

\subsection{Observations}

\textbf{Overall performance.}
Across the $36$ model--budget--benchmark settings, \methodprop{} scores
highest in $32$. On BBH, GPQA, MATH, MMLU-Pro and MuSR it ranks first for
every model and budget, and it beats the best of Random, Uniform and EPM in
$35$ of $36$ settings. The margins are largest where reasoning matters most:
on Llama-2-7B, BBH rises from $35.93$ to $40.95$ at 25K and from $35.65$ to
$41.59$ at 50K over the strongest heuristic, and on Qwen3.5-9B-Base
\methodprop{} ranks first on all six benchmarks at both budgets. It also has the highest average score in all six model--budget
blocks, ahead of the next-best method by $1.21$ to $1.52$ points on Qwen2-7B
and Llama-2-7B and by $3.19$ (25K) and $5.59$ (50K) points on Qwen3.5-9B-Base.
Results at a $100$K budget on Llama-2-7B are in \Cref{app:more-results}.

\textbf{Gains by capability.}
\methodprop{} ranks first on every \tagR{}, \tagK{} and \tagM{} benchmark,
for all three models and both budgets ($30$ of $30$ settings). On Qwen2-7B
and Llama-2-7B the leads are small, mostly $0.08$ to $1.14$ points (the
largest is $2.85$ on Llama-2-7B MuSR at 50K), but they are consistent:
the runner-up changes from cell to cell among LESS, SMART, Uniform and EPM,
while \methodprop{} is never behind. On Qwen3.5-9B-Base the margins widen to
$0.96$--$3.56$ points on \tagR{}, $2.35$--$4.53$ on \tagK{} (e.g.\ MMLU-Pro
$56.84$ vs.\ $52.83$ for LESS at 50K) and $0.38$--$2.98$ on \tagM{}.

\begin{wrapfigure}{r}{0.60\columnwidth}
  \vspace{-0.75\baselineskip}
  \centering
  \includegraphics[scale=0.8]{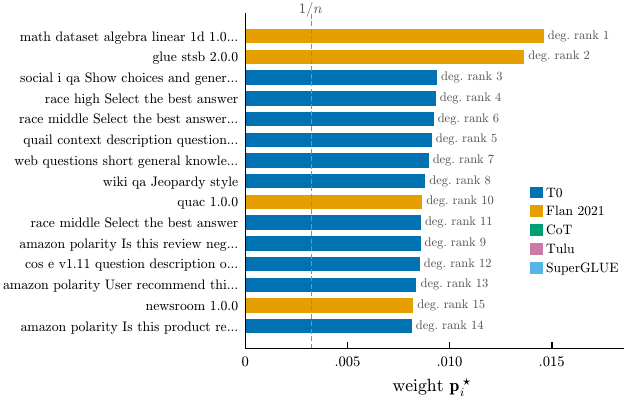}
  \caption{The $15$ highest-weighted tasks, coloured by collection, with each task's rank by degree $\bS_{[i]}$. The full structure of the mixture, including the affinity matrix reordered by weight, is analysed in \Cref{app:q2}.}
  \label{fig:q2}
  \vspace{-0.75\baselineskip}
\end{wrapfigure}

\textbf{\tagI{}.}
IFEval is the only benchmark where \methodprop{} does not always lead. On
Qwen2-7B and Llama-2-7B, SMART or Uniform is ahead by $0.14$ to $3.24$
points. On Qwen3.5-9B-Base the gap reverses: \methodprop{} leads the
strongest baseline by $9.96$ points at 25K and $11.96$ at 50K.

\textbf{Comparison with LESS.}
LESS \citep{xia2024less} selects data separately for each benchmark using gradient similarity,
and trains one checkpoint per benchmark. \methodprop{} trains a single
checkpoint without access to any benchmark data, yet outperforms LESS in all
$36$ settings. It is also cheaper: computing the affinity matrix takes
$\approx16$ GPU-hours on 8 H100s and is done once, after which
\Cref{alg:mixture} returns a mixture for any budget in seconds. LESS needs
$\approx320$ hours on 4 H100s to build its gradient datastore alone.

\textbf{Effect of training budget.}
Doubling the budget from 25K to 50K improves \methodprop{} in $16$ of $18$
model--benchmark pairs, with gains of up to $4.88$ points on
Qwen3.5-9B-Base (MuSR) and the only dips on Qwen2-7B GPQA and MMLU-Pro
($-0.61$ and $-0.34$). The baselines scale less reliably: SMART improves in
only $9$ of $18$ pairs, losing up to $3.04$ points on Llama-2-7B IFEval, and
EPM improves in $10$. A larger random or size-proportional sample can help
one benchmark while hurting another, because extra data is not necessarily
useful data. The selected mixture is also data-efficient: \methodprop{} at
25K matches or beats the best heuristic at 50K in $15$ of $18$ pairs. We also discuss in detail contamination/overlap audits to Appendix~\ref{app:potential-contamination}.

  \subsection{Effect and Stability of Spectral Shifting}
  \label{subsec:spectral-shift-ablation}
  The two spectral-shift experiments are reported with the method (\Cref{fig:shift,fig:incr}); we summarise their main takeaways here (details in \Cref{app:shift,app:incremental}).
 Lowering the shift from $\tau_0=186.30$ toward $\tau_c=17.09$ keeps the objective strictly convex on $\bDelta_n$ while shrinking the support from $206$ to $67$ tasks, with stable collection-level mass (T0: $0.62$--$0.65$; \Cref{fig:shift-mass}).

\section{Conclusion}
\noindent\methodprop{} optimizes fine-tuning task mixtures by coupling utility with an MRF-based diversity term built from behavior-derived affinities. Empirically, a single benchmark-independent mixture computed once from Qwen2-7B transfers reliably across model families (Qwen2-7B, Qwen3.5-9B, Llama2-7B) and across mixture budgets, yielding consistent gains on multi-step reasoning and instruction-following benchmarks. These results support the view that behavior-derived task affinities provide a practical inductive bias for data selection beyond semantic heuristics, while remaining compatible with standard sampling pipelines. Beyond performance, our analysis highlights how the spectrum shift controls support size and stabilizes optimization, offering an interpretable knob for trading off diversity and representativeness. Overall, \methodprop{} provides a simple, compute-efficient approach to constructing sparse mixtures that improve downstream generalization without requiring benchmark access during mixture design.

\section{Limitations}
\label{sec:limitations}
Our experiments cover $309$ instruction-tuning tasks and three base models in the $7$B--$9$B range with two instance budgets (25K/50K), demonstrating strong transfer across families and budgets despite this constrained setting. Importantly, \methodprop{} is modular: the same pipeline can be applied to larger task pools and model sizes as additional checkpoints and compute become available, and our analysis suggests the spectrum shift offers a simple mechanism to control sparsity and stability as scale increases. Affinity estimation relies on single-task checkpoints and held-out examples; however, this cost is incurred once and then amortized across mixture budgets, target models, and downstream deployments, making it practical when repeated fine-tuning runs are needed. Our evaluation centers on widely used leaderboard benchmarks to enable direct comparison with prior work; extending to additional axes (e.g., multilinguality, long-context, and safety) is a natural next step rather than a barrier to adoption. Finally, while hyperparameters such as $\rho$ and the shift level can be tuned, the method performs well with a fixed setting across all reported experiments, indicating robustness and reducing the need for benchmark-specific calibration.

  \addtocontents{toc}{\protect
  \setcounter{tocdepth}{2}} % Re-enable ToC entries

\section*{AI Use Statement}
We used generative AI tools to assist with copy-editing and \LaTeX{} editing. All scientific claims, experimental results, and conclusions were produced and verified by the authors. We reviewed and edited all AI-assisted content and take full responsibility for this submission. Every item below can be verified from the released code, with all paths given relative to the repository root. The repository includes the scripts, configuration files, and intermediate results needed to reproduce the experiments and regenerate the tables and figures reported in the paper.

\clearpage 

  % References

  \bibliographystyle{plainnat}
  \bibliography{ref}

  \clearpage

\section*{Checklist}

% %%% BEGIN INSTRUCTIONS %%%

% %%% END INSTRUCTIONS %%%

\begin{enumerate}

  \item For all models and algorithms presented, check if you include:
  \begin{enumerate}
    \item A clear description of the mathematical setting, assumptions, algorithm, and/or model. \textbf{Yes}
    \item An analysis of the properties and complexity (time, space, sample size) of any algorithm. \textbf{Yes}
    \item (Optional) Anonymized source code, with specification of all dependencies, including external libraries. \textbf{Yes}
  \end{enumerate}

  \item For any theoretical claim, check if you include:
  \begin{enumerate}
    \item Statements of the full set of assumptions of all theoretical results. \textbf{Yes}
    \item Complete proofs of all theoretical results. \textbf{Yes}
    \item Clear explanations of any assumptions. \textbf{Yes}
  \end{enumerate}

  \item For all figures and tables that present empirical results, check if you include:
  \begin{enumerate}
    \item The code, data, and instructions needed to reproduce the main experimental results (either in the supplemental material or as a URL). \textbf{Yes}
    \item All the training details (e.g., data splits, hyperparameters, how they were chosen). \textbf{Yes}
    \item A clear definition of the specific measure or statistics and error bars (e.g., with respect to the random seed after running experiments multiple times). \textbf{Yes}
    \item A description of the computing infrastructure used. (e.g., type of GPUs, internal cluster, or cloud provider). \textbf{Yes}
  \end{enumerate}

  \item If you are using existing assets (e.g., code, data, models) or curating/releasing new assets, check if you include:
  \begin{enumerate}
    \item Citations of the creator If your work uses existing assets. \textbf{Yes}
    \item The license information of the assets, if applicable. \textbf{Yes}
    \item New assets either in the supplemental material or as a URL, if applicable. \textbf{Not Applicable}
    \item Information about consent from data providers/curators. \textbf{Not Applicable}
    \item Discussion of sensible content if applicable, e.g., personally identifiable information or offensive content. \textbf{Not Applicable}
  \end{enumerate}

  \item If you used crowdsourcing or conducted research with human subjects, check if you include:
  \begin{enumerate}
    \item The full text of instructions given to participants and screenshots. \textbf{Not Applicable}
    \item Descriptions of potential participant risks, with links to Institutional Review Board (IRB) approvals if applicable. \textbf{Not Applicable}
    \item The estimated hourly wage paid to participants and the total amount spent on participant compensation. \textbf{Not Applicable}
  \end{enumerate}

\end{enumerate}

\clearpage

  % Supplement starts here (after references)
  \onecolumn \par
  \noindent

  \par
  \noindent
  \rule{\textwidth}{3pt}
  %rule{\textwidth}{1pt}
  \begin{center}
    \large\textbf{Supplementary Material: \papertitle}
  \end{center}
  \par
  \noindent
  \rule{\textwidth}{0.4pt}

  \appendix
  \addappheadtotoc

  % --- Tighter, cleaner ToC in the supplement (only if tocloft is available) ---
  \begingroup
  \makeatletter
  \@ifundefined{cftbeforesecskip}{}{%
    \setlength{\cftbeforesecskip}{0pt}%
    \setlength{\cftbeforesubsecskip}{0pt}%
    \setlength{\cftparskip}{0pt}%
    \setlength{\cftsecindent}{0pt}%
    \setlength{\cftsubsecindent}{1.2em}%
    \renewcommand{\cftsecfont}{\footnotesize}%
    \renewcommand{\cftsubsecfont}{\footnotesize}%
    \renewcommand{\cftsecpagefont}{\footnotesize}%
    \renewcommand{\cftsubsecpagefont}{\footnotesize}%
  }
  \makeatother
  \tableofcontents
  \endgroup

  \renewcommand{\thesection}{\Alph{section}} % Change section numbering to A, B, C...
  \renewcommand{\thesubsection}{\Alph{section}.\arabic{subsection}} % Change subsection numbering to A.1, A.2...

  % All appendix sections (theory, mixture design, experimental details, results,
  % further discussion) live in the supplement file.
  \input{supplemental.tex}

\end{document}

%% file: macros.tex
\def \papertitle{From Task Affinities to Optimal Allocations: Probabilistic Graphical Modelling for Supervised Finetuning
}

\newcommand{\captag}[2]{{\setlength{\fboxsep}{1.2pt}\colorbox{#1}{\strut #2}}}
\newcommand{\tagR}{\captag{grpRd}{Reasoning}}
\newcommand{\tagK}{\captag{grpKd}{Knowledge}}
\newcommand{\tagM}{\captag{grpMd}{Math}}
\newcommand{\tagI}{\captag{grpId}{Instruction following}}

\newcommand{\calA}{\mathcal{A}}
\newcommand{\calB}{\mathcal{B}}
\newcommand{\calF}{\mathcal{F}}
\newcommand{\calP}{\mathcal{P}}
\newcommand{\calQ}{\mathcal{Q}}

\newcommand{\norm}[1]{\left\bigl\lVert#1\right\bigr\rVert}

\newtheorem{proposition}{Proposition}

\AtBeginDocument{%
  \setlength{\abovedisplayskip}{4pt}%
  \setlength{\belowdisplayskip}{4pt}%
  \setlength{\abovedisplayshortskip}{3pt}%
  \setlength{\belowdisplayshortskip}{3pt}%
}

\newcommand{\method}[1]{CoreSetPFedBayes}

\newcommand{\mathcolorbox}[2]{%
  \begingroup\setlength{\fboxsep}{1pt}% padding
  \mathchoice{\colorbox{#1}{$\displaystyle #2$}}%
             {\colorbox{#1}{$\textstyle #2$}}%
             {\colorbox{#1}{$\scriptstyle #2$}}%
             {\colorbox{#1}{$\scriptscriptstyle #2$}}%
  \endgroup%
}

\newcommand{\secondbest}[1]{\uline{#1}}

\newcommand{\methodprop}{\TASKMIXPGM}
\newcommand{\TASKMIXPGM}{\ensuremath{\text{\scshape TaskMixPGM}}}

\makeatletter
\@ifundefined{newtheoremstyle}{%
  \newtheorem*{remark}{Remark}
}{%
  \newtheoremstyle{remarknodot}% name
    {3pt}{3pt}% Space above/below
    {\itshape}% Body font
    {}% Indent
    {\bfseries}% Head font
    {}% Punctuation after head (empty = no period)
    {0.5em}% Space after head
    {#1\ #2#3}% Head spec
  \theoremstyle{remarknodot}
  \newtheorem*{remark}{Remark}
  \theoremstyle{plain}% restore default for theorems/lemmas
}
\makeatother

\makeatletter
\AtBeginDocument{%
  \@ifundefined{thm@space@setup}{}{%
    \def\thm@space@setup{%
      \thm@preskip=3pt%
      \thm@postskip=3pt%
    }%
  }%
}
\makeatother

\newtheorem{thm}{Theorem}

\newtheorem{rem}{Remark}

\DeclareMathAlphabet{\pazocal}{OMS}{zplm}{m}{n}

\def\BibTeX{{\rm B\kern-.05em{\sc i\kern-.025em b}\kern-.08em
    T\kern-.1667em\lower.7ex\hbox{E}\kern-.125emX}}

\newcommand{\inner}[1]{\left\langle#1\right\rangle}

\def\A{\mathcal{A}}

\def\RR{\mathbb{R}}
\def\S{\mathcal{S}}
\def\PP{\mathbb{P}}

\def\D{\mathcal{D}}

\def\L{\mathcal{L}}

\def\supp{\mathop{\rm supp}\nolimits}

\def\bPsi{\boldsymbol{\Psi}}

\def\btheta{{\boldsymbol{\theta}}}
\def\PP{\mathbb{P}}
\def\bDelta{{\boldsymbol{\Delta}}}

\def \bp {{\mathbf{p}}}

\def\bx{{\mathbf x}}
\def\by{{\mathbf y}}

\def\calP{\mathcal P}

\def\calS{\mathcal S}

\def\calB{\mathcal B}

\def\bS{\boldsymbol S}

\def\bmu{{\boldsymbol \mu}}
\def\bnu{{\boldsymbol \nu}}

\def\argmax{\mathop{\rm arg\,max}\limits}%    a math operator.
\def\argmin{\mathop{\rm arg\,min}\limits}%    a math operator.

\def\maxop{\mathop{\rm max}\limits} %max operator
\def\minop{\mathop{\rm min}\limits}

\newcommand{\eat}[1]{}

\def\D{\mathcal{D}}

\definecolor{AccentBlue}{RGB}{42,82,190}
\definecolor{AccentOrange}{RGB}{234,106,32}
\colorlet{EqSeven}{AccentBlue}
\colorlet{EqEight}{AccentOrange}

\def\T{\mathcal{T}}
\def\V{\mathcal{V}}

\tcolorboxenvironment{exmp}{
  enhanced,
  colback=gray!10,                 % inner background color
  colframe=black,                  % inner border color
  boxrule=1pt,                   % inner border thickness
  borderline={1pt}{2pt}{black}, % outer border: {thickness}{offset}{color}
  sharp corners,
  boxsep=2pt,                      % padding inside the box
  left=2pt, right=2pt, top=2pt, bottom=2pt,
  before skip=\topsep,
  after skip=\topsep,
}

%% file: settings.tex
\usepackage{amsmath,amsthm,amssymb,amsxtra,mathtools,bm}
\usepackage{array}
\usepackage{graphicx}
\usepackage{algpseudocode}
\usepackage{tikzsymbols}
\usepackage{array,tabularx,multirow}
\usepackage{makecell}
\usepackage{array}

\usepackage{array}
\newcolumntype{P}[1]{>{\centering\arraybackslash}p{#1}}

\usepackage{verbatim}
\usepackage{enumerate}
\usepackage{parskip}

\usepackage{color,soul}
\definecolor{clemson-orange}{RGB}{234,106,32}
\definecolor{highlight-orange}{RGB}{255,150,150}
\definecolor{chicago-maroon}{RGB}{128,0,0}
\definecolor{cincinnati-red}{RGB}{190,0,0}
\definecolor{soft-cyan}{RGB}{68,85,90}
\definecolor{firebrick}{RGB}{178,34,34}
\definecolor{crimson}{RGB}{220,20,60}
\definecolor{cerrulean}{rgb}{0.165,0.322,0.745}
\definecolor{jaam}{rgb}{0.45,0.0,0.45}

\makeatletter
\@ifpackageloaded{hyperref}{%
  \hypersetup{%
    colorlinks = true,
    citecolor  = mahogany,
    linkcolor  = forestgreenweb,
    urlcolor   = forestgreenweb,
  }%
}{}
\makeatother

\usepackage{parskip}

\usepackage{thmtools}
\usepackage{thm-restate}
\newif\ifsolutions \solutionstrue

\def\final{0}
\newcommand{\reviewer}[3]{
  \expandafter\newcommand\csname #1\endcsname[1]{
    \ifthenelse{\equal{\final}{1}} {
      \textcolor{#3}{}
    } {
      \textcolor{#3}{\begin{center} \textbf{#2} ##1 \end{center}}
    }
  }
}

\reviewer{anirbit}{}{jaam}

\renewcommand{\norm}[1]{\left\| {#1} \right\|}

\def\1{\bm{1}}

\def\rq{{\textnormal{q}}}

\def\vp{{\bm{p}}}
\def\vq{{\bm{q}}}

\def\vx{{\bm{x}}}

\def\mX{{\bm{X}}}

\DeclareMathAlphabet{\mathsfit}{\encodingdefault}{\sfdefault}{m}{sl}
\SetMathAlphabet{\mathsfit}{bold}{\encodingdefault}{\sfdefault}{bx}{n}

%% file: supplemental.tex
\allowdisplaybreaks

\newpage

  \par
  \noindent
  \rule{\textwidth}{3pt}
  %rule{\textwidth}{1pt}
  \begin{center}
    \large\textbf{Supplementary Material: \papertitle}
  \end{center}
  \par
  \noindent
  \rule{\textwidth}{0.4pt}

\section{Organization of the Appendix}
The appendix is organized in five parts.
\begin{itemize}
  \item \textbf{Theory.} \Cref{app:theory} derives the KKT conditions and the
  interior closed form, and proves monotonicity, restricted strong concavity
  and weak submodularity of the support-restricted set function, with the
  greedy guarantee and a numerical and empirical check (\Cref{app:weak-submod-exp}).
  \item \textbf{Mixture design.} \Cref{app:mixture-design} explains how the
  mixture used in every experiment is obtained: why both potentials share one
  affinity matrix (\Cref{app:affinity-geometry-details}), the projection view of
  the objective (\Cref{sec:mixture}), how much diagonal shift is needed and why
  $\tau_c\ll\tau_0$ (\Cref{app:shift}), the trade-off ratio $\rho$ as a coverage
  control and the operating point (\Cref{sec:b2}), the solver and its
  verification (\Cref{app:algorithm}), and how the mixture is turned into a
  training set (\Cref{app:sampling}).
  \item \textbf{Experimental details.} \Cref{section:Experimental Details}
  describes how the affinity matrix was computed and what it costs
  (\Cref{app:affinity-impl}), the final fine-tuning runs
  (\Cref{app:finetune}), the baselines (\Cref{app:baselines}) and the evaluation
  protocol (\Cref{app:evaluation}).
  \item \textbf{Additional results and analysis.} \Cref{section:Additional Results}
  analyses the structure of the learned mixture (\Cref{section:mixture_analysis}),
  the role of the pairwise term (\Cref{app:pairwise}) and incremental
  composition (\Cref{app:incremental}), and reports additional fine-tuning
  results (\Cref{app:more-results}).
  \item \textbf{Further discussion.} \Cref{section:Comparative Analysis across various Notions of Task Similarities}
  compares notions of task similarity, \Cref{app:potential-contamination}
  discusses benchmark contamination, \Cref{app:additional-related-work}
  covers additional related work, \Cref{BroadImpact} the broader impact,
  \Cref{section:code} links to the code, and \Cref{sec:datasets_tasks} lists
  all 309 tasks.
\end{itemize}

% ############################################################################
%  PART I: THEORY (unchanged)
% ############################################################################

\newpage

\label{section:Monotonicity and Submodular Properties of Energy Potential}

\section{Theoretical Results and Proofs}
\label{app:theory}

\subsection{ Solution for Quadratic Minimization over the Simplex}

We consider the problem of minimizing a quadratic energy function over the probability
simplex $\bDelta_{n}$:
\begin{equation}
    \label{eq:objective}\min_{\bp \in \Delta_n}\;\; E(\bp) := -\bm{\Psi}_{\text{un}}
    ^{\top} \bp + \frac{1}{2}\bp^{\top} \bm{\Psi}^{(\succeq 0)}_{\mathrm{pair}}\bp
\end{equation}
where $\bm{\Psi}_{\texttt{un}}\in \mathbb{R}^{n}$ denotes a unary potential
vector and $\bm{\Psi}^{(\succeq 0)}_{\mathrm{pair}} \in \mathbb{R}^{n \times n}$ is a symmetric
positive semi-definite (PSD) matrix encoding pairwise interactions. For further analysis of the proof, we use the notation of $\bm{\Psi}^{(\succeq 0)}_{\mathrm{pair}}$ and $\bm{\Psi}_{\mathrm{pair}}$ interchangeably.

\subsection{Lagrangian and First-Order Conditions}\label{Appendix Sec: Lagrangian}

To enforce the affine constraint $\bp^{\top} \mathbf{1}_{n} = 1$, and inequality
constraints $\bp \geq 0$, we consider the KKT conditions for optimality. Define
the Lagrangian:
\begin{equation}
    \label{eq:lagrangian}\mathcal{L}(\bp, \nu, \bm{\mu}) = -\bm{\Psi}_{\text{un}}
    ^{\top} \bp + \frac{1}{2}\bp^{\top} \bm{\Psi}_{\text{pair}}\bp + \nu(\bp^{\top}
    \mathbf{1}_{n} - 1) - \bm{\mu}^{\top} \bp
\end{equation}
with dual variables $\nu \in \mathbb{R}$ (equality) and $\bm{\mu}\in \mathbb{R}^{n}
_{+}$ (inequality).

The \textbf{KKT optimality conditions} are:
{\setlength{\abovedisplayskip}{3pt}%
\setlength{\belowdisplayskip}{3pt}%
\setlength{\abovedisplayshortskip}{2pt}%
\setlength{\belowdisplayshortskip}{2pt}%

\noindent\textbf{Stationarity.}
\begin{equation}
  \label{eq:kkt_stationarity}
  \nabla_{\bp}\mathcal{L}= -\bm{\Psi}_{\text{un}}+ \bm{\Psi}_{\text{pair}}\bp
  + \nu \mathbf{1}_{n} - \bm{\mu}= 0.
\end{equation}
\begin{align*}
  \frac{\partial \mathcal{L}}{\partial \bp_{[i]}}
  &= -\bm{\Psi}_{\text{un}[i]}+ \sum_{j=1}^{n} \bm{\Psi}_{\text{pair}[ij]}\bp_{[j]}+ \nu - \bmu_{[i]}= 0,\\
  \Rightarrow\quad
  \nu &= \bm{\Psi}_{\text{un}[i]}- \sum_{j=1}^{n} \bm{\Psi}_{\text{pair}[ij]}\,\bp_{[j]}+ \bmu_{[i]}.
\end{align*}

\noindent\textbf{Primal feasibility.}\; $0 \leq \bp_{[i]}\leq 1$ for $i\in[n]$ and $\sum_{i=1}^{n}\bp_{[i]}=1$.

\noindent\textbf{Dual feasibility.}\; $\nu\in\mathbb{R}$ and $\bm{\mu}_{[i]}\ge 0$.

\noindent\textbf{Complementary slackness.}\; $\bm{\mu}_{[i]}\bp_{[i]}=0$ for all $i\in[n]$.

\noindent\textbf{Coordinate-wise cases.}
\begin{itemize}
  \setlength\itemsep{1pt}
  \setlength\topsep{2pt}
  \setlength\parskip{0pt}
  \item (\textbf{Active}) If $\bp_{[i]}>0$, then $\bm{\mu}_{[i]}=0$ and
  $\nu = \bm{\Psi}_{\text{un}[i]}- \sum_{j=1}^{n} \bm{\Psi}_{\text{pair}[ij]}\, \bp_{[j]}$.
  \item (\textbf{Inactive}) If $\bp_{[i]}=0$, then $\bm{\mu}_{[i]}\ge 0$, i.e.\
  $\bm{\Psi}_{\text{un}[i]}- \sum_{j} \bm{\Psi}_{\text{pair}[ij]}\, \bp_{[j]}\le\nu$.
\end{itemize}
}

Thus every task with positive weight has the same \emph{net utility}
$\bm{\Psi}_{\text{un}[i]}-(\bm{\Psi}_{\text{pair}}\bp)_{[i]}=\nu$, and every excluded task has
net utility at most $\nu$. When $\bm{\Psi}_{\text{pair}}$ is positive definite on $\mathbf 1^\perp$
(i.e.\ $\tau>\tau_c$, Lemma~\ref{lem:cpsd}), the problem is strictly convex on $\Delta_n$,
so these conditions are also sufficient and identify the unique minimiser.
If $\mathcal A=\supp(\bp^\ast)$ is known, the active block solves the linear system
$\bm{\Psi}_{\text{pair},\mathcal A\mathcal A}\,\bp_{\mathcal A}=\bm{\Psi}_{\text{un},\mathcal A}-\nu\mathbf 1$,
$\mathbf 1^\top\bp_{\mathcal A}=1$, i.e.\ \eqref{eq:closed_form} with all quantities restricted to $\mathcal A$.

\subsection{Solution under Interior Assumption}\label{Appendix: closedform Interior}

We first consider the case where the solution lies in the relative interior of
the simplex; that is, $\bp^{\ast} > 0$ and hence $\bm{\mu}= \mathbf{0}$. Substituting
into \eqref{eq:kkt_stationarity}, we obtain:
\begin{equation}
    \bm{\Psi}_{\text{pair}}\bp = \bm{\Psi}_{\text{un}}- \nu \mathbf{1}_{n}
\end{equation}
Assuming $\bm{\Psi}_{\text{pair}}$ is invertible (positive definiteness is not needed for this step),
we may solve:
\begin{equation}
    \bp = \bm{\Psi}_{\text{pair}}^{-1}\bm{\Psi}_{\text{un}}- \nu \bm{\Psi}_{\text{pair}}
    ^{-1}\mathbf{1}_{n}
\end{equation}
Imposing the constraint $\bp^{\top} \mathbf{1}_{n} = 1$, we find:
\begin{align}
    \label{eq:constraint_balance}\mathbf{1}_{n}^{\top} \bp & = \mathbf{1}_{n}^{\top} \bm{\Psi}_{\text{pair}}^{-1}\bm{\Psi}_{\text{un}}- \nu \mathbf{1}_{n}^{\top} \bm{\Psi}_{\text{pair}}^{-1}\mathbf{1}_{n} = 1
\end{align}
Letting
\[
    a := \mathbf{1}_{n}^{\top} \bm{\Psi}_{\text{pair}}^{-1}\bm{\Psi}_{\text{un}},
    \quad b := \mathbf{1}_{n}^{\top} \bm{\Psi}_{\text{pair}}^{-1}\mathbf{1}_{n},
\]
we obtain $\nu = \frac{a - 1}{b}$.

Substituting back into the expression for $\bp$, we conclude:
\begin{equation}
    \label{eq:closed_form}\bp^\ast = \bm{\Psi}_{\text{pair}}^{-1} \bm{\Psi}_{\text{un}} - \frac{\mathbf{1}_{n}^{\top} \bm{\Psi}_{\text{pair}}^{-1}\bm{\Psi}_{\text{un}}- 1}{\mathbf{1}_{n}^{\top} \bm{\Psi}_{\text{pair}}^{-1}\mathbf{1}_{n}} \cdot \bm{\Psi}_{\text{pair}}^{-1} \mathbf{1}_n
\end{equation}

\subsection{Discussion}

The closed-form expression \eqref{eq:closed_form} satisfies the affine constraint
by construction, and requires $\bm{\Psi}_{\text{pair}}$ to be invertible. If $\bp^{\ast} \ge 0$
componentwise and $\bm{\Psi}_{\text{pair}}$ is positive definite on $\mathbf 1^\perp$ ($\tau>\tau_c$),
it satisfies all KKT conditions with $\bm\mu=\mathbf 0$ and is therefore the unique global minimizer.
Invertibility alone is not enough: if $\bm{\Psi}_{\text{pair}}$ is indefinite on $\mathbf 1^\perp$, the
stationary point can be a saddle.
Otherwise, if any coordinate is negative, the interior assumption fails, and active-set
refinement or projection onto the simplex is required. In practice, one may use projection-based
algorithms (e.g., conditional gradient, projected gradient descent) or
iteratively restrict to the support set of nonnegative entries and resolve
\eqref{eq:closed_form} over that face of the simplex.

\subsection{Monotonicity and Submodular Properties of Energy Potential}
\label{app:submod-proofs}

\paragraph{Setting and notation.}
Throughout this section $\mathbb{E}(\vp)=\bPsi_{\texttt{un}}^{\top}\vp-\tfrac12\vp^{\top}\bPsi_{\texttt{pair}}\vp$
with $\bPsi_{\texttt{pair}}$ symmetric; in the algorithm
$\bPsi_{\texttt{pair}}=\lambda\widetilde{\bS}=\lambda(\bS+\tau\mathbf I)$ and
$\bPsi_{\texttt{un}}=\beta\widetilde{\bS}\one$.
For a nonempty $\A\subseteq\V$ let
$\calF(\A):=\{\vp\in\bDelta_n:\supp(\vp)\subseteq\A\}$ (a face of the simplex),
$f(\A):=\max_{\vp\in\calF(\A)}\mathbb{E}(\vp)$, and let $\zeta^{(\A)}$ denote a maximiser.
$f(\emptyset)$ is undefined because $\calF(\emptyset)=\emptyset$; this is handled
in Corollary~\ref{cor:greedy-app}.
Since any two points of $\bDelta_n$ differ by a zero-sum vector, curvature only
matters along $\one^\perp$. For $k\ge2$ we therefore define the
\emph{zero-sum restricted} eigenvalues
\begin{equation}
  m_k:=\min_{\substack{\bm h\in\one^\perp\setminus\{0\}\\ \|\bm h\|_0\le k}}
        \frac{\bm h^\top\bPsi_{\texttt{pair}}\bm h}{\|\bm h\|_2^2},
  \qquad
  M_k:=\max_{\substack{\bm h\in\one^\perp\setminus\{0\}\\ \|\bm h\|_0\le k}}
        \frac{\bm h^\top\bPsi_{\texttt{pair}}\bm h}{\|\bm h\|_2^2}.
  \label{eq:zs-eigs}
\end{equation}
$m_k$ is non-increasing and $M_k$ non-decreasing in $k$.
With $\bm B$ an orthonormal basis of $\one^\perp$ and $\tau_c=-\lambda_{\min}(\bm B^\top\bS\bm B)$
as in Lemma~\ref{lem:cpsd},
\begin{equation}
  m_k\;\ge\;\lambda\,(\tau-\tau_c),\qquad
  M_k\;\le\;\lambda\,\big(\tau+\lambda_{\max}(\bm B^\top\bS\bm B)\big)
  \qquad\text{for all }k,
  \label{eq:zs-eig-bounds}
\end{equation}
because restricting the Rayleigh quotient to sparse vectors can only raise its
minimum and lower its maximum. In particular $m_k>0$ for every $k$ whenever
$\tau>\tau_c$, which is a much weaker requirement than $\widetilde{\bS}\succ0$
($\tau>\tau_0$).

\lemmMonotonicity*
\phantomsection\label{app:proof-monotone}
\begin{proof}
Let $\emptyset\neq\calP\subseteq\calQ$. Every $\vp$ with
$\supp(\vp)\subseteq\calP$ also has $\supp(\vp)\subseteq\calQ$, so
$\calF(\calP)\subseteq\calF(\calQ)$. Both faces are nonempty and compact and
$\mathbb{E}$ is continuous, so both maxima are attained, and maximising the
same function over a larger set cannot decrease the maximum:
$f(\calQ)=\max_{\calF(\calQ)}\mathbb{E}\ge\mathbb{E}(\zeta^{(\calP)})=f(\calP)$.
No concavity or definiteness assumption is needed.
\end{proof}

\lemmFiniteRscRsm*
\begin{proof}
Let $\vp,\vq\in\bDelta_n$ with $|\supp(\vp)\cup\supp(\vq)|\le k$ and set
$\bm h=\vp-\vq$. Then $\one^\top\bm h=1-1=0$ and $\|\bm h\|_0\le k$. Since
$\nabla\mathbb{E}(\vq)=\bPsi_{\texttt{un}}-\bPsi_{\texttt{pair}}\vq$, expanding the quadratic gives the exact identity
\[
  \mathbb{E}(\vp)-\mathbb{E}(\vq)-\nabla\mathbb{E}(\vq)^\top\bm h
  \;=\;-\tfrac12\,\bm h^\top\bPsi_{\texttt{pair}}\bm h .
\]
By the definition \eqref{eq:zs-eigs},
$m_k\|\bm h\|_2^2\le\bm h^\top\bPsi_{\texttt{pair}}\bm h\le M_k\|\bm h\|_2^2$, which is the
claim. Both constants are finite for any symmetric $\bPsi_{\texttt{pair}}$
($|m_k|,|M_k|\le\|\bPsi_{\texttt{pair}}\|_2$), and $m_k>0$ whenever
$\tau>\tau_c$ by \eqref{eq:zs-eig-bounds}.
\end{proof}

\begin{rem}[Sign convention]
Because $\mathbb{E}$ is \emph{concave}, the Bregman remainder is non-positive;
restricted strong \emph{concavity} and smoothness of $\mathbb{E}$ are
restricted strong convexity and smoothness of $-\mathbb{E}$, which is the form
used by \citep{elenberg2018restricted} and \citep{gurumoorthy2019efficient}.
\end{rem}

\subsection{Weak Submodularity of Set function $f$}
\label{app:weak-submod-proof}

We follow the template of \citep[Thm.~4.3]{gurumoorthy2019efficient} (ProtoDash),
itself an instance of \citep{elenberg2018restricted}: bound the joint gain
$f(\mathcal{L}\cup\S)-f(\mathcal{L})$ from above by restricted strong concavity, and each
single gain $f(\mathcal{L}\cup\{j\})-f(\mathcal{L})$ from below by restricted smoothness,
both in terms of the gradient at $\zeta^{(\mathcal{L})}$. ProtoDash optimises over the
nonnegative orthant, where a new element can be added with any weight
$\alpha\ge0$ while the others stay fixed, and where optimality of
$\zeta^{(\mathcal{L})}$ makes the gradient vanish on its support. On the simplex
neither holds, and the argument needs three changes:
(i)~adding $j$ must move mass away from $\mathcal{L}$, i.e.\ along
$\bm e_j-\zeta^{(\mathcal{L})}$ with step $\alpha\in[0,1]$;
(ii)~at $\zeta^{(\mathcal{L})}$ the gradient equals the Lagrange multiplier $\nu$ (not $0$) on
the support, so the relevant quantity is the \emph{gap}
$g_j:=\nabla_j\mathbb{E}(\zeta^{(\mathcal{L})})-\nu$ (the Frank--Wolfe gap of
direction $j$);
(iii)~curvature is only needed along $\one^\perp$, which yields
\eqref{eq:zs-eigs} instead of ordinary sparse eigenvalues.
The price of the step cap $\alpha\le1$ is a factor $2$ in the ratio.

\weaksubmodularity*

\begin{proof}
Write $\zeta:=\zeta^{(\mathcal{L})}$, $\bm\nabla:=\nabla\mathbb{E}(\zeta)$, and
$\nu:=\bm\nabla^\top\zeta$. Set $m:=m_{|\mathcal{L}|+|\S|}>0$ and
$M:=M_{|\mathcal{L}|+1}$. Note $M\ge m_2\ge m>0$, since $|\mathcal{L}|+1\ge2$.

\emph{Step 1 (optimality of $\zeta$ on the face $\calF(\mathcal{L})$).}
$\zeta$ maximises a differentiable function over the face, which is a
polytope, so the KKT conditions hold without further qualification:
there are $\nu'\in\mathbb R$ and $\bm\mu\ge0$ with
$\bm\nabla_{[i]}=\nu'-\bm\mu_{[i]}$ and $\bm\mu_{[i]}\zeta_{[i]}=0$ for
$i\in\mathcal{L}$. Multiplying by $\zeta$ and summing gives $\nu'=\nu$. Hence
\begin{equation}
  \bm\nabla_{[i]}=\nu\ \ \text{if } \zeta_{[i]}>0,
  \qquad
  \bm\nabla_{[i]}\le\nu\ \ \text{if } i\in\mathcal{L},\ \zeta_{[i]}=0 .
  \label{eq:face-kkt}
\end{equation}
For $j\in\S$ define the gap $g_j:=\bm\nabla_{[j]}-\nu$ and
$g_j^+:=\max\{g_j,0\}$, and write $\bm g_\S^+:=(g_j^+)_{j\in\S}$.

\emph{Step 2 (upper bound on the joint gain).}
Let $\vp\in\calF(\mathcal{L}\cup\S)$ and $\bm h:=\vp-\zeta$. Then $\one^\top\bm h=0$ and
$\supp(\bm h)\subseteq\mathcal{L}\cup\S$. Using $\one^\top\bm h=0$,
\[
  \bm\nabla^\top\bm h=\sum_i(\bm\nabla_{[i]}-\nu)\bm h_{[i]}
  =\underbrace{\sum_{i\in\mathcal{L}}(\bm\nabla_{[i]}-\nu)\bm h_{[i]}}_{\le\,0}
   +\sum_{j\in\S}g_j\,\vp_{[j]}
  \;\le\;\sum_{j\in\S}g_j^+\,\vp_{[j]} .
\]
The first sum is non-positive by \eqref{eq:face-kkt}: its terms vanish where
$\zeta_{[i]}>0$, and where $\zeta_{[i]}=0$ we have $\bm h_{[i]}=\vp_{[i]}\ge0$ and
$\bm\nabla_{[i]}-\nu\le0$. For $j\in\S$ we used $\zeta_{[j]}=0$, so
$\bm h_{[j]}=\vp_{[j]}\ge0$. The last sum is at most
$\|\bm g_\S^+\|_2\|\bm h\|_2$ (Cauchy--Schwarz) and also at most
$g_{\max}:=\max_{j\in\S}g_j^+$, since $\sum_{j\in\S}\vp_{[j]}\le1$.
Applying Lemma~\ref{lem:finite-rsc-rsm} with $k=|\mathcal{L}|+|\S|$:
\[
  \mathbb{E}(\vp)-\mathbb{E}(\zeta)
  \le\bm\nabla^\top\bm h-\tfrac m2\|\bm h\|_2^2
  \le\|\bm g_\S^+\|_2\,\|\bm h\|_2-\tfrac m2\|\bm h\|_2^2
  \le\frac{\|\bm g_\S^+\|_2^2}{2m},
\]
and also $\mathbb{E}(\vp)-\mathbb{E}(\zeta)\le\bm\nabla^\top\bm h\le g_{\max}$.
Maximising over $\vp$,
\begin{equation}
  f(\mathcal{L}\cup\S)-f(\mathcal{L})\;\le\;\min\Big\{\frac{\|\bm g_\S^+\|_2^2}{2m},\;g_{\max}\Big\}.
  \label{eq:joint-upper}
\end{equation}

\emph{Step 3 (lower bound on each single gain).}
Fix $j\in\S$ with $g_j>0$ and let $\bm d_j:=\bm e_j-\zeta$. For
$\alpha\in[0,1]$, $\zeta+\alpha\bm d_j\in\calF(\mathcal{L}\cup\{j\})$. Since
$\mathbb{E}$ is quadratic,
\[
  \mathbb{E}(\zeta+\alpha\bm d_j)-\mathbb{E}(\zeta)
  =\alpha\,\bm\nabla^\top\bm d_j-\tfrac{\alpha^2}{2}\,\bm d_j^\top\bPsi_{\texttt{pair}}\bm d_j
  \;\ge\;\alpha\,g_j-\tfrac{\alpha^2}{2}\,M D_j ,
\]
where we used $\bm\nabla^\top\bm d_j=\bm\nabla_{[j]}-\nu=g_j$, and
$\bm d_j\in\one^\perp$ with $\|\bm d_j\|_0\le|\mathcal{L}|+1$, so
$\bm d_j^\top\bPsi_{\texttt{pair}}\bm d_j\le M D_j$ with
$D_j:=\|\bm d_j\|_2^2=1+\|\zeta\|_2^2\in(1,2]$.
Optimising over $\alpha\in[0,1]$:
if $g_j\le MD_j$, take $\alpha=g_j/(MD_j)$ to get a gain of at least
$g_j^2/(2MD_j)\ge g_j^2/(4M)$;
if $g_j>MD_j$, take $\alpha=1$ to get a gain of at least $g_j-MD_j/2>g_j/2$.
Hence
\begin{equation}
  f(\mathcal{L}\cup\{j\})-f(\mathcal{L})\;\ge\;\phi_j:=\min\Big\{\frac{(g_j^+)^2}{4M},\;\frac{g_j^+}{2}\Big\}
  \quad\text{for every } j\in\S ,
  \label{eq:single-lower}
\end{equation}
which holds trivially (with $\phi_j=0$) when $g_j\le0$, by monotonicity.

\emph{Step 4 (combining).}
If $g_{\max}\le 2M$, then for every $j$,
$g_j^+/2\ge (g_j^+)^2/(4M)$, so
$\sum_{j\in\S}\phi_j\ge\|\bm g_\S^+\|_2^2/(4M)
\ge\frac{m}{2M}\cdot\frac{\|\bm g_\S^+\|_2^2}{2m}$.
If $g_{\max}>2M$, the maximising $j^\ast$ satisfies $g_{j^\ast}>2M\ge MD_{j^\ast}$, so
$\sum_{j\in\S}\phi_j\ge g_{\max}/2\ge\frac12 g_{\max}\ge\frac{m}{2M}g_{\max}$,
using $m\le M$. In both cases, by \eqref{eq:joint-upper},
\[
  \sum_{j\in\S}\big[f(\mathcal{L}\cup\{j\})-f(\mathcal{L})\big]
  \;\ge\;\frac{m_{|\mathcal{L}|+|\S|}}{2\,M_{|\mathcal{L}|+1}}\,\big[f(\mathcal{L}\cup\S)-f(\mathcal{L})\big],
\]
which is the claimed bound on $\gamma_{\mathcal{L},\S}$. The bound in terms of
$\tau$ follows from \eqref{eq:zs-eig-bounds}, after the factor $\lambda$
cancels.
\end{proof}

\begin{corollary}[Greedy guarantee]
\label{cor:greedy-app}
Let $\tau>\tau_c$ and let greedy start from $\A_1=\{a_1\}$ with
$a_1\in\argmax_j f(\{j\})$, then add $\A_{t+1}=\A_t\cup\{\argmax_{j}f(\A_t\cup\{j\})\}$.
Let $\mathrm{OPT}_{r}:=\max_{|\A|\le r,\ \A\neq\emptyset}f(\A)$ and
$\gamma_k:=m_{2k-1}/(2M_{k+1})$. Then for every $k\ge2$
\[
  f(\A_k)-f(\A_1)\;\ge\;\big(1-e^{-\gamma_k}\big)\,\big(\mathrm{OPT}_{k-1}-f(\A_1)\big).
\]
\end{corollary}
\begin{proof}
On $\V\setminus\{a_1\}$ define $g(\mathcal{T}):=f(\mathcal{T}\cup\{a_1\})-f(\{a_1\})$. Then
$g(\emptyset)=0$, and $g$ is monotone by Lemma~\ref{lem:monotonicity}.
Greedy on $f$ after step 1 is greedy on $g$.
For $\mathcal{T}\subseteq\A_k\setminus\{a_1\}$ and $|\S|\le k-1$, the ratio
$\gamma^{g}_{\mathcal{T},\S}=\gamma^{f}_{\mathcal{T}\cup\{a_1\},\S}$, and
Theorem~\ref{thm:weak-submodularity} bounds it below by $\gamma_k$, since
$|\mathcal{T}\cup\{a_1\}|+|\S|\le 2k-1$ and $|\mathcal{T}\cup\{a_1\}|+1\le k+1$.
The greedy bound for weakly submodular functions \citep{das2011submodular,elenberg2018restricted}
gives $g(\A_k\setminus\{a_1\})\ge(1-e^{-\gamma_k})\max_{|\mathcal{T}|\le k-1}g(\mathcal{T})$.
For any $\A^\ast$ with $|\A^\ast|\le k-1$, monotonicity gives
$\max_{|\mathcal{T}|\le k-1}g(\mathcal{T})\ge f(\A^\ast\cup\{a_1\})-f(\{a_1\})\ge f(\A^\ast)-f(\{a_1\})$.
\end{proof}

\begin{rem}[Relation to the plotted curvature ratio]
The empirical curvature curves report
$\lambda_{\min}\big((\bS+\tau\mathbf I)_{\A}\big)/\lambda_{\max}\big((\bS+\tau\mathbf I)_{\A}\big)$,
the ratio of \emph{unrestricted} principal-submatrix eigenvalues. Restricting
the Rayleigh quotient to $\one^\perp$ can only raise the minimum and lower the
maximum, so this ratio is a conservative proxy for the constants in
Theorem~\ref{thm:weak-submodularity} (up to the factor $2$). By
\eqref{eq:zs-eig-bounds} the restricted constants stay positive at every support
size for any $\tau>\tau_c$; a negative proxy therefore does not imply that the
guarantee fails.
\end{rem}
% ---- begin inlined: Appendix/weak_submodularity_experiment.tex ----
\subsection{Weak submodularity in numbers and an empirical check}
\label{app:weak-submod-exp}

\begin{figure*}[t]
  \centering
  \begin{subfigure}[t]{0.24\textwidth}\centering
    \includegraphics[width=\linewidth]{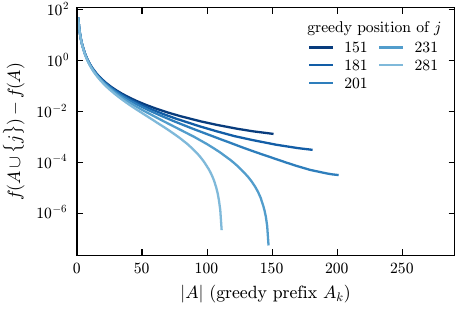}
    \caption{Diminishing returns.}\label{fig:submod-dim}
  \end{subfigure}\hfill
  \begin{subfigure}[t]{0.24\textwidth}\centering
    \includegraphics[width=\linewidth]{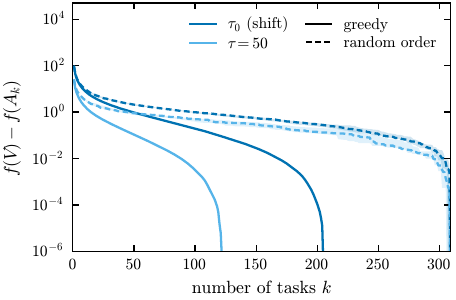}
    \caption{Greedy vs.\ random order.}\label{fig:submod-greedy}
  \end{subfigure}\hfill
  \begin{subfigure}[t]{0.24\textwidth}\centering
    \includegraphics[width=\linewidth]{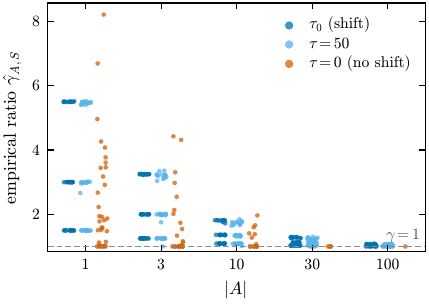}
    \caption{Empirical $\hat\gamma_{\A,\S}$.}\label{fig:submod-gamma}
  \end{subfigure}\hfill
  \begin{subfigure}[t]{0.24\textwidth}\centering
    \includegraphics[width=\linewidth]{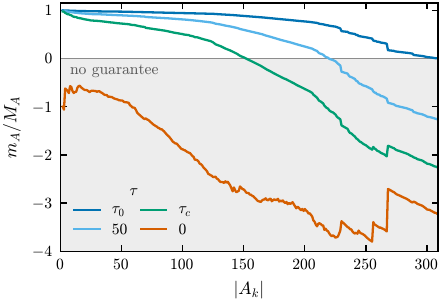}
    \caption{Curvature proxy $m_{\A}/M_{\A}$.}\label{fig:submod-curv}
  \end{subfigure}
  \caption{Weak submodularity of $f$. (a) Gain of five fixed tasks along
  greedy prefixes at $\tau_0$ (legend: greedy position of $j$). (b) Gap
  $f(\V)-f(\A_k)$ for exact greedy (solid) and random orders (dashed, band
  over $5$ random orders). (c) $\hat\gamma_{\A,\S}$ \eqref{eq:gamma_L_kappa} on
  the $180$ random-set pairs per $\tau$ ($12$ draws of $\A$ per size,
  $|\S|\in\{2,5,10\}$). Pairs with $f(\Bar{\A}\cup\Bar{\S})=f(\Bar{\A})$ are
  omitted ($0$, $1$ and $105$ for $\tau_0$, $50$, $0$). (d)
  $m_{\A}/M_{\A}$ for $(\bS+\tau\mathbf I)_{\A}$ along the $\tau_0$ greedy
  order, for $\tau\in\{\tau_0,50,\tau_c,0\}$, with $\tau_c=17.09$ the
  critical shift of \eqref{eq:tauc}. This principal-submatrix ratio is a
  conservative proxy for the zero-sum constants of Theorem~\ref{thm:weak-submodularity};
  below zero (shaded) the proxy gives no guarantee.}
  \label{fig:submod}
\end{figure*}

\textbf{The guarantee in numbers.}
With $\tau=\tau_0$, the zero-sum curvature constants of $\bS+\tau\mathbf I$
are $m=\tau_0-\tau_c=169.21$ and
$M=\lambda_{\max}(\bm U^\top\bS\bm U)+\tau_0=24.80+186.30=211.10$. The
Theorem~\ref{thm:weak-submodularity} then gives $\gamma\ge m/(2M)=0.40$, so
$k$-step greedy attains at least $1-e^{-0.40}\approx0.33$ of the best
$(k{-}1)$-task value. This is a worst-case bound over all supports.
Empirically $\hat\gamma\ge1$ on every sampled pair
(\Cref{fig:submod-gamma}).

\textbf{Empirical check of weak submodularity.}
We run exact greedy on $f$ (\Cref{eq:weak-submodularity}), re-solving
$f(\Bar{\A}\cup\{j\})$ for every candidate $j$ at each step, and compare it
with $5$ random insertion orders (seeds $0$--$4$). We estimate the
submodularity ratio \citep{das2011submodular} for
$\tau\in\{\tau_0,50,0\}$ on two families of disjoint pairs $(\A,\S)$
(\Cref{fig:submod-gamma}).
\emph{Greedy prefixes} ($880$ pairs): $\A$ is the first
$a\in\{1,2,5,10,25,50,100,150,200,250\}$ greedy tasks and
$|\S|\in\{2,5,10,25\}$, with $\S$ drawn uniformly $20$ times plus two
adversarial choices (the next $|\S|$ greedy tasks and the $|\S|$ tasks with
the largest individual gains). \emph{Random sets} ($180$ pairs): $12$ uniform
draws of $\A$ for each $|\A|\in\{1,3,10,30,100\}$, with
$|\S|\in\{2,5,10\}$. We discard pairs where $\S$ adds nothing to $\A$
(denominator of \eqref{eq:gamma_L_kappa} below $10^{-10}$). At $\tau=0$,
$f$ is non-convex, so we take the best of the warm start, four Dirichlet
restarts and every single-task vertex of $\Bar{\A}$; its $\hat\gamma$ is
therefore approximate.

\emph{Findings.}
(i) Greedy gains are non-increasing at all $307$ steps.
(ii) Greedy reaches $91\%$ of the total gain by $k=10$, $98.9\%$ by $k=50$
and $f(\V)$ at $k\approx206$; random orders reach $f(\V)$ only near $k=n$.
(iii) No retained pair has $\hat\gamma<1$. At $\tau_0$, $742/880$ and
$180/180$ pairs are retained (median $\hat\gamma$ $1.30$ and $1.28$); at
$\tau=50$, $577/880$ and $179/180$ (medians $1.36$, $1.26$); at $\tau=0$,
$0/880$ and $75/180$ (median $1.34$). (iv) The conservative proxy
$\lambda_{\min}/\lambda_{\max}$ of $(\bS+\tau\mathbf I)_{\A}$, the
principal submatrix of $\bS+\tau\mathbf I$ with rows and columns indexed by
$\A$ (\Cref{fig:submod-curv}), stays above $0.44$ for $|\A|\le250$ at
$\tau_0$, falls to $4\times10^{-6}$ at $|\A|=n$, and is negative throughout
at $\tau=0$. The exact zero-sum constants, however, stay positive at every
support size whenever $\tau>\tau_c$, so the guarantee covers the whole
greedy path at $\tau_0$ and $\tau=50$. At $\tau=0<\tau_c$ there is no
guarantee: $\hat\gamma\ge1$ holds only because $f$ collapses to the best
single task in $\A$, and greedy returns that one task. The shift is what
makes the selected set non-degenerate.

% ---- end inlined: Appendix/weak_submodularity_experiment.tex ----

% ############################################################################
%  PART II: MIXTURE DESIGN (the algorithm and operating point used for all results)
%  Mixture file: p_star_affinity.npy, rho = 0.02424, tau = tau_0 + 1e-3,
%  Qwen2-7B PMI affinity S (n = 309).
% ############################################################################

\newpage

\section{Mixture Design: Spectrum Shift, Convexity and the Operating Point}
\label{app:mixture-design}

This section explains, step by step, how the single mixture $\vp^\ast$ used
in every fine-tuning experiment is obtained from the affinity matrix $\bS$.
\Cref{app:affinity-geometry-details} motivates the objective,
\Cref{sec:mixture} rewrites it as a projection, \Cref{app:shift} explains why a
diagonal shift is needed and how large it must be, \Cref{sec:b2} explains how
the trade-off ratio $\rho$ controls coverage and fixes the operating point,
\Cref{app:algorithm} describes and verifies the solver, and
\Cref{app:sampling} turns the mixture into a training set.
Throughout, $\bm d:=\bS\one$ denotes the vector of task degrees
($\bm d_{[i]}=\bS_{[i]}$ in the main text).

\subsection{Design rationale}
\label{app:affinity-geometry-details}

\paragraph{Why both potentials derive from the same affinity graph.}
Deriving representativeness and redundancy from a single similarity
matrix is deliberate and follows a well-established design in subset
selection. Graph-centrality summarization scores sentences by their
degree in a similarity graph, and coverage--diversity objectives for
extractive summarization and data subset selection likewise compute both
the coverage reward and the redundancy penalty from one similarity kernel.
Three considerations motivate this choice here.
(i) \emph{Consistency}: if representativeness and redundancy were measured
with different notions of similarity, a task could be judged central under
one and non-redundant under the other for incompatible reasons; a single
behavioral $\bm S$ guarantees the tradeoff is internally coherent.
(ii) \emph{No target leakage}: every quantity is computed from the training
pool alone, so the resulting mixture is benchmark-independent by
construction, which is precisely what enables a single general-purpose
checkpoint.
(iii) \emph{Interpretability}: as we show next, sharing $\bm S$ yields a
closed-form characterization with a single, meaningful control knob.

\paragraph{Reading the affinity.}
$\bS_{[ij]}$ is the symmetrised average log-likelihood ratio between the
single-task models of $\T_i$ and $\T_j$, each evaluated on the other's
held-out data (\Cref{eq:PMI_matrix}). It is pointwise mutual information in
the sense of \citep{church1990word}, between the task identity and the observed
response, not mutual information between two random variables. Because each
model fits its own task best, $95\%$ of the off-diagonal entries are negative
(mean $-0.43$), and every degree $\bm d_{[i]}$ is negative, ranging from
$-1241$ to $-3.4$. A task is therefore \emph{representative} when its degree is
\emph{least negative}: the other tasks' models predict its data almost as well
as its own model does. By the offset invariance below, only differences
between degrees matter.

\paragraph{Restoring self-affinity.}
Both scores in \eqref{eq:PMI_matrix} are defined between \emph{distinct} tasks, so
we set $S_{[ii]}=0$. A hollow matrix has zero trace and is therefore
indefinite (unless $\bm S=\bm 0$), which makes the quadratic in \eqref{Eq: original opt}
nonconvex. This is an artifact of leaving the diagonal empty rather than a
property of the tasks: every task is, trivially, maximally similar to
itself. We therefore restore a uniform self-affinity $\widetilde{\bm S}=\bm S+\tau\bm I$
as in \eqref{eq:selfaff}. This is the spectrum shift of \citep{chen2009learning,wu2005analysis},
now read as filling the missing diagonal. Replacing $\bm S$ by $\widetilde{\bm S}$ in
\emph{both} potentials changes $\Psi_{\mathrm{un}}$ only by the constant
vector $\beta\tau\mathbf 1_n$, which is constant on $\Delta_n$ and hence
does not affect the minimizer. The net effect is the pairwise term
$\tfrac{\lambda\tau}{2}\|\bm p\|_2^2$, a preference for spreading
mass that is consistent with the diversity goal. Positive definiteness on all of
$\mathbb R^n$ ($\tau>\tau_0$) makes Proposition~\ref{prop:projection} a projection
in a norm, while convexity on $\Delta_n$ needs only $\tau>\tau_c$ (Lemma~\ref{lem:cpsd}).
Without any diagonal the problem is non-convex on $\Delta_n$ and its optimum
collapses onto a single task (\Cref{fig:shift}).

\paragraph{A single general-purpose checkpoint vs.\ per-target selection.}
Per-target methods such as LESS select a different subset, and train a
different model, for each benchmark using that benchmark's data.
\methodprop{} produces one mixture and one checkpoint evaluated on every
benchmark. The affinities are computed once per task pool and reused for every
budget and base model (\Cref{app:affinity-impl}), whereas a gradient datastore
must be rebuilt per base model and the selection repeated per benchmark.

\subsection{Mixture design as an affinity-geometry projection}
\label{sec:mixture}

\paragraph{Objective.}
Let $\bm S\in\mathbb R^{n\times n}$ be the symmetric, hollow PMI task-affinity
matrix and $\Delta_n=\{\bm p\in\mathbb R^n_{\ge0}:\mathbf 1^\top\bm p=1\}$.
With $\bPsi_{\texttt{un}}=\beta\widetilde{\bS}\one$ and
$\bPsi_{\texttt{pair}}=\lambda\widetilde{\bS}$, \Cref{Eq: original opt} reads
\begin{equation}
  \label{eq:obj}
  \bm p^\star=\arg\min_{\bm p\in\Delta_n}\;
  -\beta\,\mathbf 1^\top\widetilde{\bm S}\bm p
  +\tfrac{\lambda}{2}\,\bm p^\top\widetilde{\bm S}\bm p ,
\end{equation}
\begin{equation}
  \label{eq:selfaff}
  \widetilde{\bm S}=\bm S+\tau\bm I .
\end{equation}
Two observations fix the meaning of $\tau$. First, on $\Delta_n$ the unary
term differs from $-\beta\,\mathbf 1^\top\bm S\bm p$ only by the constant
$\beta\tau$, so the diagonal does not alter representativeness. Second,
$\bm p^\top\widetilde{\bm S}\bm p=\bm p^\top\bm S\bm p+\tau\lVert\bm p\rVert_2^2$,
and $\lVert\bm p\rVert_2^2$ is the collision probability of $\bm p$ (the
inverse of its order-2 effective size), so $\tau$ acts as a quadratic-entropy
(diversity) regulariser on the mixture.

\begin{proposition}[Mixture design as affinity-geometry projection]
\label{prop:projection}
Let $\widetilde{\bm S}\succ 0$, $\rho:=\beta/\lambda$,
and $\|\bm x\|_{\widetilde{\bm S}}^2 := \bm x^\top\widetilde{\bm S}\bm x$.
Then, for all $\bm p\in\mathbb R^n$,
\[
  -\beta\,\mathbf 1^\top\widetilde{\bm S}\bm p
  + \tfrac{\lambda}{2}\,\bm p^\top\widetilde{\bm S}\bm p
  \;=\;
  \tfrac{\lambda}{2}\,\big\|\bm p-\rho\mathbf 1_n\big\|_{\widetilde{\bm S}}^2
  \;-\; \tfrac{\beta^2}{2\lambda}\,\mathbf 1^\top\widetilde{\bm S}\mathbf 1 ,
\]
so the optimal mixture is the unique $\widetilde{\bm S}$-projection of the
scaled uniform allocation $\rho\mathbf 1_n$ onto the simplex:
\[
  \bm p^\star \;=\; \arg\min_{\bm p\in\Delta_n}\,
  \big\|\bm p-\rho\mathbf 1_n\big\|_{\widetilde{\bm S}} .
\]
\end{proposition}
\begin{proof}
Expand the right-hand side using the symmetry of $\widetilde{\bm S}$ and
$\lambda\rho=\beta$; the constant does not depend on $\bm p$.
Since $\widetilde{\bm S}\succ0$, $\|\cdot\|^2_{\widetilde{\bm S}}$ is strictly
convex and $\Delta_n$ is compact and convex, so the minimiser exists and is
unique.
\end{proof}

\begin{corollary}[Uniform $\leftrightarrow$ minimum-redundancy interpolation]
\label{cor:interp}
Let $\bm u=\tfrac1n\mathbf 1_n$ be the uniform mixture and
\[
  \bm q \;:=\; \arg\min_{\bm p:\,\mathbf 1^\top\bm p=1}\bm p^\top\widetilde{\bm S}\bm p
  \;=\; \frac{\widetilde{\bm S}^{-1}\mathbf 1_n}{\mathbf 1^\top\widetilde{\bm S}^{-1}\mathbf 1_n}
\]
be the minimum-redundancy mixture. The stationary point of
\eqref{eq:obj} under the equality constraint alone is
\[
  \tilde{\bm p} \;=\; n\rho\,\bm u \;-\; (n\rho-1)\,\bm q .
\]
Hence $\rho=1/n$ returns the uniform mixture
($\tilde{\bm p}=\bm u\in\Delta_n$, so $\bm p^\star=\bm u$);
$\rho<1/n$ gives a convex combination of $\bm u$ and $\bm q$ (diversity);
and $\rho>1/n$ extrapolates from $\bm u$ away from $\bm q$
(representativeness). The nonnegativity constraint then determines the
support of $\bm p^\star$.
\end{corollary}
\begin{proof}
Stationarity gives $\lambda\widetilde{\bm S}\bm p=\beta\widetilde{\bm S}\mathbf 1-\nu'\mathbf 1$,
i.e.\ $\bm p=\rho\mathbf 1-(\nu'/\lambda)\widetilde{\bm S}^{-1}\mathbf 1$.
Imposing $\mathbf 1^\top\bm p=1$ gives
$\nu'/\lambda=(n\rho-1)/(\mathbf 1^\top\widetilde{\bm S}^{-1}\mathbf 1)$,
and substituting yields the claim.
\end{proof}

\begin{remark}[What $\bm q$ down-weights]
\label{rem:degree}
In general $\bm q$ need not be ordered by task degree $\bm d$. When the
diagonal dominates ($\lVert\bm S\rVert_2<\tau$), the Neumann series gives
$\widetilde{\bm S}^{-1}\mathbf 1=\tau^{-1}\big(\mathbf 1-\tau^{-1}\bm d\big)
+O(\lVert\bm S\rVert_2^2/\tau^3)$, so to first order $\bm q$ down-weights
high-degree tasks and $\tilde{\bm p}$ for $\rho>1/n$ up-weights them.
At the operating point ($\tau=\tau_0+10^{-3}$, $n\rho=7.49$) the ordering is
nearly exact: Spearman$(\bm q,\bm d)=-0.994$ and
Spearman$(\tilde{\bm p},\bm d)=0.994$.
\end{remark}

\begin{remark}[Connection to mean--variance portfolio design]
\label{rem:portfolio}
Reading $\widetilde{\bm S}$ as a covariance, $\bm q$ is the global
minimum-variance portfolio and Corollary~\ref{cor:interp} is the two-fund
separation theorem: every equality-constrained solution is an affine
combination of the two funds $\bm u$ and $\bm q$. The simplex constraint is the usual
long-only (no short-selling) restriction, and $\rho$ plays the role of the
return--risk trade-off.
\end{remark}

\subsection{How much shift is needed: \texorpdfstring{$\tau_0$}{tau0} versus \texorpdfstring{$\tau_c$}{tauc}}
\label{app:shift}

\paragraph{Two thresholds.}
Problem~\eqref{eq:obj} is optimised over $\Delta_n$, whose affine hull has
direction space $\mathbf 1^\perp$. Strict convexity therefore requires
positive definiteness of $\widetilde{\bm S}$ only on $\mathbf 1^\perp$, not on
$\mathbb R^n$.

\begin{lemma}[Convexity on the simplex]
\label{lem:cpsd}
Let $\bm B\in\mathbb R^{n\times(n-1)}$ have orthonormal columns spanning
$\mathbf 1^\perp$ and define
\[
  \tau_0:=-\lambda_{\min}(\bm S),\qquad
  \tau_c:=-\lambda_{\min}(\bm B^\top\bm S\bm B)\;\le\;\tau_0 .
\]
The objective \eqref{eq:obj} is strictly convex on $\Delta_n$ if and only
if $\tau>\tau_c$, and $\widetilde{\bm S}\succ0$ if and only if $\tau>\tau_0$.
\end{lemma}
\begin{proof}
For $\bm p,\bm p+\bm h\in\Delta_n$ we have $\bm h\in\mathbf 1^\perp$, and the
second-order term of \eqref{eq:obj} along $\bm h$ is
$\tfrac\lambda2\bm h^\top\widetilde{\bm S}\bm h
=\tfrac\lambda2\bm z^\top(\bm B^\top\bm S\bm B+\tau\bm I)\bm z$ with
$\bm h=\bm B\bm z$. This is positive for all $\bm z\neq0$ iff
$\tau>\tau_c$ \citep{debreu1952}. The second claim is immediate, and
$\tau_c\le\tau_0$ by Cauchy interlacing.
\end{proof}

\noindent($\bm B$ plays the role of $\bm U$ in the main text, \Cref{eq:tauc}.)
Three regimes follow. For $\tau>\tau_0$ the objective is strictly convex on
$\mathbb R^n$ and Proposition~\ref{prop:projection} is a projection in a norm.
For $\tau_c<\tau\le\tau_0$ the minimiser over $\Delta_n$ is still unique, the
identity of Proposition~\ref{prop:projection} still holds algebraically, and
$\|\cdot\|_{\widetilde{\bm S}}$ is a norm on $\mathbf 1^\perp$ only;
Corollary~\ref{cor:interp} holds wherever $\widetilde{\bm S}$ is nonsingular,
since it uses only the equality-constrained KKT system. For $\tau\le\tau_c$
the problem is non-convex on $\Delta_n$ and projected gradient may stop at
different local minima.

\begin{figure*}[t]
  \centering
  \begin{subfigure}[t]{0.32\textwidth}\centering
    \includegraphics[width=\linewidth]{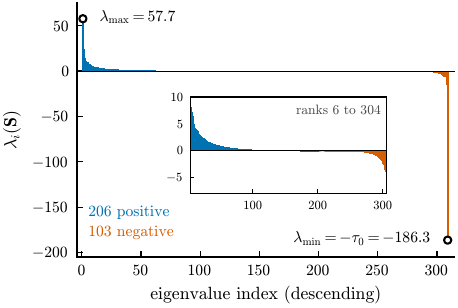}
    \caption{Eigenvalues of $\bS$.}\label{fig:spec-spectrum}
  \end{subfigure}\hfill
  \begin{subfigure}[t]{0.32\textwidth}\centering
    \includegraphics[width=\linewidth]{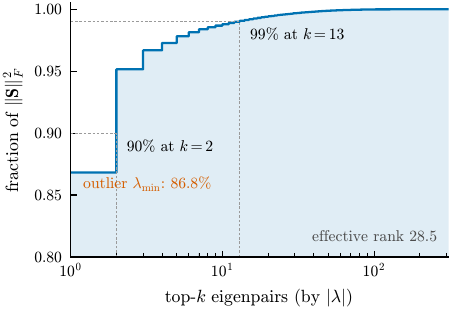}
    \caption{Cumulative spectral energy.}\label{fig:spec-energy}
  \end{subfigure}\hfill
  \begin{subfigure}[t]{0.32\textwidth}\centering
    \includegraphics[width=\linewidth]{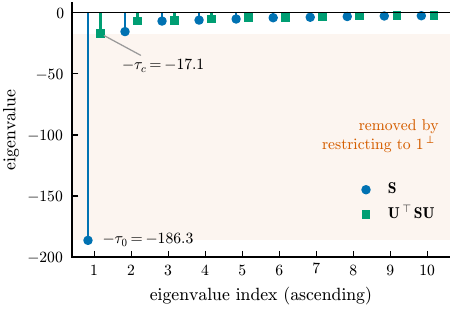}
    \caption{$\bS$ vs.\ $\bS$ restricted to $\one^\perp$.}\label{fig:spec-perp}
  \end{subfigure}
  \caption{Eigen-structure of the PMI affinity matrix $\bS$ ($n=309$).
  (a) Sorted eigenvalues (blue: positive, orange: negative), with an inset on
  ranks $6$--$304$. $\bS$ is strongly indefinite, with one outlier
  $\lambda_{\min}=-\tau_0=-186.3$; the next eigenvalue is $-15.4$.
  (b) Fraction of $\norm{\bS}_F^2$ captured by the $k$ largest-magnitude
  eigenpairs. The outlier alone carries $86.8\%$; $2$ eigenpairs give $90\%$
  and $13$ give $99\%$ (entropy effective rank $28.5$).
  (c) The $10$ smallest eigenvalues of $\bS$ and of its restriction
  $\bm B^\top\bS\bm B$ to $\one^\perp$. Restriction removes the outlier and
  raises the threshold from $-\tau_0=-186.3$ to $-\tau_c=-17.1$.}
  \label{fig:spectrum}
\end{figure*}

\begin{figure}[t]
  \centering
  \begin{subfigure}[t]{0.49\columnwidth}\centering
    \includegraphics[width=\linewidth]{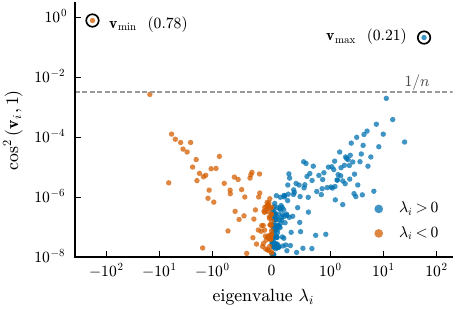}
    \caption{Alignment with $\one$.}\label{fig:spec-alignment}
  \end{subfigure}\hfill
  \begin{subfigure}[t]{0.49\columnwidth}\centering
    \includegraphics[width=\linewidth]{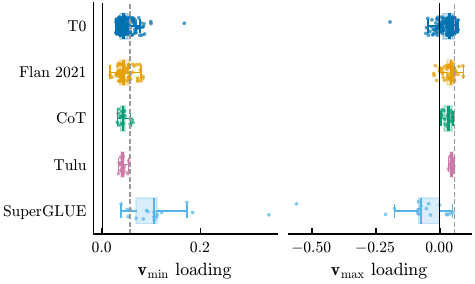}
    \caption{Loadings by collection.}\label{fig:spec-loadings}
  \end{subfigure}
  \caption{(a) $\cos^2(\bm v_i,\one)$ against $\lambda_i$ for all $309$
  eigenvectors of $\bS$ (dashed: $1/n$). Only $\bm v_{\min}$ ($0.78$) and
  $\bm v_{\max}$ ($0.21$) are substantially aligned with $\one$.
  (b) Loadings of $\bm v_{\min}$ and $\bm v_{\max}$ by source collection
  (dashed: $1/\sqrt n$, the loading of a uniform vector). $\bm v_{\min}$ is
  close to uniform for T0, Flan, CoT and Tulu (larger on SuperGLUE), while
  $\bm v_{\max}$ separates SuperGLUE from the other collections.}
  \label{fig:spec-extra}
\end{figure}

\paragraph{Why $\tau_c\ll\tau_0$: the spectrum of $\bS$.}
For our affinity matrix, $\tau_0=186.30$ but $\tau_c=17.09$, an order of
magnitude smaller (\Cref{fig:spectrum}). $\bS$ has $206$ positive and $103$
negative eigenvalues, $\lambda_{\max}=57.68$, and a single negative outlier,
$\lambda_{\min}=-186.30$; the next eigenvalue is $-15.42$. The outlier alone
carries $86.8\%$ of $\|\bS\|_F^2$. Its eigenvector $\bm v_{\min}$ satisfies
$\bm v_{\min}^\top\mathbf 1/\sqrt n=0.884$ ($\cos^2=0.78$): it lies mostly
along $\mathbf 1$, the one direction that is fixed on $\Delta_n$, and it loads
almost uniformly on every task (\Cref{fig:spec-extra}). Every other
eigenvector is essentially orthogonal to $\mathbf 1$ ($\cos^2\le2.7\times10^{-3}$,
below the $1/n$ expected for a random direction). Restricting $\bS$ to
$\mathbf 1^\perp$ removes the outlier (\Cref{fig:spec-perp}), so the PSD
threshold $\tau_0$ is driven by a direction the optimisation never moves in.

\paragraph{Offset invariance.}
This is the spectral face of the invariance in \Cref{para:offset-invariance}:
replacing $\bS$ by $\bS+c\mathbf 1\mathbf 1^\top$ changes \eqref{eq:obj} by a
constant on $\Delta_n$ and leaves $\tau_c$ unchanged, while
$\lambda_{\min}(\bS+c\mathbf 1\mathbf 1^\top)\to-\tau_c$ as $c\to\infty$.
The PMI scores carry such an offset (mean entry $-0.43$), but it is not
uniform across rows: subtracting only the mean entry raises
$\lambda_{\min}$ from $-186.3$ to $-98.6$, and the full restriction to
$\mathbf 1^\perp$ is needed to reach $-\tau_c$.

\label{app:spectrum}

\begin{figure}[t]
  \centering
  \includegraphics[scale=1.1]{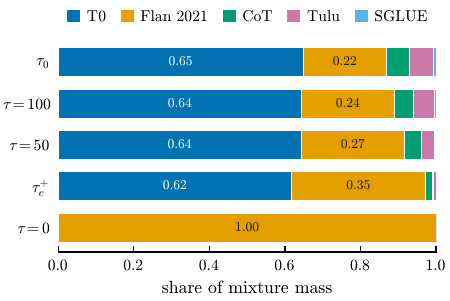}
  \caption{Share of mixture mass per source collection as the shift
  decreases. Down to $\tau_c^{+}$ the collection-level allocation is stable
  (T0 $0.62$--$0.65$), with \textsc{Flan2021} growing from $0.22$ to $0.35$.
  Without the shift all mass goes to one \textsc{Flan2021} task.}
  \label{fig:shift-mass}
\end{figure}

\paragraph{What the shift does to the mixture.}
\Cref{fig:shift} in the main text reports $\vp^\ast(\tau)$ as $\tau$ decreases
from $\tau_0$ at fixed $\rho$. The mixture contracts smoothly: its support
falls from $206$ tasks at $\tau_0$ to $165$, $123$ and $67$ at $\tau=100$, $50$
and $\tau_c^{+}$, and its effective size $e^{H(\vp^\ast)}$ from $179.6$ to
$44.6$, while the collection-level allocation stays stable (T0 receives
$0.62$--$0.65$ of the mass; \Cref{fig:shift-mass}). Without any shift
($\tau=0$) the problem is non-convex on $\Delta_n$: projected gradient from $28$
initialisations (uniform, $\vp^\ast(\tau_0)$, $20$ Dirichlet draws, single-task
vertices) always ends at a single-task vertex, $27$ of them at the same task
(\textsc{Flan2021 math\_dataset\_algebra\_linear\_1d}), at
$\mathrm{TV}(\vp^\ast(0),\vp^\ast(\tau_0))=0.985$. Removing the shift is
therefore not a meaningful ablation of the mixture; it changes the problem
from a unique, broad minimiser to an order-dependent single task.

\label{app:shift-mass}

\paragraph{The shift we use.}
All experiments use $\tau=\tau_0+10^{-3}=186.30$, the smallest shift for which
$\widetilde{\bS}\succ0$. This keeps Proposition~\ref{prop:projection} valid
verbatim, gives the broadest mixture among the convex choices at a fixed
$\rho$, and leaves the breadth of the mixture to be set by $\rho$ alone
(\Cref{sec:b2}). Any $\tau\in(\tau_c,\tau_0]$ also yields a unique mixture;
smaller $\tau$ gives a sparser one.

\subsection{The trade-off ratio \texorpdfstring{$\rho$}{rho} as a coverage control}
\label{sec:b2}

\paragraph{Only the ratio matters.}
Dividing \eqref{eq:obj} by $\lambda$ leaves the minimiser unchanged, so
$\vp^\ast$ depends on $\beta$ and $\lambda$ only through $\rho=\beta/\lambda$,
and we set $\rho$ directly. By Proposition~\ref{prop:projection}, $\vp^\ast$ is
the point of $\Delta_n$ closest to $\rho\mathbf 1$ in the affinity geometry, so
$\rho$ is the per-task mass the objective would \emph{like} every task to
receive. Its natural unit is $1/n$: $n\rho$ is the total mass requested,
relative to the one unit the simplex can provide.

\paragraph{How $n\rho$ controls coverage.}
Corollary~\ref{cor:interp} gives the picture. At $n\rho=1$ the request is
exactly feasible and $\vp^\ast=\bm u$: every task is covered equally. For
$n\rho>1$ the request exceeds the budget by a factor $n\rho$, the unconstrained
solution $\tilde{\bm p}=n\rho\,\bm u-(n\rho-1)\bm q$ moves away from the
minimum-redundancy mixture $\bm q$ towards high-degree tasks
(Remark~\ref{rem:degree}), and the nonnegativity constraint switches off the
tasks that $\tilde{\bm p}$ would make negative. Larger $n\rho$ therefore means
fewer active tasks, concentrated on the most representative ones; $n\rho$
close to $1$ means nearly uniform coverage. The KKT conditions make this
precise: every active task has the same net utility
$\rho(\bm d+\tau\mathbf 1)_{[i]}-(\widetilde{\bS}\vp^\ast)_{[i]}=\nu$
(\Cref{Appendix Sec: Lagrangian}), so a task is included exactly when its degree
clears a threshold that rises with $n\rho$. Equivalently, the KKT fixed point is
$\vp^\ast=\big[(\rho\bm d-\bS\vp^\ast)/\tau-\theta\mathbf 1\big]_+$: a thresholded,
degree-driven allocation \citep{martins2016} corrected by the pairwise term
$\bS\vp^\ast$. \Cref{tab:rho} shows the resulting path: as $n\rho$ grows from
$1.85$ to $618$, the support shrinks from $302$ to $6$ tasks and the effective
size from $289.8$ to $2.4$, while the effect of the pairwise term shrinks from
TV $0.097$ to $0.001$.

\begin{table}[t]
\centering
\caption{The $\rho$ path at $\tau=\tau_0+10^{-3}$. $\rho$ alone controls
breadth. ``Interaction effect'' is the TV between $\vp^\ast$ and the solution
of the same problem with $\bS$ removed from the quadratic term,
$\Pi_{\Delta_n}\!\big(\rho(\bm d+\tau\mathbf 1)/\tau\big)$; it grows as
$n\rho$ decreases. The operating point is in bold.}
\label{tab:rho}
\small
\begin{tabular}{rrrrr}
\toprule
$\rho$ & $n\rho$ & active & $e^{H}$ & interaction effect\\
\midrule
0.0060 & 1.85 & 302 & 289.8 & 0.097\\
0.0100 & 3.09 & 275 & 250.4 & 0.065\\
0.0150 & 4.63 & 246 & 218.8 & 0.050\\
0.0200 & 6.18 & 225 & 194.5 & 0.039\\
0.0220 & 6.80 & 210 & 186.1 & 0.036\\
\textbf{0.02424} & \textbf{7.49} & \textbf{206} & \textbf{179.6} & \textbf{0.035}\\
0.0270 & 8.34 & 201 & 171.6 & 0.032\\
0.0300 & 9.27 & 194 & 163.5 & 0.030\\
0.0400 & 12.36 & 169 & 141.0 & 0.021\\
0.0700 & 21.63 & 127 & 105.7 & 0.011\\
0.2000 & 61.80 & 73 & 51.0 & 0.006\\
0.5000 & 154.5 & 33 & 18.1 & 0.003\\
2.0000 & 618.0 & 6 & 2.4 & 0.001\\
\bottomrule
\end{tabular}
\end{table}

\paragraph{The operating point $\rho=0.02424$.}
All experiments use $\rho=0.02424$ ($n\rho=7.49$; reported as $0.0242$ in the
main text), together with $\tau=\tau_0+10^{-3}$. The value is a declared
\emph{coverage} choice, made before fine-tuning: it keeps two thirds of the
pool active ($206$ of $309$ tasks, effective size $e^{H(\vp^\ast)}=179.6$), so
that the budget covers most of the pool's breadth, while still excluding the
tasks whose degree is lowest. The affinity matrix alone does not identify a
unique $\rho$: intrinsic criteria, such as knees of the support-size or
effective-size curves along the $\rho$ path, disagree with one another by an
order of magnitude. We therefore report the entire path (\Cref{tab:rho}) and
the behaviour of the mixture along it (\Cref{tab:pairwise}). At this point the
pairwise term is active but moderate: removing it moves $3.5\%$ of the mass,
and within its support $\vp^\ast$ stays within TV $0.011$ of a degree-only
allocation with the same support size (Spearman$(\vp^\ast,\bm d)=0.999$ on the
support). In other words, at $n\rho=7.49$, $\rho$ mainly sets \emph{how many}
tasks receive weight, and the weights within that support follow task degree
(\Cref{app:pairwise}).

\subsection{Solver and its verification}
\label{app:algorithm}

\paragraph{Algorithm.}
For $\tau>\tau_c$, \eqref{eq:obj} is a strictly convex quadratic program over
$\Delta_n$, and we solve it exactly with \Cref{alg:mixture}: accelerated
projected gradient (FISTA; \citealp{beck2009fast}) on
$\tfrac12\vp^\top\widetilde{\bS}\vp-\rho\,\one^\top\widetilde{\bS}\vp$ (the objective
divided by $\lambda$), with gradient $\widetilde{\bS}(\vp-\rho\one)$, step $1/L$ with
$L=\lambda_{\max}(\widetilde{\bS})$, the exact Euclidean projection onto
$\Delta_n$ by sorting \citep{duchi2008efficient,condat2016fast}, uniform
initialisation, and the stopping rule
$\lVert\vp_{j+1}-\vp_j\rVert_1<\epsilon$. Plain projected gradient with step
$\eta\le1/L$ converges at rate $O(1/t)$, FISTA at $O(1/t^2)$, and both
converge linearly once $\tau>\tau_c$ by strong convexity on the simplex. Each
iteration costs one matrix--vector product and one $O(n\log n)$ projection.

\paragraph{Why not the closed form.}
The equality-constrained stationary point $\tilde{\bm p}$
(Corollary~\ref{cor:interp}, \eqref{eq:closed_form}) is the minimiser only when
$\tilde{\bm p}\ge0$. At our operating point $66$ of its $309$ entries are
negative, so it is not feasible, and renormalising its positive part does not
give a minimiser of \eqref{eq:obj} (it violates the KKT conditions of
\Cref{Appendix Sec: Lagrangian}). We therefore always solve the constrained
problem exactly.

\paragraph{Verification of the deployed mixture.}
With $\rho=0.02424$, $\tau=\tau_0+10^{-3}$ and $\epsilon=10^{-14}$,
\Cref{alg:mixture} converges in $38$ iterations, in under a second on a CPU.
The solution has $206$ active tasks; the gradient is constant on the support to
$3\times10^{-14}$ and exceeds it by at least $4.7\times10^{-3}$ on every inactive
task, so the full KKT conditions hold, including complementary slackness, and
the solution is the unique minimiser. Re-running the algorithm reproduces the
mixture file used for fine-tuning to TV $5\times10^{-16}$; a looser tolerance
$\epsilon=10^{-12}$ changes it by TV $2.5\times10^{-13}$. Rounding $\rho$ to
$0.0242$ would change it by TV $3.9\times10^{-4}$ (still $206$ tasks), which is
why we report the unrounded value here.

\subsection{From the mixture to a training set}
\label{app:sampling}

Given $\vp^\ast$ and a budget $\mathcal B$, per-task counts are obtained as
follows:
(i) Draw $5$ independent samples $\bm k^{(r)}\sim\mathrm{Mult}(\mathcal B,\vp^\ast)$
with seeds $42,\dots,46$ and average them, which reduces the sampling noise of a
single draw while keeping $\mathbb E[k_i]=\mathcal B\,\vp^\ast_{[i]}$.
(ii) Round the average to integers summing exactly to $\mathcal B$ by the
largest-remainder method.
(iii) If a count exceeds its task's size, cap it at $|\mathcal D_i|$ and
redistribute the deficit proportionally among the tasks with headroom, until no
task is over capacity. At $\mathcal B=25$K no task is capped; at
$\mathcal B=50$K one task is (\textsc{tulu\_hardcoded}, $240$ examples, expected
$289$).
(iv) Sample $k_i$ examples from task $\T_i$ uniformly without replacement, with a
per-task seed derived from the base seed and the task index.
Tasks outside the support of $\vp^\ast$ contribute nothing; every active task
has an expected count of at least one at both budgets.

% ############################################################################
%  PART III: EXPERIMENTAL DETAILS
% ############################################################################

\section{Experimental Details}
\label{section:Experimental Details}

\subsection{Computing the affinity matrix}
\label{app:affinity-impl}

\paragraph{One affinity, reused for all models.}
$\bS$ is computed once, from Qwen2-7B single-task models, and the resulting
mixture is reused unchanged for all three base models (Qwen2-7B,
Llama-2-7B and Qwen3.5-9B). We do not recompute $\bS$ per model, so the
Llama-2-7B and Qwen3.5-9B results also test whether a mixture derived on
one model transfers to other model families and sizes.

\paragraph{Single-task models.}
For each of the $309$ tasks we train a LoRA adapter on Qwen2-7B (rank $8$,
$\alpha=32$, dropout $0.1$, on the query, key, value and output projections)
for one epoch on the task's full training split, with learning rate
$2\times10^{-5}$, cosine schedule, weight decay $0.01$, bf16, per-device batch
$16$ on $4$ GPUs, maximum sequence length $4096$ and seed $23$. Task sizes range
from $240$ to $363{,}846$ examples (median $10{,}000$, $6.32$M in total). Each
adapter is about $18$\,MB.

\paragraph{Cross-evaluation.}
Each adapter is merged into the base model and scores every task on
$m=\min(2048,|\mathcal D_j|)$ held-out examples (never used to train any
adapter), drawn with a fixed seed; $277$ of the $309$ tasks exceed the cap.
Scoring uses maximum length $512$ and computes the mean per-token
log-probability of the response, averaged over examples. This gives the
$n\times n$ log-likelihood matrix $\bm L$, and
$\bS=\tfrac12\big[(\bm L-\one\,\mathrm{diag}(\bm L)^\top)+(\cdot)^\top\big]$ with
zero diagonal (\Cref{eq:PMI_matrix}). Because the scoring examples are held
out, the diagonal $\bm L_{jj}$ measures generalisation of the own-task model
rather than memorisation, and is not inflated relative to the cross terms.

\paragraph{Cost and scaling.}
The procedure needs $n$ LoRA runs (cost linear in the pool size,
$\sum_i|\mathcal D_i|$) and $n^2=95{,}481$ model--task evaluations of at most $m$
examples each (cost $O(n^2m)$), which dominates as $n$ grows but parallelises
trivially across adapters. Empirically, computing the complete PMI matrix for Qwen2-7B on 8-H100 GPUs required approximately 16 hours of wall-clock time, corresponding to about 128 H100-GPU-hours in total, or \(0.41\) GPU-hours per adapter row. 
% \pc{Replace this estimate with the measured LoRA and scoring hours of the Qwen2-7B run, and make the cost statement in the main text (Comparison with LESS) consistent with it.}
% The procedure requires \(n\) task-specific LoRA runs, so the adaptation cost grows linearly with the task pool and the total amount of training data, \(\sum_i |D_i|\). Computing the affinity matrix requires \(n^2=95{,}481\) model–task evaluations for \(n=309\), with at most \(m\) sampled examples per evaluation, giving a scoring cost of \(O(n^2m)\). These evaluations parallelize naturally across task-specific adapters. 

The cost is paid once per task pool. Adding a task requires one LoRA run and
$2n$ new evaluations, without recomputing existing entries. Storage is
negligible: the symmetric matrix has $n(n-1)/2=47{,}586$ entries
($\approx0.18$\,MB in float32).

\subsection{Final fine-tuning on the mixture}
\label{app:finetune}

The reported models are fine-tuned on training sets sampled from
$\vp^\ast$ (\Cref{app:sampling}) at budgets of $25$K and $50$K examples, for one
epoch, with AdamW (learning rate $2\times10^{-5}$, weight decay $0.01$), bf16,
DDP and gradient checkpointing, and maximum sequence length $2048$. To reduce
run-to-run variability, for each base model, budget, and method we repeat
fine-tuning three times on the same mixture with different seeds (affecting
data order, dropout, and other stochastic components) and report the mean; when
shown, the $\pm$ subscript denotes the standard deviation across these three
runs.
All runs use LoRA with rank \(r=8\), \(\alpha=32\), and dropout 0.1. For Qwen2-7B and Llama-2-7B, LoRA is applied to the attention projections \texttt{q\_proj}, \texttt{k\_proj}, \texttt{v\_proj}, and \texttt{o\_proj}. For Qwen3.5-9B, we additionally adapt \texttt{in\_proj\_qkv}, \texttt{in\_proj\_z}, \texttt{in\_proj\_b}, \texttt{in\_proj\_a}, \texttt{out\_proj}, and the MLP projections \texttt{gate\_proj}, \texttt{up\_proj}, and \texttt{down\_proj}. All other optimization and training hyperparameters are held fixed across models and methods. Every method within a table, including all baselines, is trained with the same configuration on the same base model, so differences reflect only the training data.

% \pc{Fill in the final fine-tuning configuration of the new runs (it is not on the affinity server): GPU type and count, per-device batch size, gradient accumulation and effective batch size, warmup ratio (the previous draft listed both 1\% and 3\%), and whether fine-tuning was full-parameter or LoRA (with target modules). Earlier Qwen3.5-9B runs used LoRA ($r=8$, $\alpha=32$, dropout $0.1$) on all attention and MLP projections with an effective batch of 32--64.}

\subsection{Baselines}
\label{app:baselines}

All baselines draw from the same $309$-task pool and budget $\mathcal B$.
\textsc{Uniform}, \textsc{EPM} and \textsc{Random} are defined in the main text
(\Cref{itm:uniform,itm:epm,itm:random}) and use the same capping and
within-task sampling as \Cref{app:sampling}. \textsc{SMART}
\citep{renduchintala-etal-2024-smart} allocates the budget across tasks with its
submodular objective and selects non-redundant examples within each task.
\textsc{LESS} \citep{xia2024less} uses a LoRA warm-up on a random 5\% of the pool (seed 3), Adam-preconditioned training gradients projected to 4,096 dimensions, and standard SGD gradients on validation examples to form each benchmark’s average target gradient. It selects the top 0.25\% (25K) or 0.5\% (50K) examples by cosine similarity between training and validation gradients, then trains one checkpoint per benchmark.
% !!! [CRITICAL -- LESS audit, 6 Oct 2026] !!!
% /mnt/warm_storage/saral/less_q3: 316,638 NaN train gradients; NaN validation gradients for bbh/musr/math_lvl5;
% NaN rows sorted first -> every 25K/50K LESS subset is an arbitrary flan2021-only sample (identical for
% GPQA/IFEval/MMLU-Pro). GPQA/IFEval/MuSR/MATH targets are drawn from evaluation items. Re-run before reporting.
% \pc{The current LESS selections are invalid (NaN influence scores ranked first; targets drawn from evaluation items). Re-run LESS before reporting it.}

\subsection{Evaluation}
\label{app:evaluation}

We evaluate with the EleutherAI \texttt{lm-evaluation-harness}
\citep{sutawika2024eleutherai} on the six Open LLM Leaderboard benchmarks:
BBH and MuSR (reasoning), GPQA and MMLU-Pro (knowledge), MATH-hard (Level-5
MATH) and IFEval (instruction following; we reported instruction-level-loose accuracy), using the leaderboard task
definitions and few-shot settings. All numbers are averages over the three
fine-tuning repeats described above.
% \pc{State the harness version and the IFEval metric reported (prompt- or instruction-level, strict or loose).}

\subsection{Comparison with LESS: Lack of Cross-Model Amortization}

LESS \citep{xia2024less} requires computing and storing per-example LoRA gradients for the entire candidate pool at each warm-up checkpoint, making gradient construction scale linearly with the pool size and number of checkpoints. This gradient-storage stage alone required approximately 320 hours of wall-clock time on 4 H100 GPUs. Importantly, these gradients are model-specific; gradients computed using Llama2 or Qwen2 cannot be reused directly when switching to Qwen3.5, since both the parameterization and resulting optimization geometry change. Thus, LESS requires repeating the warm-up and gradient-construction stages for new models. The datastore also incurs substantial storage overhead, since per-example gradient representations must be materialized and retained across warm-up checkpoints. The gradient construction for Qwen3.5-9B model required roughly 460 hours on 4 H100s under the same data and sequence-length . This model-dependent recomputation represents a substantial disadvantage relative to a model-agnostic \methodprop{} method, for which the selection artifact can be reused across model families without incurring this cost again.

% ############################################################################
%  PART IV: ADDITIONAL RESULTS AND ANALYSIS
% ############################################################################

\section{Additional Results and Analysis}
\label{section:Additional Results}

\subsection{Structure of the learned mixture}
\label{section:mixture_analysis}
\label{app:q2}

\paragraph{Sparse, broad selection.}
The deployed mixture assigns positive weight to $206$ of the $309$ tasks
(effective size $179.6$; entropy $5.19$ against the uniform maximum
$\ln309=5.73$). The sparsity is not imposed by any cardinality constraint: it
follows from the KKT threshold of \Cref{sec:b2}, which excludes the $103$ tasks
with the lowest degree.

\begin{figure}[h!]
  \centering
  \includegraphics[width=0.35\linewidth,height=0.35\linewidth]{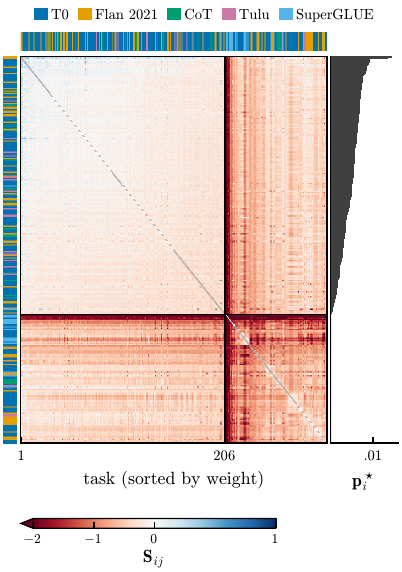}
  \caption{$\bS$ reordered by learned weight: the $206$ active tasks in
  decreasing $\vp^\ast_{[i]}$, followed by the $103$ inactive tasks
  (average-linkage clustering). Side strips show the source collection and
  $\vp^\ast_{[i]}$. Colour scale clipped at $-2$.}
  \label{fig:q2-heatmap}
\end{figure}

\paragraph{Which tasks are kept.}
Reordering $\bS$ by learned weight (\Cref{fig:q2-heatmap}) shows the structure
the optimiser exploits. The $206$ active tasks form a block of weak mutual
affinity ($\bS_{ij}$ near $0$), while the $103$ inactive tasks have strongly
negative affinity to almost every task, including each other: they are
specialised tasks whose models transfer poorly and whose data other tasks'
models predict poorly. The highest-weighted tasks, the ``attractors'', are broad
reading-comprehension and QA prompts from T0 (RACE, QuAIL, Social IQa,
WebQuestions, WikiQA) and Flan (linear-algebra math, STS-B, QuAC, Newsroom);
their weight ranks follow their degree ranks almost exactly (the top $15$ by
weight are the top $15$ by degree; \Cref{fig:q2}).

\paragraph{Collection-level allocation.}
T0 receives $0.649$ of the mass ($135$ of $193$ tasks active), Flan~2021
$0.221$ ($40/63$), Tulu $0.066$ ($15/18$), CoT $0.059$ ($14/18$) and SuperGLUE
$0.006$ ($2/17$). SuperGLUE prompts are the least transferable in this pool, a
bias of degree-driven weighting noted in the limitations.

\subsection{Role of the pairwise term}
\label{app:pairwise}

We compare $\vp^\ast$ with two variants that keep only the unary (degree)
information. \emph{Pairwise-removed} replaces $\widetilde{\bS}$ by $\tau\mathbf I$
in the quadratic term, giving $\Pi_{\Delta_n}\!\big(\rho(\bm d+\tau\one)/\tau\big)$.
\emph{Degree, support-matched} is $\Pi_{\Delta_n}(a\,\bm d)$ with the scale $a$
chosen so its support has the same size as $\vp^\ast$. At every $\rho$ we tested,
$\vp^\ast$ lies within TV $0.011$ of the support-matched degree allocation, except
at $n\rho=618$ where only $6$ tasks are active (\Cref{tab:pairwise}). The pairwise
term mainly determines \emph{how many} tasks are active, through $\rho$, and
changes the weights within that support only slightly. At the operating point,
removing it moves $3.5\%$ of the mass and the support-matched degree rule
moves $1.1\%$.
%\pc{Downstream results for the prepared ablation mixtures (degree-only support-matched, uniform-on-support, size-proportional-on-support, greedy $k\in\{50,100\}$; files in \texttt{figs\_pmi\_qwen/ablations/}) are still to be added.}

\begin{table}[t]
\centering\small
\caption{Effect of the pairwise term along the $\rho$ path
($\tau=\tau_0+10^{-3}$). TV is measured from $\vp^\ast(\rho)$ to the
pairwise-removed solution at the same $\rho$ (no-pair) and to the degree-only
allocation with the same support size (deg.).}
\label{tab:pairwise}
\begin{tabular}{rrrrrr}
\toprule
$n\rho$ & $\rho$ & $|\mathrm{supp}|$ & $e^{H}$ & TV no-pair & TV deg. \\
\midrule
1.2 & 0.0039 & 308 & 306.8 & 0.120 & 0.003 \\
2 & 0.0065 & 299 & 285.2 & 0.093 & 0.009 \\
4 & 0.0129 & 252 & 229.5 & 0.054 & 0.008 \\
\textbf{7.49} & \textbf{0.0242} & \textbf{206} & \textbf{179.6} & \textbf{0.035} & \textbf{0.011} \\
10 & 0.0324 & 190 & 157.4 & 0.027 & 0.009 \\
20 & 0.0647 & 131 & 109.5 & 0.012 & 0.005 \\
100 & 0.3236 & 54 & 32.3 & 0.005 & 0.004 \\
618 & 2.0000 & 6 & 2.4 & 0.001 & 0.023 \\
\bottomrule
\end{tabular}
\end{table}

\subsection{Incremental composition}
\label{app:incremental}

\Cref{fig:incr} adds tasks one at a time, in greedy order and in five random
orders, and re-solves on the growing support, measuring both the task affinity
of the main text and its renormalised version $\widetilde{\mathrm{TV}}_k$.
With the shift, the existing mixture is almost unaffected by each new task:
$\widetilde{\mathrm{TV}}_k$ has median $0.0026$ for $k<10$ and $0.0010$ for
$k\ge150$, and never exceeds $0.007$. The support grows steadily to the $206$
tasks of $\vp^\ast(\tau_0)$, and the final mixture equals $\vp^\ast(\tau_0)$
(TV $<10^{-13}$) for every order. Without the shift every intermediate mixture is
a single task: in random orders the mixture stays the same at $98\%$ of steps,
and at the remaining $4$--$9$ steps per order all mass jumps to the newly added
task. The final task then depends on the order; in $3$ of $5$ orders it is not
the globally best task.

\subsection{Larger budget on Llama-2-7B}
\label{app:more-results}

\Cref{tab:llama-100k} extends the Llama-2-7B comparison to a $100$K budget.
%\pc{The 25K and 50K PMI rows here match \Cref{tab:llama-25k-50k} exactly. Confirm that they, and the 100K row, were trained on the current mixture (\texttt{p\_star\_affinity.npy}) rather than the earlier one. The LESS rows inherit the problem described in \Cref{app:baselines}.}

\begin{table*}[!htbp]
\centering
\setlength{\tabcolsep}{3pt}
\caption{Llama-2-7b: Instruction-tuning performance on Leaderboard subsets using batch size 8. Scores are reported as mean $\pm$ std. over three fine-tuning repeats on the same mixture with different seeds).}
\label{tab:llama-100k}
\rowcolors{2}{gray!15}{white}
\resizebox{\textwidth}{!}{%
\begin{tabular}{lc*{5}{c} % Leaderboard subcolumns
}
\toprule
\multicolumn{1}{c}{Dataset}
% & \multicolumn{1}{c}{ }
  & \multicolumn{6}{c}{Leaderboard} \\
\cmidrule(lr){2-7}
Size / Method &  BBH    & GPQA   & IFEval & Math   & MMLU-Pro & MUSR   \\
\midrule

\multicolumn{7}{l}{\textbf{25K}}\\
Random     & 0.3482$_{\pm0.0059}$ & 0.2626$_{\pm0.0128}$ & 0.3465$_{\pm N/A}$ & 0.0098$_{\pm0.0027}$ & 0.1877$_{\pm0.0036}$ & 0.3677$_{\pm0.0172}$ \\
Uniform    & 0.3501$_{\pm0.0059}$ & 0.2701$_{\pm0.0129}$ & 0.3501$_{\pm N/A}$ & 0.0151$_{\pm0.0034}$ & 0.1768$_{\pm0.0035}$ & 0.4027$_{\pm0.0175}$ \\
EPM        & 0.3593$_{\pm0.0059}$ & 0.2601$_{\pm0.0127}$ & 0.3405$_{\pm N/A}$ & 0.0151$_{\pm0.0033}$ & 0.1836$_{\pm0.0035}$ & 0.4286$_{\pm0.0177}$ \\
LESS       & $0.4059_{\pm0.0055}$ & $0.2412_{\pm0.0088}$ & $0.3417_{\pm N/A}$ & $0.0001_{\pm0.0004}$ & $0.1816_{\pm0.0062}$ & $0.4157_{\pm0.0175}$ \\
\rowcolor{black!6}Ours (PMI) & \textbf{0.4095$_{\pm0.0054}$} & \textbf{0.2718$_{\pm0.0129}$} & \textbf{0.3561$_{\pm N/A}$} & \textbf{0.0159$_{\pm0.0034}$} & \textbf{0.1924$_{\pm0.0036}$} & \textbf{0.4298$_{\pm0.0177}$} \\

\midrule
\multicolumn{7}{l}{\textbf{50K}}\\
Random     & 0.3565$_{\pm0.0060}$ & 0.2668$_{\pm0.0128}$ & 0.3581$_{\pm N/A}$ & 0.0134$_{\pm0.0033}$ & 0.1811$_{\pm0.0036}$ & 0.3770$_{\pm0.0172}$ \\
Uniform    & 0.3480$_{\pm0.0059}$ & 0.2785$_{\pm0.0130}$ & 0.3901$_{\pm N/A}$ & 0.0161$_{\pm0.0037}$ & 0.1895$_{\pm0.0036}$ & 0.4057$_{\pm0.0176}$ \\
EPM        & 0.3532$_{\pm0.0059}$ & 0.2634$_{\pm0.0128}$ & 0.3507$_{\pm N/A}$ & 0.0128$_{\pm0.0031}$ & 0.1799$_{\pm0.0036}$ & 0.4206$_{\pm0.0176}$ \\
LESS       & $0.4133_{\pm0.0051}$ & $0.2475_{\pm0.0088}$ & $0.3407_{\pm N/A}$ & $0.0136_{\pm0.0003}$ & $0.1906_{\pm0.0011}$ & $0.4226_{\pm0.0176}$ \\
\rowcolor{black!6}Ours (PMI) & \textbf{0.4159$_{\pm0.0060}$} & \textbf{0.2794$_{\pm0.0130}$} & 0.3577$_{\pm N/A}$ & \textbf{0.0189$_{\pm0.0037}$} & \textbf{0.1967$_{\pm0.0035}$} & \textbf{0.4511$_{\pm0.0178}$} \\

\midrule
\multicolumn{7}{l}{\textbf{100K}}\\
Random      & 0.3458$_{\pm0.0062}$ & 0.2621$_{\pm0.0125}$ & 0.3705$_{\pm0.0078}$ & 0.0113$_{\pm0.0030}$ & 0.1893$_{\pm0.0035}$ & 0.4101$_{\pm0.0179}$ \\
Uniform     & 0.3569$_{\pm0.0057}$ & 0.2710$_{\pm0.0131}$ & 0.3801$_{\pm0.0084}$ & 0.0189$_{\pm0.0036}$ & 0.1890$_{\pm0.0037}$ & 0.3730$_{\pm0.0168}$  \\
EPM         & 0.3439$_{\pm0.0060}$ & 0.2844$_{\pm0.0134}$ & 0.3717$_{\pm0.0077}$ & 0.0098$_{\pm0.0027}$ & 0.1873$_{\pm0.0036}$ & 0.4259$_{\pm0.0174}$  \\
LESS         &  0.4195$_{\pm0.0057}$ & $0.2617_{\pm0.127}$ & 0.3504$_{\pm N/A}$ & 0.0151$_{\pm0.0033}$ & 0.1970$_{\pm0.0011}$ & 0.4341$_{\pm0.0175}$  \\
Ours (PMI)   &  \textbf{0.4201$_{\pm0.006}$} & \textbf{0.2853$_{\pm0.0125}$} & 0.3595$_{\pm N/A}$ & \textbf{0.0196$_{\pm0.0032}$} & \textbf{0.1984$_{\pm0.0036}$} & \textbf{0.4665$_{\pm0.0178}$}  \\

\bottomrule
\end{tabular}%
}
\end{table*}

The PMI mixture is strongest on BBH, MMLU-Pro and MuSR at every budget, reaching
$0.4201$ on BBH and $0.4665$ on MuSR at $100$K, and improves over Uniform and EPM
on BBH by up to $7$ points. On several benchmarks the PMI mixture at $50$K
matches or exceeds the heuristic mixtures at $100$K (e.g.\ MuSR and Math), which
indicates that the selected tasks carry more useful signal per example. The
heuristic baselines do not improve consistently with budget: Random and EPM gain
on some benchmarks while losing on others as the budget grows, because extra
examples drawn without regard to task structure are not necessarily useful.

% ############################################################################
%  PART V: FURTHER DISCUSSION
% ############################################################################

% ---- begin inlined: Appendix/comparative_similarities.tex ----
\section{Comparing Notions of Task Similarity}
\label{section:Comparative Analysis across various Notions of Task Similarities}
\label{app:similarity-notions}

\subsection{Similarity across Task Vectors via Linearized finetuning}
Large-scale pretrained language models (PLMs) such as GPT-2 are widely adapted to downstream tasks via full-model fine-tuning. However, multi-task or per-task retraining remains computationally burdensome. \emph{Task arithmetic}(~\cite{ortizjimenez2023taskarithmetic_tangent}) introduces a simple yet effective approach: given a pretrained checkpoint initialization $\btheta_0$ and task-specific fine-tuned weights $\btheta_t^*$, the \emph{task vector} is defined as:
\[
\tau_t := \btheta_t^* - \btheta_0
\]
These vectors enable model editing via linear composition:
\begin{itemize}
    \item \textbf{Addition:} $\btheta_0 + \sum_{t \in \mathcal{T}} \tau_t$ synthesizes multi-task behaviors.
    \item \textbf{Negation:} $\btheta_0 - \tau_s$ induces task-specific forgetting.
\end{itemize}
While effective, the underlying mechanisms behind this arithmetic remain poorly understood.

\textbf{Linearized Fine-Tuning}: \citep{ortizjimenez2023taskarithmetic_tangent} posit that \emph{tangent-space fine-tuning} disentangles task behaviors more effectively by constraining updates to the local linear approximation of the model. Let $f(x;\btheta)$ denote a PLM with parameters $\btheta \in \mathbb{R}^m$, the corresponding
\textbf{nonlinear task vector} is given by $\tau_t^{\mathrm{nl}} := \btheta_t^* - \btheta_0$.

In contrast, \emph{linearized fine-tuning} restricts optimization to the first-order Taylor expansion:
\[
f_{\mathrm{lin}}(x; \btheta) := f(x;\btheta_0) + \nabla_\btheta f(x;\btheta_0)^\top (\btheta - \btheta_0)
\]
This surrogate is optimized using Jacobian-vector products (JVP), yielding a linearized task vector:
\[
\tau_t^{\mathrm{lin}} := \btheta_t^{\mathrm{lin}*} - \btheta_0
\]

Task vectors are generally useful as they can enable model editing as well provide a well defined representation of the finetuning task at hand, dependent on the model parameters.
Ideally, the goal would be to select multiple linearly independent task vectors such that they represent generalizably well across a range of IFT datasets and does generalizably well across different benchmark datasets. The algorithm is presented as Algorithm 1 in Section~\ref{subsec:algorithms}.

\textbf{Similarity Structure of Task Embeddings}

Directly computing any similarity metric  over $m \sim 10^6 \text{ to } 10^9$ parameters, is computationally expensive. Thus, we first isolate the most informative layer (chosen via task‐vector analysis using \textbf{layer-wise subsetting} and then project its high-dimensional slice task vector $\tau \in \mathbb R^m$ to a much lower-dimensional vector $\tilde\tau=R\,\tau\in\mathbb R^k$ using a \textbf{Gaussian random matrix} $R\in\mathbb R^{k\times m}$ with $k \ll m$. This projection technique is known to preserve similarity distances in expectation, providing a reliable and efficient approximation for comparing vector directions in the reduced space.

\textbf{Cosine Similarity across Task Vectors}: To analyze inter-task relationships, we examine the cosine similarity between task vectors:
\[
\mathrm{sim}(\tau_A, \tau_B) := \frac{\tau_A^\top \tau_B}{\|\tau_A\|_2 \cdot \|\tau_B\|_2} \in [-1, 1]
\]
This metric probes the angular alignment between task-specific directions in parameter space. High similarity indicates shared representational updates; near-orthogonality suggests disentangled task pathways.

\paragraph{Analyzing Task Vector Relationships via Cosine Similarity and PMI:}

\begin{figure}[!t]
    \centering
    \begin{subfigure}[t]{0.95\textwidth}
        \centering
        \includegraphics[width=\linewidth,height=0.32\textheight,keepaspectratio]{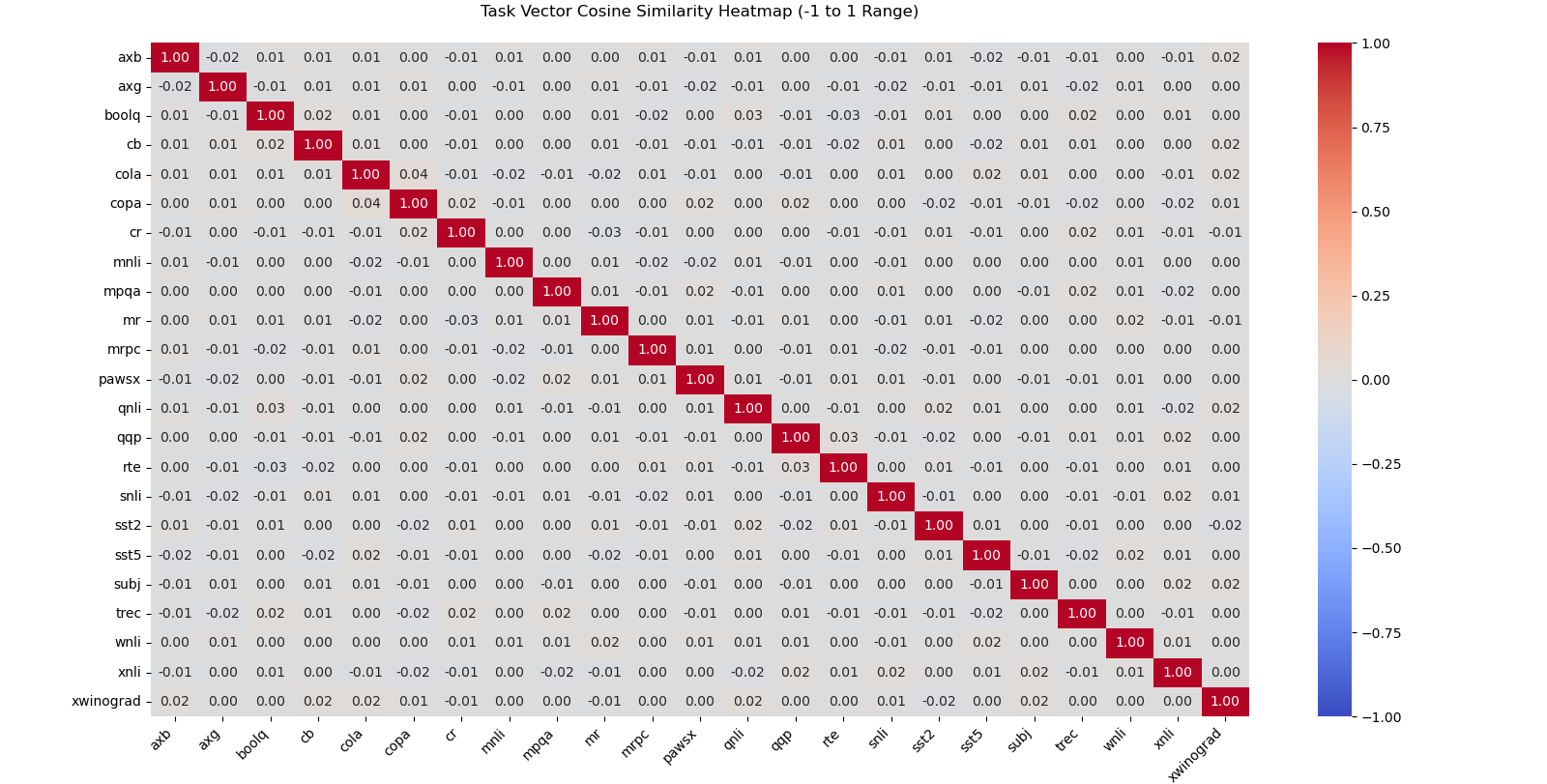}
        \caption{Cosine Similarity}
        \label{fig:cosine}
    \end{subfigure}

    \vspace{0.75em}

    \begin{subfigure}[t]{0.95\textwidth}
        \centering
        \includegraphics[width=\linewidth,height=0.32\textheight,keepaspectratio]{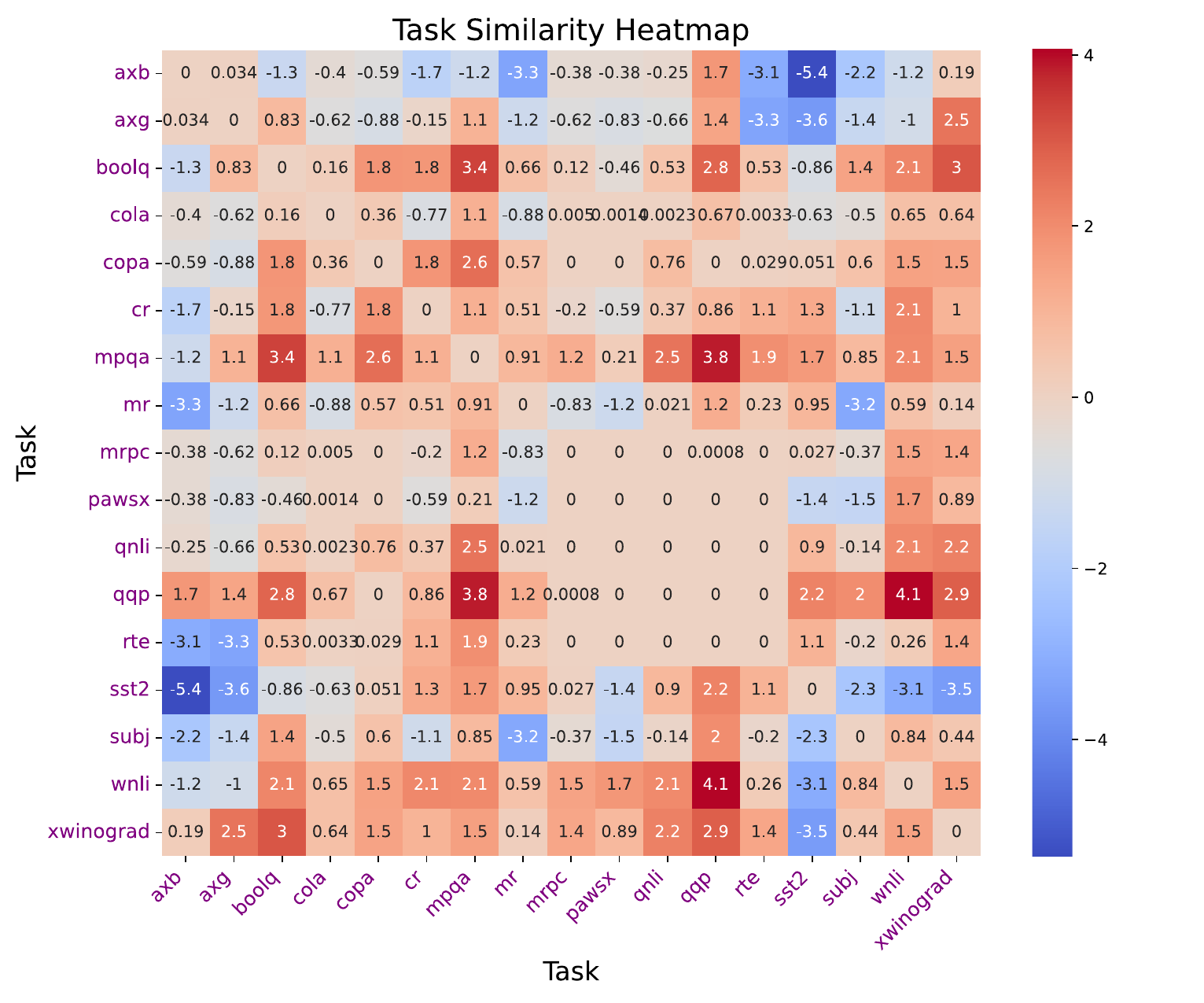}
        \caption{PMI}
        \label{fig:pmi}
    \end{subfigure}

    \caption{Comparison of task similarity metrics using cosine similarity and PMI-based heatmaps. Cosine scores are generally low and fail to distinguish task structure. PMI highlights asymmetric task alignment and reveals clearer task groupings across models.}
    \label{fig:task_similarity_heatmaps}
\end{figure}

While \textbf{Cosine Similarity} is a commonly used metric for comparing vector representations, it falls short in capturing nuanced differences in model behavior when applied to classification probability distributions. Cosine only measures the angular similarity between two vectors and is therefore invariant to vector magnitude. Hence, two models assigning vastly different probabilities but in the same proportional direction can still yield a high cosine score, misleadingly implying strong similarity. This limitation becomes evident in our experimental heatmap (Figure~\ref{fig:cosine}), where task relationships are not clearly differentiated as many unrelated tasks appear spuriously similar due to their shared vector directionality. Moreover, cosine similarity does not adequately account for uncertainty or confidence in model outputs. \\
\noindent
To address these issues, we used Pointwise Mutual Information \textbf{(PMI)}, which offers better theoretical grounding and practical discriminability. As shown in Figure~\ref{fig:pmi}, PMI captures directional alignment of model predictions with respect to task-specific specialization and yields a more interpretable heatmap where related tasks cluster more meaningfully and task-specific behaviors are more distinctly captured.

Concretely, the cosine heatmap appears overly uniform and thus masking important task groupings; whereas the PMI map exposes clear blocks of high intra-group similarity and low inter-group coupling. These results confirm that, for fine-grained task-similarity assessment in large models, information-theoretic measures substantially outperform simple angular alignment.\\

Figure~\ref{fig:task_similarity_heatmaps} visualizes the effects (on SGLUE tasks), comparing cosine and PMI heatmaps to illustrate their differing sensitivity to inter-task relationships.

\subsection{Algorithms for computing PMI}
\label{subsec:algorithms}

\textbf{Algorithm 1} performs fine-tuning by linearizing the model around its pretrained parameters. Instead of recomputing the full forward pass, it uses a Jacobian-vector product (JVP) to approximate the effect of parameter updates, allowing faster gradient-based updates in the “tangent space” of the original model.

\begin{tcolorbox}[
  colback=white!98!gray!2,
  colframe=black,
  title=\textbf{Algorithm 1: Linearized (Tangent-Space) Fine-Tuning},
  boxsep=2pt,
  left=2pt,right=2pt,top=2pt,bottom=2pt,
  before skip=4pt, after skip=4pt
]
\small
\textbf{Require:} Pretrained weights $\theta_0$, dataset $\mathcal{D}_t$ \\
\begin{tabular}{rl}
1: & Initialize $\theta \leftarrow \theta_0$ \\
2: & \textbf{while} not converged \textbf{do} \\
3: & \quad Sample mini-batch $(x, y) \sim \mathcal{D}_t$ \\
4: & \quad Compute base output $o_0 = f(x; \theta_0)$ \\
5: & \quad Compute JVP: $g = \mathrm{JVP}\bigl(f(\cdot; \theta_0),\, \theta - \theta_0; x\bigr)$ \\
6: & \quad $\hat{o} = o_0 + g$ \\
7: & \quad $\theta \leftarrow \theta - \eta\, \nabla_\theta \ell(\hat{o}, y)$ \\
8: & \textbf{end while} \\
9: & \textbf{return} $\theta_t^{\mathrm{lin}*}$ \\
\end{tabular}
\end{tcolorbox}

\textbf{Algorithm 2} quantifies how similarly two models $M_A$ and $M_B$ score the same labeled examples, using a pointwise mutual information (PMI)–inspired score. By averaging the log-ratio of predicted probabilities on each other’s held-out data, it produces a symmetric similarity score $S_{AB}$.

\begin{tcolorbox}[
  colback=white!98!gray!2,
  colframe=black,
  title=\textbf{Algorithm 2: PMI-Based Inter-Model Similarity $S_{AB}$},
  boxsep=2pt,
  left=2pt,right=2pt,top=2pt,bottom=2pt,
  before skip=4pt, after skip=4pt
]
\small
\textbf{Require:} Models $M_A$, $M_B$; datasets $\mathcal{D}^A$, $\mathcal{D}^B$ \\
\textbf{Ensure:} Similarity score $S_{AB}$

\vspace{0.2em}
\begin{tabular}{rl}
1: & Initialize accumulator $\texttt{sum}_B \gets 0$ \\
2: & \textbf{for all} $(x, y) \in \mathcal{D}^B$ \textbf{do} \\
3: & \quad Compute $p_A \gets M_A(x)$ and extract $p_A(y)$ \\
4: & \quad Compute $p_B \gets M_B(x)$ and extract $p_B(y)$ \\
5: & \quad Update $\texttt{sum}_B \mathrel{+}= \log\left( \frac{p_A(y)}{p_B(y)} \right)$ \\
6: & \textbf{end for} \\
7: & Set $\Delta_B \gets \tfrac{1}{|\mathcal{D}^B|} \cdot \texttt{sum}_B$ \\
8: & Initialize accumulator $\texttt{sum}_A \gets 0$ \\
9: & \textbf{for all} $(x, y) \in \mathcal{D}^A$ \textbf{do} \\
10: & \quad Compute $p_A \gets M_A(x)$ and extract $p_A(y)$ \\
11: & \quad Compute $p_B \gets M_B(x)$ and extract $p_B(y)$ \\
12: & \quad Update $\texttt{sum}_A \mathrel{+}= \log\left( \frac{p_B(y)}{p_A(y)} \right)$ \\
13: & \textbf{end for} \\
14: & Set $\Delta_A \gets \tfrac{1}{|\mathcal{D}^A|} \cdot \texttt{sum}_A$ \\
15: & \textbf{return} $S_{AB} \gets \tfrac{1}{2}(\Delta_A + \Delta_B)$ \\
\end{tabular}
\end{tcolorbox}
% ---- end inlined: Appendix/comparative_similarities.tex ----

\section{Potential Data Contamination}
\label{app:potential-contamination}

A fair evaluation requires that the fine-tuning mixture does not contain the benchmark questions it is later evaluated on. This section examines that risk for every source collection in our pool and reports an instance-level overlap audit.

We distinguish \emph{task- or capability-level relatedness} from \emph{instance-level evaluation leakage}. The former is expected in general-purpose instruction tuning: for example, our pool contains AQuA and StrategyQA, which exercise multi-step reasoning skills also required by BBH. Such relatedness does not imply that BBH evaluation questions or answers were observed during fine-tuning. We further describe documented safeguards for upstream sources.

\textbf{MMLU and BBH held-out.} The Flan Collection defines MMLU and BBH as held-out tasks that are not included in Flan-2022 fine-tuning \citep[Sec.~2.2]{longpre2023flan}. Its Appendix~B \citep[Appendix~B]{longpre2023flan} further states that MMLU tasks were removed from the Super-Natural Instructions component to preserve this separation. The Scaling Flan paper \cite{chung2022scaling} further reports removing 44 tasks related to MMLU and explicitly states that MMLU and BBH are held-out and absent from its fine-tuning data.

\textbf{Composition of the CoT mixture.} Scaling Flan lists the nine CoT training datasets as AQuA, CREAK, ECQA, ESNLI, GSM8K, QASC, QED, SenseMaking, and StrategyQA; MATH and BBH are not listed as CoT training sources \citep[A.4]{chung2022scaling}.

\textbf{Chronology of legacy sources.} The official T0 release dates to October 2021, while GLUE and SuperGLUE date to 2018 and 2019, respectively. By comparison, GPQA (Nov.~2023), IFEval (Nov.~2023), MuSR (Oct.~2023), and MMLU-Pro (Jun.~2024) post-date these sources. Therefore, as a chronological inference, the newly authored GPQA questions, newly constructed IFEval prompts, and newly generated MuSR narratives could not have appeared in the frozen official T0, GLUE/SuperGLUE, or earlier CoT releases.

\textbf{Tulu V3 decontamination.} For the later Tulu V3 component, we used the official decontaminated release. The Tulu V3 technical report states that a training dataset is excluded when its overlap exceeds 2\% of the Tulu evaluation suite, and the official Open-Instruct implementation supports full-text, token $n$-gram, and embedding-based matching with removal of matched training instances. We treat this as meaningful risk mitigation rather than proof of zero contamination, because the implementation notes that retrieval limits may prevent removal of every possible match.

\paragraph{Instance-level overlap audit.}

We conducted an instance-level overlap audit between the available \methodprop{} source pool and the final evaluation benchmarks. The audit covers 6{,}072{,}160 training and validation examples from 309 source tasks. For every evaluation example, we applied: (i) normalized exact matching (full-text equality after lowercasing and removing punctuation and prompt wrappers), (ii) token overlap using a 13-gram criterion (as in prior GPT-3 contamination analyses), and (iii) semantic matching to detect paraphrases and translated/reformatted copies (cf. Yang et al., 2023).

We audited the BBH leaderboard subset, GPQA leaderboard subsets, IFEval, MATH-hard level-5, MMLU-Pro, and MuSR. Table~\ref{tab:overlap-audit-app} summarizes the first-stage results.

\begin{table}[t]
  \centering
  \small
  \setlength{\tabcolsep}{5pt}
  \begin{tabular}{lrrrrr}
    \toprule
    \textbf{Benchmark} & \textbf{Eval. ex.} & \textbf{Exact} & \textbf{Exact rate} & \textbf{13-gram cand.} & \textbf{13-gram rate}\\
    \midrule
    BBH & 5{,}761 & 0 & 0.000\% & 30 & 0.521\%\\
    GPQA & 1{,}192 & 0 & 0.000\% & 1 & 0.084\%\\
    IFEval & 541 & 0 & 0.000\% & 5 & 0.924\%\\
    MATH-hard & 1{,}324 & 0 & 0.000\% & 275 & 20.770\%\\
    MMLU-Pro & 12{,}032 & 0 & 0.000\% & 112 & 0.931\%\\
    MuSR & 756 & 0 & 0.000\% & 1 & 0.132\%\\
    \bottomrule
  \end{tabular}
  \caption{First-stage overlap audit between the TaskMixPGM source pool (309 tasks) and evaluation benchmarks. ``13-gram candidates'' are lexical hits requiring adjudication; a shared 13-gram may arise from repeated templates, mathematical expressions, or standard instructional language and does not by itself establish leakage.}
  \label{tab:overlap-audit-app}
\end{table}

The audit found no normalized exact duplicates in any evaluated benchmark. The 13-gram overlap rates are below 1\% for five of the six benchmarks. MATH-hard has a higher candidate rate (20.77\%); these cases require additional adjudication, so we do not treat every lexical hit as confirmed contamination.

\section{Additional Related Work}
\label{app:additional-related-work}

\textbf{Mixture optimisation and task affinity.}
Domain-reweighting methods choose mixture proportions with proxy models:
DoReMi \citep{xie2023doremi} reweights domains by minimax excess loss of a
small proxy, and RegMix \citep{liu2024regmix} regresses downstream loss on
proxy-trained mixtures. Both require training runs per candidate mixture
and target pretraining domains. Skill-It \citep{chen2023skillit}
estimates a graph of skill dependencies from training dynamics and uses it
to order and mix data, the closest conceptual neighbour of our affinity
graph. Task affinity grouping \citep{fifty2021efficiently} likewise
measures inter-task transfer from behaviour (lookahead loss changes), but
uses it to partition tasks for multi-task training rather than to weight
them. Our objective is a quadratic quality--diversity trade-off of the kind
determinantal point processes formalise \citep{kulesza2012determinantal}.
Unlike a DPP, it returns continuous proportions on the simplex, has a
closed-form convexity threshold $\tau_c$, and needs no proxy training
beyond the single-task adapters.

\section{Broader Impact}
\label{BroadImpact}

\methodprop{} changes \emph{which} data a model is fine-tuned on, not how the model is trained, so its effects follow from the mixtures it produces.

\textbf{Cheaper and more transparent mixture design.} Choosing an SFT mixture is usually done by hand or by expensive search over training runs. \methodprop{} computes a mixture once, from single-task adapters and without benchmark data, and reuses it for every budget and base model. Because the mixture follows from an explicit affinity matrix, practitioners can inspect why a task received weight, which tasks it is redundant with, and how the allocation changes with the trade-off parameter.

\textbf{Auditing training pools.} The affinity graph exposes clusters of near-duplicate tasks and tasks that the pool covers poorly. This can help curators find redundancy and gaps in large instruction collections before training.

\textbf{Limitations and risks.} \methodprop{} optimizes over the task pool it is given; it cannot add capabilities that the pool lacks, and it inherits the biases and quality issues of the underlying datasets. By favouring representative tasks it may also down-weight small or specialized tasks, including safety- or language-specific data, unless these are protected explicitly. We recommend inspecting the resulting mixture for such omissions before deployment.

\section{Code}
\label{section:code}
An anonymized version of our code is available at \url{https://anonymous.4open.science/r/taskpgm_ais-B555}.

\section{Dataset and Task Names}
\label{sec:datasets_tasks}
This section lists all datasets/tasks in our pool, along with their short task identifiers and one-line descriptions.

\input{Appendix/datasets_tasks_table}

%% file: Appendix/datasets_tasks_table.tex
{\scriptsize
% Allow line breaks at underscores in long task identifiers (prevents column overlap)
\begingroup\let\origunderscore\_\renewcommand{\_}{\origunderscore\allowbreak}
\setlength{\tabcolsep}{2pt}
\renewcommand{\arraystretch}{1.05}
\setlength{\LTpre}{0pt}
\setlength{\LTpost}{0pt}
\setlength{\LTleft}{0pt}
\setlength{\LTright}{0pt}
\begin{longtable}{@{}>{\raggedleft\arraybackslash}p{2.6em}@{\hspace{4pt}}>{\ttfamily\raggedright\arraybackslash}p{0.33\textwidth}@{}>{\raggedright\arraybackslash}p{\dimexpr\textwidth-2.6em-4pt-0.33\textwidth\relax}@{}}
\caption{Dataset and task names.}\label{tab:datasets_tasks}\\
\toprule
\textbf{\#} & \normalfont\textbf{Task} & \textbf{Description}\\
\midrule
\rowcolors{2}{rowgray}{white}
\endfirsthead

\multicolumn{3}{@{}l}{\small\textit{Table \thetable\ continued from previous page.}}\\
\toprule
\textbf{\#} & \normalfont\textbf{Task} & \textbf{Description}\\
\midrule
\rowcolors{2}{rowgray}{white}
\endhead

\midrule
\multicolumn{3}{r@{}}{\small\textit{Continued on next page.}}\\
\endfoot

\bottomrule
\endlastfoot

\rowcolor{gray!20}\multicolumn{3}{@{}l}{\textbf{COT}}\\
\addlinespace[2pt]
\rowcolors{2}{rowgray}{white}

1  & cot\_creak           & Chain-of-thought reasoning on the CREAK entity-based commonsense fact-verification dataset. \\
2  & cot\_creak\_ii       & Second prompt variant of cot\_creak with alternate instruction formatting. \\
3  & cot\_ecqa            & Chain-of-thought on ECQA: generates explanations for commonsense QA answer choices. \\
4  & cot\_ecqa\_ii        & Second prompt variant of cot\_ecqa. \\
5  & cot\_esnli           & Chain-of-thought on e-SNLI: produces natural language explanations for NLI labels. \\
6  & cot\_esnli\_ii       & Second prompt variant of cot\_esnli. \\
7  & cot\_gsm8k           & Chain-of-thought on GSM8K: step-by-step solutions to grade-school math word problems. \\
8  & cot\_gsm8k\_ii       & Second prompt variant of cot\_gsm8k. \\
9  & cot\_qasc            & Chain-of-thought on QASC: multi-hop fact-composition for science question answering. \\
10 & cot\_qasc\_ii        & Second prompt variant of cot\_qasc. \\
11 & cot\_sensemaking     & Chain-of-thought for sensemaking: determines which of two statements makes more sense. \\
12 & cot\_sensemaking\_ii & Second prompt variant of cot\_sensemaking. \\
13 & cot\_strategyqa      & Chain-of-thought on StrategyQA: yes/no questions requiring implicit multi-step reasoning. \\
14 & cot\_strategyqa\_ii  & Second prompt variant of cot\_strategyqa. \\
15 & stream\_aqua         & Streaming chain-of-thought on AQuA-RAT algebraic word problems with rationale generation. \\
16 & stream\_aqua\_ii     & Second prompt variant of stream\_aqua. \\
17 & stream\_qed          & Streaming chain-of-thought on QED: structured reasoning for open-domain QA. \\
18 & stream\_qed\_ii      & Second prompt variant of stream\_qed. \\
\addlinespace[2pt]
\rowcolor{gray!20}\multicolumn{3}{@{}l}{\textbf{FLAN2021}}\\
\addlinespace[2pt]
\rowcolors{2}{rowgray}{white}

19 & aeslc\_1.0.0                                  & Email subject line compression from body text (AESLC dataset). \\
20 & ag\_news\_subset\_1.0.0                        & News topic classification across four categories (AG News). \\
21 & ai2\_arc\_ARC\_Challenge\_1.0.0                & Challenging grade-school science multiple-choice QA (ARC Challenge). \\
22 & ai2\_arc\_ARC\_Easy\_1.0.0                     & Easier grade-school science multiple-choice QA (ARC Easy). \\
23 & anli\_r1\_0.1.0                               & Adversarial NLI round 1: entailment classification on hard examples. \\
24 & anli\_r2\_0.1.0                               & Adversarial NLI round 2: harder entailment classification. \\
25 & anli\_r3\_0.1.0                               & Adversarial NLI round 3: hardest adversarial entailment examples. \\
26 & bool\_q\_1.0.0                                & Yes/no question answering given a passage (BoolQ). \\
27 & cnn\_dailymail\_3.4.0                          & Abstractive news summarization (CNN/DailyMail). \\
28 & coqa\_1.0.0                                   & Conversational QA with follow-up questions over a passage (CoQA). \\
29 & cosmos\_qa\_1.0.0                             & Commonsense reading comprehension about everyday situations. \\
30 & definite\_pronoun\_resolution\_1.1.0           & Coreference: resolves which entity a definite pronoun refers to. \\
31 & drop\_2.0.0                                   & Discrete reasoning over paragraphs: arithmetic and counting QA. \\
32 & fix\_punct                                    & Punctuation restoration in text without punctuation marks. \\
33 & gem\_common\_gen\_1.1.0                        & Constrained text generation using a given set of common concepts. \\
34 & gem\_dart\_1.1.0                              & Data-to-text generation from structured RDF triples (DART). \\
35 & gem\_e2e\_nlg\_1.1.0                          & Restaurant description generation from structured meaning representations. \\
36 & gem\_web\_nlg\_en\_1.1.0                       & Natural language generation from RDF triples (WebNLG, English). \\
37 & gem\_wiki\_lingua\_english\_en\_1.1.0           & Cross-lingual summarization of WikiHow articles (English). \\
38 & gigaword\_1.2.0                               & Headline generation from the first sentence of newswire articles. \\
39 & glue\_cola\_2.0.0                             & Linguistic acceptability classification of English sentences (CoLA). \\
40 & glue\_mnli\_2.0.0                             & Multi-genre NLI: entailment, contradiction, or neutral (MNLI). \\
41 & glue\_mrpc\_2.0.0                             & Sentence-pair paraphrase detection (Microsoft Research Paraphrase Corpus). \\
42 & glue\_qnli\_2.0.0                             & Question-answer NLI: does the sentence contain the answer? \\
43 & glue\_qqp\_2.0.0                              & Quora question pair duplication detection. \\
44 & glue\_sst2\_2.0.0                             & Binary sentiment classification of movie reviews (SST-2). \\
45 & glue\_stsb\_2.0.0                             & Semantic textual similarity scored on a 1--5 scale (STS-B). \\
46 & glue\_wnli\_2.0.0                             & Coreference-based NLI derived from Winograd schemas (WNLI). \\
47 & hellaswag\_1.1.0                              & Commonsense NLI: selecting the most plausible continuation of a sentence. \\
48 & huggingface\_xsum                             & Extreme single-sentence news summarization (XSum). \\
49 & imdb\_reviews\_plain\_text\_1.0.0              & Binary sentiment classification of IMDb movie reviews. \\
50 & lambada\_1.0.0                                & Predict the final word of a passage requiring long-range context. \\
51 & math\_dataset\_algebra\_\_linear\_1d\_1.0.0    & Solve one-variable linear algebra equations (DeepMind Math). \\
52 & multi\_news\_1.0.0                            & Multi-document news summarization (Multi-News). \\
53 & natural\_questions\_open\_1.0.0               & Open-domain QA from Google search queries (Natural Questions). \\
54 & newsroom\_1.0.0                               & News article summarization with diverse summary types (Newsroom). \\
55 & openbookqa\_0.1.0                             & Open-book science QA requiring retrieval of core science facts. \\
56 & opinion\_abstracts\_idebate                   & Summarizing debate arguments into a single opinion abstract. \\
57 & opinion\_abstracts\_rotten\_tomatoes           & Summarizing critic reviews into a consensus opinion abstract. \\
58 & para\_crawl\_enes                             & English--Spanish parallel sentence translation (ParaCrawl). \\
59 & paws\_wiki\_1.1.0                             & Paraphrase detection with lexical overlap adversaries (PAWS-Wiki). \\
60 & piqa\_1.0.0                                   & Physical intuition QA: choosing the correct way to accomplish a task. \\
61 & quac\_1.0.0                                   & Question answering in context: QA over Wikipedia sections (QuAC). \\
62 & samsum\_1.0.0                                 & Abstractive summarization of messenger-style dialogues (SAMSum). \\
63 & sentiment140\_1.0.0                           & Binary sentiment classification of tweets (Sentiment140). \\
64 & snli\_1.1.0                                   & Sentence-pair NLI on image caption premises (Stanford NLI). \\
65 & squad\_v1.1\_3.0.0                            & Extractive reading comprehension on Wikipedia passages (SQuAD v1.1). \\
66 & squad\_v2.0\_3.0.0                            & Extractive QA with unanswerable questions (SQuAD v2.0). \\
67 & story\_cloze\_2016\_1.0.0                     & Commonsense reasoning: select the correct ending for a short story. \\
68 & super\_glue\_cb\_1.0.2                        & Three-way NLI on CommitmentBank short premises (CB). \\
69 & super\_glue\_copa\_1.0.2                      & Causal reasoning: choose the cause or effect of a given premise (COPA). \\
70 & super\_glue\_multirc\_1.0.2                   & Multi-sentence reading comprehension with binary answer validation. \\
71 & super\_glue\_record\_1.0.2                    & Cloze-style QA over news with named-entity answer candidates (ReCoRD). \\
72 & super\_glue\_rte\_1.0.2                       & Recognizing textual entailment (two-class, SuperGLUE RTE). \\
73 & super\_glue\_wic\_1.0.2                       & Word-in-context: does a polysemous word share meaning in two sentences? \\
74 & super\_glue\_wsc.fixed\_1.0.2                 & Winograd Schema Challenge coreference resolution (fixed split). \\
75 & trec\_1.0.0                                   & Question type classification into six coarse categories (TREC). \\
76 & trivia\_qa\_rc\_1.1.0                         & Trivia QA with supporting evidence from Wikipedia/web (reading comprehension). \\
77 & true\_case                                    & Restores true casing to all-lowercase input text. \\
78 & unified\_qa\_science\_inst                    & Unified QA format applied to science exam questions with instructions. \\
79 & winogrande\_1.1.0                             & Large-scale Winograd Schema pronoun resolution at various difficulty levels. \\
80 & word\_segment                                 & Word segmentation: insert spaces into a string of concatenated words. \\
81 & yelp\_polarity\_reviews\_0.2.0                & Binary sentiment classification of Yelp business reviews. \\
\addlinespace[2pt]
\rowcolor{gray!20}\multicolumn{3}{@{}l}{\textbf{SGLUE}}\\
\addlinespace[2pt]
\rowcolors{2}{rowgray}{white}

82  & axb        & Broad-coverage NLI diagnostic test on linguistic phenomena (AX-b). \\
83  & axg        & Gender-bias NLI diagnostic test (AX-g). \\
84  & boolq      & Yes/no QA given a passage (BoolQ, SuperGLUE split). \\
85  & cola       & Linguistic acceptability classification (CoLA, SuperGLUE split). \\
86  & copa       & Causal commonsense reasoning: cause or effect selection (COPA). \\
87  & cr         & Binary sentiment classification of customer product reviews (CR). \\
88  & mpqa       & Opinion polarity detection at the phrase level (MPQA). \\
89  & mr         & Binary sentiment classification of movie reviews (MR). \\
90  & mrpc       & Sentence-pair paraphrase detection (MRPC, SuperGLUE split). \\
91  & pawsx      & Cross-lingual paraphrase detection across multiple languages (PAWS-X). \\
92  & qnli       & Question-answer NLI (QNLI, SuperGLUE split). \\
93  & qqp        & Quora question-pair duplication detection (QQP, SuperGLUE split). \\
94  & rte        & Recognizing textual entailment (RTE, SuperGLUE split). \\
95  & sst2       & Binary sentiment classification of movie review sentences (SST-2). \\
96  & subj       & Subjectivity classification: is the sentence subjective or objective? \\
97  & wnli       & Winograd NLI coreference-based entailment (WNLI). \\
98  & xwinograd  & Cross-lingual Winograd Schema pronoun resolution. \\
\addlinespace[2pt]
\rowcolor{gray!20}\multicolumn{3}{@{}l}{\textbf{T0}}\\
\addlinespace[2pt]
\rowcolors{2}{rowgray}{white}

99  & adversarial\_qa\_dbert\_answer\_the\_following\_q      & Adversarial QA with DBERT backbone: answer a question given a passage. \\
100 & adversarial\_qa\_dbert\_based\_on                      & Adversarial QA (DBERT): answer framed as ``based on the passage\ldots'' \\
101 & adversarial\_qa\_dbert\_generate\_question             & Generate a question whose answer is the given span (DBERT). \\
102 & adversarial\_qa\_dbert\_question\_context\_answer      & QA formatted with explicit question, context, and answer fields (DBERT). \\
103 & adversarial\_qa\_dbert\_tell\_what\_it\_is             & Adversarial QA (DBERT): identify what the answer refers to. \\
104 & adversarial\_qa\_dbidaf\_answer\_the\_following\_q     & Same as above but using the BiDAF adversarial backbone model. \\
105 & adversarial\_qa\_dbidaf\_based\_on                    & BiDAF adversarial QA with ``based on'' prompt framing. \\
106 & adversarial\_qa\_dbidaf\_generate\_question            & Generate a question from an answer span (BiDAF adversarial). \\
107 & adversarial\_qa\_dbidaf\_question\_context\_answer     & Explicit QCA format (BiDAF adversarial). \\
108 & adversarial\_qa\_dbidaf\_tell\_what\_it\_is            & Identify the referent of the answer span (BiDAF adversarial). \\
109 & adversarial\_qa\_droberta\_answer\_the\_following\_q   & Same as above but using the RoBERTa adversarial backbone model. \\
110 & adversarial\_qa\_droberta\_based\_on                  & RoBERTa adversarial QA with ``based on'' framing. \\
111 & adversarial\_qa\_droberta\_generate\_question          & Generate a question from an answer span (RoBERTa adversarial). \\
112 & adversarial\_qa\_droberta\_question\_context\_answer   & Explicit QCA format (RoBERTa adversarial). \\
113 & adversarial\_qa\_droberta\_tell\_what\_it\_is          & Identify the referent of the answer span (RoBERTa adversarial). \\
114 & amazon\_polarity\_Is\_this\_product\_review\_positive  & Binary product review sentiment: is the review positive? \\
115 & amazon\_polarity\_Is\_this\_review                    & Binary sentiment classification of Amazon product reviews. \\
116 & amazon\_polarity\_Is\_this\_review\_negative           & Binary product review sentiment: is the review negative? \\
117 & amazon\_polarity\_User\_recommend\_this\_product       & Predict whether a user would recommend the product based on review. \\
118 & amazon\_polarity\_convey\_negative\_or\_positive\_sentiment & Classify the sentiment conveyed in an Amazon review. \\
119 & amazon\_polarity\_flattering\_or\_not                 & Determine whether an Amazon review is flattering or not. \\
120 & amazon\_polarity\_negative\_or\_positive\_tone         & Classify the overall tone of an Amazon product review. \\
121 & amazon\_polarity\_user\_satisfied                     & Predict whether the reviewer is satisfied with the product. \\
122 & amazon\_polarity\_would\_you\_buy                     & Predict purchase intent from a product review. \\
123 & app\_reviews\_categorize\_rating\_using\_review        & Predict a star rating from a mobile app review. \\
124 & app\_reviews\_convert\_to\_rating                     & Convert the sentiment of an app review to a numeric rating. \\
125 & app\_reviews\_convert\_to\_star\_rating               & Convert an app review to an explicit star rating label. \\
126 & app\_reviews\_generate\_review                        & Generate a plausible app review text given a rating. \\
127 & cos\_e\_v1.11\_aligned\_with\_common\_sense           & Commonsense QA: select the answer aligned with common sense (CoS-E). \\
128 & cos\_e\_v1.11\_description\_question\_option\_id      & CoS-E formatted with description, question, and option ID. \\
129 & cos\_e\_v1.11\_description\_question\_option\_text    & CoS-E formatted with description, question, and option text. \\
130 & cos\_e\_v1.11\_explain\_why\_human                    & Generate a human-style explanation for why a commonsense answer is correct. \\
131 & cos\_e\_v1.11\_generate\_explanation\_given\_text     & Generate a rationale given the question and correct answer. \\
132 & cos\_e\_v1.11\_i\_think                              & CoS-E with first-person ``I think\ldots'' style explanation framing. \\
133 & cos\_e\_v1.11\_question\_description\_option\_id      & CoS-E formatted with question, description, and option ID. \\
134 & cos\_e\_v1.11\_question\_description\_option\_text    & CoS-E formatted with question, description, and option text. \\
135 & cos\_e\_v1.11\_question\_option\_description\_id      & CoS-E formatted with question, option, and description (ID output). \\
136 & cos\_e\_v1.11\_question\_option\_description\_text    & CoS-E formatted with question, option, and description (text output). \\
137 & cos\_e\_v1.11\_rationale                             & Produce the rationale for a CoS-E commonsense answer choice. \\
138 & dbpedia\_14\_given\_a\_choice\_of\_categories\_       & DBpedia topic classification given a list of category choices. \\
139 & dbpedia\_14\_given\_a\_list\_of\_category\_what\_does\_the\_title\_belong\_to & Predict DBpedia category from article title given a category list. \\
140 & dbpedia\_14\_given\_list\_what\_category\_does\_the\_paragraph\_belong\_to & Predict DBpedia category from a paragraph given a category list. \\
141 & dbpedia\_14\_pick\_one\_category\_for\_the\_following\_text & Pick one DBpedia ontology category for the provided text. \\
142 & dream\_answer\_to\_dialogue                          & Select the answer to a question about a dialogue (DREAM). \\
143 & dream\_baseline                                      & Baseline QA on the DREAM dialogue comprehension dataset. \\
144 & dream\_generate\_first\_utterance                    & Generate the opening utterance of a dialogue given context. \\
145 & dream\_generate\_last\_utterance                     & Generate the closing utterance of a dialogue given context. \\
146 & dream\_read\_the\_following\_conversation\_and\_answer\_the\_question & Standard reading comprehension on DREAM dialogues. \\
147 & duorc\_ParaphraseRC\_answer\_question                 & Answer a paraphrased question about a movie plot (DuoRC). \\
148 & duorc\_ParaphraseRC\_build\_story\_around\_qa         & Generate a story that supports a given QA pair (DuoRC Paraphrase). \\
149 & duorc\_ParaphraseRC\_decide\_worth\_it                & Decide whether a QA pair is worth including (DuoRC Paraphrase). \\
150 & duorc\_ParaphraseRC\_extract\_answer                  & Extract the answer span from a paraphrased plot (DuoRC). \\
151 & duorc\_ParaphraseRC\_generate\_question               & Generate a question from a plot and answer (DuoRC Paraphrase). \\
152 & duorc\_ParaphraseRC\_generate\_question\_by\_answer   & Generate a question conditioned solely on the answer (DuoRC Paraphrase). \\
153 & duorc\_ParaphraseRC\_movie\_director                  & Predict the movie director from a plot description (DuoRC Paraphrase). \\
154 & duorc\_ParaphraseRC\_question\_answering              & Standard QA over paraphrased movie plots (DuoRC). \\
155 & duorc\_ParaphraseRC\_title\_generation                & Generate a movie title from its plot (DuoRC Paraphrase). \\
156 & duorc\_SelfRC\_answer\_question                       & Answer a question about a movie plot using the same-source RC (DuoRC SelfRC). \\
157 & duorc\_SelfRC\_build\_story\_around\_qa               & Generate a story supporting a QA pair (DuoRC SelfRC). \\
158 & duorc\_SelfRC\_decide\_worth\_it                      & Decide whether a QA pair is worth including (DuoRC SelfRC). \\
159 & duorc\_SelfRC\_extract\_answer                        & Extract the answer span from the same-source plot (DuoRC SelfRC). \\
160 & duorc\_SelfRC\_generate\_question                     & Generate a question from a plot and answer (DuoRC SelfRC). \\
161 & duorc\_SelfRC\_generate\_question\_by\_answer         & Generate a question conditioned solely on the answer (DuoRC SelfRC). \\
162 & duorc\_SelfRC\_movie\_director                        & Predict the movie director from a same-source plot (DuoRC SelfRC). \\
163 & duorc\_SelfRC\_question\_answering                    & Standard QA over same-source movie plots (DuoRC SelfRC). \\
164 & duorc\_SelfRC\_title\_generation                      & Generate a movie title from its same-source plot (DuoRC SelfRC). \\
165 & kilt\_tasks\_hotpotqa\_combining\_facts               & Multi-hop QA by combining two supporting facts (HotpotQA via KILT). \\
166 & kilt\_tasks\_hotpotqa\_complex\_question              & Answer complex multi-hop questions using Wikipedia (HotpotQA/KILT). \\
167 & kilt\_tasks\_hotpotqa\_final\_exam                    & Multi-hop QA framed as a final-exam question (HotpotQA/KILT). \\
168 & kilt\_tasks\_hotpotqa\_formulate                      & Formulate an answer from two supporting Wikipedia passages. \\
169 & kilt\_tasks\_hotpotqa\_straighforward\_qa             & Straightforward QA format over HotpotQA (KILT). \\
170 & qasc\_is\_correct\_1                                 & Verify whether a given fact combination is correct (QASC, format 1). \\
171 & qasc\_is\_correct\_2                                 & Verify whether a given fact combination is correct (QASC, format 2). \\
172 & qasc\_qa\_with\_combined\_facts\_1                    & Answer a science question using one pre-combined supporting fact. \\
173 & qasc\_qa\_with\_separated\_facts\_1                   & Answer a science question using two separated supporting facts (format 1). \\
174 & qasc\_qa\_with\_separated\_facts\_2                   & Answer a science question using two separated supporting facts (format 2). \\
175 & qasc\_qa\_with\_separated\_facts\_3                   & Answer a science question using two separated supporting facts (format 3). \\
176 & qasc\_qa\_with\_separated\_facts\_4                   & Answer a science question using two separated supporting facts (format 4). \\
177 & qasc\_qa\_with\_separated\_facts\_5                   & Answer a science question using two separated supporting facts (format 5). \\
178 & quail\_context\_description\_question\_answer\_id     & Reading comprehension with context, description, question; answer as ID (QUAIL). \\
179 & quail\_context\_description\_question\_answer\_text   & Same as above but answer as text (QUAIL). \\
180 & quail\_context\_description\_question\_text           & QUAIL reading comprehension: output the answer text given context and description. \\
181 & quail\_context\_question\_answer\_description\_id     & QUAIL format with context, question, answer, and description (ID output). \\
182 & quail\_context\_question\_answer\_description\_text   & QUAIL format with context, question, answer, and description (text output). \\
183 & quail\_context\_question\_description\_answer\_id     & QUAIL: context, question, description; answer as option ID. \\
184 & quail\_context\_question\_description\_answer\_text   & QUAIL: context, question, description; answer as option text. \\
185 & quail\_context\_question\_description\_text           & QUAIL: generate answer text from context, question, and description. \\
186 & quail\_description\_context\_question\_answer\_id     & QUAIL with description first, then context and question; answer as ID. \\
187 & quail\_description\_context\_question\_answer\_text   & QUAIL with description first; answer as text. \\
188 & quail\_description\_context\_question\_text           & QUAIL with description-first ordering; free-text answer. \\
189 & quail\_no\_prompt\_id                                & QUAIL reading comprehension without an explicit prompt template; answer as ID. \\
190 & quail\_no\_prompt\_text                              & QUAIL reading comprehension without a prompt template; answer as text. \\
191 & quarel\_choose\_between                              & Physical comparison QA: choose between two objects based on a scenario (QuaRel). \\
192 & quarel\_do\_not\_use                                 & QuaRel task variant instructing model not to use certain attributes. \\
193 & quarel\_heres\_a\_story                              & QuaRel framed as a story comprehension question. \\
194 & quarel\_logic\_test                                  & QuaRel framed as a logic test over qualitative relationships. \\
195 & quarel\_testing\_students                            & QuaRel framed as a student-testing scenario. \\
196 & quartz\_answer\_question\_based\_on                  & Answer a science question based on a provided principle (QuaRTz). \\
197 & quartz\_answer\_question\_below                      & QuaRTz QA with a ``answer the question below'' instruction framing. \\
198 & quartz\_given\_the\_fact\_answer\_the\_q             & QuaRTz: given a qualitative fact, answer the associated question. \\
199 & quartz\_having\_read\_above\_passage                 & QuaRTz QA framed as comprehension after reading the passage above. \\
200 & quartz\_paragraph\_question\_plain\_concat           & Plain concatenation of paragraph and question for QuaRTz QA. \\
201 & quartz\_read\_passage\_below\_choose                 & QuaRTz: read a passage and choose the correct answer. \\
202 & quartz\_use\_info\_from\_paragraph\_question         & Answer a QuaRTz question using information extracted from the paragraph. \\
203 & quartz\_use\_info\_from\_question\_paragraph         & Answer a QuaRTz question; question presented before the paragraph. \\
204 & quoref\_Answer\_Friend\_Question                     & Answer a coreference-dependent question as if asked by a friend (Quoref). \\
205 & quoref\_Answer\_Question\_Given\_Context             & Standard coreference-aware QA given a context passage (Quoref). \\
206 & quoref\_Answer\_Test                                 & Quoref QA framed as an answer test. \\
207 & quoref\_Context\_Contains\_Answer                    & Verify that the context contains the answer to a coreference question. \\
208 & quoref\_Find\_Answer                                 & Find the answer span in a passage for a coreference-dependent question. \\
209 & quoref\_Found\_Context\_Online                       & Quoref QA framed as finding context online to answer a question. \\
210 & quoref\_Given\_Context\_Answer\_Question             & Quoref: given context, answer the coreference-dependent question. \\
211 & quoref\_Guess\_Answer                                & Quoref framed as guessing the answer to a coreference question. \\
212 & quoref\_Guess\_Title\_For\_Context                   & Predict the title of the passage given its coreference-rich content. \\
213 & quoref\_Read\_And\_Extract\_                         & Read a passage and extract the answer to a coreference question. \\
214 & quoref\_What\_Is\_The\_Answer                        & Simple ``what is the answer'' prompt for Quoref QA. \\
215 & race\_high\_Is\_this\_the\_right\_answer             & Verify if a given answer is correct for a high-school reading question (RACE). \\
216 & race\_high\_Read\_the\_article\_and\_answer\_the\_question\_no\_option\_ & High-school RACE QA without explicit answer options. \\
217 & race\_high\_Select\_the\_best\_answer                & High-school RACE multiple-choice reading comprehension. \\
218 & race\_high\_Select\_the\_best\_answer\_generate\_span\_ & RACE (high): select the best answer and generate the answer span. \\
219 & race\_high\_Select\_the\_best\_answer\_no\_instructions\_ & RACE (high) multiple choice without explicit instructions. \\
220 & race\_high\_Taking\_a\_test                          & RACE (high) framed as a student taking an exam. \\
221 & race\_high\_Write\_a\_multi\_choice\_question\_for\_the\_following\_article & Generate a multiple-choice question from a high-school article. \\
222 & race\_high\_Write\_a\_multi\_choice\_question\_options\_given\_ & Generate a multiple-choice question with pre-given options (RACE high). \\
223 & race\_middle\_Is\_this\_the\_right\_answer            & Verify if a given answer is correct for a middle-school reading question (RACE). \\
224 & race\_middle\_Read\_the\_article\_and\_answer\_the\_question\_no\_option\_ & Middle-school RACE QA without explicit answer options. \\
225 & race\_middle\_Select\_the\_best\_answer              & Middle-school RACE multiple-choice reading comprehension. \\
226 & race\_middle\_Select\_the\_best\_answer\_generate\_span\_ & RACE (middle): select the best answer and generate the span. \\
227 & race\_middle\_Select\_the\_best\_answer\_no\_instructions\_ & RACE (middle) multiple choice without explicit instructions. \\
228 & race\_middle\_Taking\_a\_test                        & RACE (middle) framed as a student taking an exam. \\
229 & race\_middle\_Write\_a\_multi\_choice\_question\_for\_the\_following\_article & Generate a multiple-choice question from a middle-school article. \\
230 & race\_middle\_Write\_a\_multi\_choice\_question\_options\_given\_ & Generate a multiple-choice question with pre-given options (RACE middle). \\
231 & ropes\_background\_new\_situation\_answer            & QA using background knowledge applied to a new situation (ROPES). \\
232 & ropes\_background\_situation\_middle                 & ROPES QA with background and situation interleaved. \\
233 & ropes\_given\_background\_situation                  & ROPES: answer given background knowledge and a specific situation. \\
234 & ropes\_new\_situation\_background\_answer            & ROPES with new situation presented before the background. \\
235 & ropes\_plain\_background\_situation                  & Plain concatenation of background and situation for ROPES QA. \\
236 & ropes\_plain\_bottom\_hint                           & ROPES with a hint placed at the bottom of the prompt. \\
237 & ropes\_plain\_no\_background                        & ROPES QA without providing the background passage. \\
238 & ropes\_prompt\_beginning                             & ROPES with the prompt instruction placed at the beginning. \\
239 & ropes\_prompt\_bottom\_hint\_beginning               & ROPES with both a bottom hint and a beginning prompt instruction. \\
240 & ropes\_prompt\_bottom\_no\_hint                      & ROPES with prompt at the bottom and no background hint. \\
241 & ropes\_prompt\_mix                                   & ROPES with a mixed prompt ordering strategy. \\
242 & ropes\_read\_background\_situation                   & ROPES: read the background and situation, then answer the question. \\
243 & sciq\_Direct\_Question                               & Answer a science multiple-choice question directly (SciQ). \\
244 & sciq\_Direct\_Question\_Closed\_Book\_               & Answer a SciQ science question without the supporting passage. \\
245 & sciq\_Multiple\_Choice                               & Standard multiple-choice science QA (SciQ). \\
246 & sciq\_Multiple\_Choice\_Closed\_Book\_               & Closed-book multiple-choice science QA (SciQ). \\
247 & sciq\_Multiple\_Choice\_Question\_First              & SciQ with the question presented before the answer options. \\
248 & social\_i\_qa\_Check\_if\_a\_random\_answer\_is\_valid\_or\_not & Validate whether a random answer is correct for a social scenario (SocialIQA). \\
249 & social\_i\_qa\_Generate\_answer                      & Generate an answer to a social commonsense question (SocialIQA). \\
250 & social\_i\_qa\_Generate\_the\_question\_from\_the\_answer & Generate the question that matches a given answer (SocialIQA). \\
251 & social\_i\_qa\_I\_was\_wondering                    & SocialIQA framed with an ``I was wondering\ldots'' conversational prompt. \\
252 & social\_i\_qa\_Show\_choices\_and\_generate\_answer  & SocialIQA: present all choices and generate the correct answer. \\
253 & social\_i\_qa\_Show\_choices\_and\_generate\_index   & SocialIQA: present all choices and output the index of the correct one. \\
254 & web\_questions\_get\_the\_answer                     & Retrieve a short factual answer to a web search question (WebQuestions). \\
255 & web\_questions\_potential\_correct\_answer           & Generate a potential correct answer for a WebQuestions query. \\
256 & web\_questions\_question\_answer                     & Standard QA format on the WebQuestions factoid dataset. \\
257 & web\_questions\_short\_general\_knowledge\_q         & Short general-knowledge QA from WebQuestions. \\
258 & web\_questions\_whats\_the\_answer                   & ``What's the answer'' prompt style for WebQuestions. \\
259 & wiki\_bio\_comprehension                             & Comprehension questions about a person based on their Wikipedia biography. \\
260 & wiki\_bio\_guess\_person                             & Identify the person described in a Wikipedia biography snippet. \\
261 & wiki\_bio\_key\_content                              & Extract the key content from a Wikipedia biography. \\
262 & wiki\_bio\_what\_content                             & Describe what content a Wikipedia biography covers. \\
263 & wiki\_bio\_who                                       & Answer ``who is this'' given a Wikipedia biography excerpt. \\
264 & wiki\_hop\_original\_choose\_best\_object\_affirmative\_1 & Multi-hop: choose the best object in an affirmative frame (WikiHop, format 1). \\
265 & wiki\_hop\_original\_choose\_best\_object\_affirmative\_2 & Multi-hop: choose the best object in an affirmative frame (WikiHop, format 2). \\
266 & wiki\_hop\_original\_choose\_best\_object\_affirmative\_3 & Multi-hop: choose the best object in an affirmative frame (WikiHop, format 3). \\
267 & wiki\_hop\_original\_choose\_best\_object\_interrogative\_1 & Multi-hop: choose the best object in an interrogative frame (WikiHop, format 1). \\
268 & wiki\_hop\_original\_choose\_best\_object\_interrogative\_2 & Multi-hop: choose the best object in an interrogative frame (WikiHop, format 2). \\
269 & wiki\_hop\_original\_explain\_relation               & Explain the relationship between two entities using Wikipedia (WikiHop). \\
270 & wiki\_hop\_original\_generate\_object                & Generate the object entity given a subject and relation (WikiHop). \\
271 & wiki\_hop\_original\_generate\_subject               & Generate the subject entity given an object and relation (WikiHop). \\
272 & wiki\_hop\_original\_generate\_subject\_and\_object  & Generate both subject and object entities for a relation (WikiHop). \\
273 & wiki\_qa\_Decide\_good\_answer                       & Decide whether a candidate answer is a good answer (WikiQA). \\
274 & wiki\_qa\_Direct\_Answer\_to\_Question               & Provide a direct answer to a WikiQA factoid question. \\
275 & wiki\_qa\_Generate\_Question\_from\_Topic            & Generate a question given a Wikipedia topic (WikiQA). \\
276 & wiki\_qa\_Is\_This\_True\_                           & Verify whether a candidate answer is true for a WikiQA question. \\
277 & wiki\_qa\_Jeopardy\_style                            & Answer a WikiQA question in Jeopardy-style (answer as a question). \\
278 & wiki\_qa\_Topic\_Prediction\_Answer\_Only            & Predict the Wikipedia topic from the answer alone (WikiQA). \\
279 & wiki\_qa\_Topic\_Prediction\_Question\_Only          & Predict the Wikipedia topic from the question alone (WikiQA). \\
280 & wiki\_qa\_Topic\_Prediction\_Question\_and\_Answer\_Pair & Predict the Wikipedia topic given both the question and answer (WikiQA). \\
281 & wiki\_qa\_automatic\_system                          & WikiQA formatted for automatic IR-based answer selection. \\
282 & wiki\_qa\_exercise                                   & WikiQA framed as an exercise for answer selection. \\
283 & wiki\_qa\_found\_on\_google                          & WikiQA framed as retrieving an answer found on Google. \\
284 & wiqa\_does\_the\_supposed\_perturbation\_have\_an\_effect & Classify whether a described perturbation affects a process (WIQA). \\
285 & wiqa\_effect\_with\_label\_answer                    & Predict the effect of a perturbation on a process; output a label (WIQA). \\
286 & wiqa\_effect\_with\_string\_answer                   & Predict the effect of a perturbation on a process; output a string (WIQA). \\
287 & wiqa\_what\_is\_the\_final\_step\_of\_the\_following\_process & Predict the final step of a described scientific process (WIQA). \\
288 & wiqa\_what\_is\_the\_missing\_first\_step             & Identify the missing first step in a partially described process (WIQA). \\
289 & wiqa\_what\_might\_be\_the\_first\_step\_of\_the\_process & Generate a plausible first step for a process (WIQA). \\
290 & wiqa\_what\_might\_be\_the\_last\_step\_of\_the\_process & Generate a plausible last step for a process (WIQA). \\
291 & wiqa\_which\_of\_the\_following\_is\_the\_supposed\_perturbation & Identify which option is the perturbation in a process description (WIQA). \\
\addlinespace[2pt]
\rowcolor{gray!20}\multicolumn{3}{@{}l}{\textbf{TULU}}\\
\addlinespace[2pt]
\rowcolors{2}{rowgray}{white}

292 & aya              & Multilingual instruction-following data from the Aya initiative. \\
293 & coconot          & Instruction-following training data with conflict/contradiction handling. \\
294 & evol\_codealpaca  & Evolved code instruction data built on Code Alpaca via instruction evolution. \\
295 & flanv2           & Instruction tuning data from the FLAN v2 collection. \\
296 & no\_robots        & High-quality human-written instruction-response pairs (No Robots dataset). \\
297 & numina\_math\_tir & Mathematical reasoning with tool-integrated reasoning steps (NuminaMath-TIR). \\
298 & openassistant\_guanaco & Human preference data from OpenAssistant filtered to the Guanaco subset. \\
299 & sciriff          & Scientific instruction-following data across NLP, biology, and chemistry tasks. \\
300 & table\_gpt       & Instruction data for table understanding and manipulation tasks (TableGPT). \\
301 & tulu\_hardcoded   & Hardcoded identity and safety responses for the Tulu model. \\
302 & tulu\_personas\_algebra & Persona-conditioned algebra problem-solving instruction data. \\
303 & tulu\_personas\_code & Persona-conditioned code generation and debugging instruction data. \\
304 & tulu\_personas\_if & Persona-conditioned instruction-following data. \\
305 & tulu\_personas\_math & Persona-conditioned general math instruction data. \\
306 & tulu\_personas\_math\_grade & Persona-conditioned grade-school math instruction data. \\
307 & wildchat         & Real user–chatbot conversations from the WildChat corpus. \\
308 & wildguardmix     & Safety and refusal training data mixing WildGuard examples. \\
309 & wildjailbreak    & Adversarial jailbreak attack and defense instruction data (WildJailbreak). \\
\end{longtable}
\endgroup
}

%% file: ref.bib
@article{sutawika2024eleutherai,
  title={Eleutherai/lm-evaluation-harness: v0. 4.9},
  author={Sutawika, Lintang and Schoelkopf, Hailey and Gao, Leo and Abbasi, Baber and Biderman, Stella and Tow, Jonathan and Lovering, Charles and Phang, Jason and Thite, Anish and Wang, Thomas and others},
  journal={Zenodo},
  year={2024}
}

@inproceedings{wu2005analysis,
  title={An analysis of transformation on non-positive semidefinite similarity matrix for kernel machines},
  author={Wu, Gang and Chang, Edward Y and Zhang, Zhihua},
  booktitle={Proceedings of the 22nd international conference on machine learning},
  volume={8},
  year={2005},
  organization={Citeseer Cham}
}

@inproceedings{chen2009learning,
  title={Learning kernels from indefinite similarities},
  author={Chen, Yihua and Gupta, Maya R and Recht, Benjamin},
  booktitle={Proceedings of the 26th Annual International Conference on Machine Learning},
  pages={145--152},
  year={2009}
}

@book{kindermann1980markov,
  title={Markov random fields and their applications},
  author={Kindermann, Ross and Snell, J Laurie},
  volume={1},
  year={1980},
  publisher={American Mathematical Society}
}

@inproceedings{duchi2008efficient,
  title={Efficient projections onto the l 1-ball for learning in high dimensions},
  author={Duchi, John and Shalev-Shwartz, Shai and Singer, Yoram and Chandra, Tushar},
  booktitle={Proceedings of the 25th international conference on Machine learning},
  pages={272--279},
  year={2008}
}

@inproceedings{gurumoorthy2019efficient,
  title     = {Efficient Data Representation by Selecting Prototypes with Importance Weights},
  author    = {Gurumoorthy, Karthik S. and Dhurandhar, Amit and Cecchi, Guillermo and Aggarwal, Charu},
  booktitle = {IEEE International Conference on Data Mining (ICDM)},
  pages     = {260--269},
  year      = {2019},
  note      = {arXiv:1707.01212}
}

@article{elenberg2018restricted,
  title   = {Restricted Strong Convexity Implies Weak Submodularity},
  author  = {Elenberg, Ethan R. and Khanna, Rajiv and Dimakis, Alexandros G. and Negahban, Sahand},
  journal = {The Annals of Statistics},
  volume  = {46},
  number  = {6B},
  pages   = {3539--3568},
  year    = {2018}
}

@inproceedings{das2011submodular,
  title     = {Submodular meets Spectral: Greedy Algorithms for Subset Selection, Sparse Approximation and Dictionary Selection},
  author    = {Das, Abhimanyu and Kempe, David},
  booktitle = {Proceedings of the 28th International Conference on Machine Learning (ICML)},
  pages     = {1057--1064},
  year      = {2011}
}

@article{beck2009fast,
  title={A fast iterative shrinkage-thresholding algorithm for linear inverse problems},
  author={Beck, Amir and Teboulle, Marc},
  journal={SIAM journal on imaging sciences},
  volume={2},
  number={1},
  pages={183--202},
  year={2009},
  publisher={SIAM}
}

@article{das2018approximate,
  title={Approximate Submodularity and Its Applications: Subset Selection, Sparse Approximation and Dictionary Selection},
  author={Das, A. and Kempe, D.},
  journal={Journal of Machine Learning Research},
  volume={19},
  number={3},
  pages={1--34},
  year={2018}
}

@inproceedings{xie2023doremi,
  title     = {{DoReMi}: Optimizing Data Mixtures Speeds Up Language Model Pretraining},
  author    = {Xie, Sang Michael and Pham, Hieu and Dong, Xuanyi and Du, Nan and Liu, Hanxiao and Lu, Yifeng and Liang, Percy and Le, Quoc V. and Ma, Tengyu and Yu, Adams Wei},
  booktitle = {Advances in Neural Information Processing Systems},
  year      = {2023}
}

@inproceedings{chen2023skillit,
  title     = {Skill-it! {A} Data-Driven Skills Framework for Understanding and Training Language Models},
  author    = {Chen, Mayee F. and Roberts, Nicholas and Bhatia, Kush and Wang, Jue and Zhang, Ce and Sala, Frederic and R{\'e}, Christopher},
  booktitle = {Advances in Neural Information Processing Systems},
  year      = {2023}
}

@inproceedings{fifty2021efficiently,
  title     = {Efficiently Identifying Task Groupings for Multi-Task Learning},
  author    = {Fifty, Christopher and Amid, Ehsan and Zhao, Zhe and Yu, Tianhe and Anil, Rohan and Finn, Chelsea},
  booktitle = {Advances in Neural Information Processing Systems},
  year      = {2021}
}

@article{kulesza2012determinantal,
  title   = {Determinantal Point Processes for Machine Learning},
  author  = {Kulesza, Alex and Taskar, Ben},
  journal = {Foundations and Trends in Machine Learning},
  volume  = {5},
  number  = {2--3},
  pages   = {123--286},
  year    = {2012}
}

@article{church1990word,
  title   = {Word Association Norms, Mutual Information, and Lexicography},
  author  = {Church, Kenneth Ward and Hanks, Patrick},
  journal = {Computational Linguistics},
  volume  = {16},
  number  = {1},
  pages   = {22--29},
  year    = {1990}
}

@book{horn2012matrix,
  title={Matrix analysis},
  author={Horn, Roger A and Johnson, Charles R},
  year={2012},
  publisher={Cambridge university press}
}

@article{nemhauser1978analysis,
  title={An Analysis of Approximations for Maximizing Submodular Set Functions---I},
  author={Nemhauser, G.L. and Wolsey, L.A. and Fisher, M.L.},
  journal={Mathematical Programming},
  volume={14},
  pages={265--294},
  year={1978}
}

@article{condat2016fast,
  title={Fast projection onto the simplex and the l 1 ball},
  author={Condat, Laurent},
  journal={Mathematical Programming},
  volume={158},
  number={1},
  pages={575--585},
  year={2016},
  publisher={Springer}
}

@article{wright2022optimization,
  title={Optimization for data analysis},
  author={Wright, Stephen J and Recht, Benjamin},
  journal={(No Title)},
  year={2022}
}

@article{beck2003mirror,
  title={Mirror descent and nonlinear projected subgradient methods for convex optimization},
  author={Beck, Amir and Teboulle, Marc},
  journal={Operations Research Letters},
  volume={31},
  number={3},
  pages={167--175},
  year={2003},
  publisher={Elsevier}
}

@inproceedings{renduchintala-etal-2024-smart,
    title = "{SMART}: Submodular Data Mixture Strategy for Instruction Tuning",
    author = "Renduchintala, H S V N S Kowndinya  and
      Bhatia, Sumit  and
      Ramakrishnan, Ganesh",
    editor = "Ku, Lun-Wei  and
      Martins, Andre  and
      Srikumar, Vivek",
    booktitle = "Findings of the Association for Computational Linguistics: ACL 2024",
    month = aug,
    year = "2024",
    address = "Bangkok, Thailand",
    publisher = "Association for Computational Linguistics",
    url = "https://aclanthology.org/2024.findings-acl.766/",
    doi = "10.18653/v1/2024.findings-acl.766",
    pages = "12916--12934"
}

@inproceedings{
liu2024tsds,
title={{TSDS}: Data Selection for Task-Specific Model Finetuning},
author={Zifan Liu and Amin Karbasi and Theodoros Rekatsinas},
booktitle={The Thirty-eighth Annual Conference on Neural Information Processing Systems},
year={2024},
url={https://openreview.net/forum?id=wjbTHLUSzU}
}

@inproceedings{zhang2025staff,
  title={Staff: Speculative coreset selection for task-specific fine-tuning},
  author={Zhang, Xiaoyu and Zhai, Juan and Ma, Shiqing and Shen, Chao and Li, Tianlin and Jiang, Weipeng and Liu, Yang},
  booktitle={The thirteenth international conference on learning representations},
  year={2025}
}

@inproceedings{
agarwal2025delift,
title={{DELIFT}: Data Efficient Language model Instruction Fine-Tuning},
author={Ishika Agarwal and Krishnateja Killamsetty and Lucian Popa and Marina Danilevsky},
booktitle={The Thirteenth International Conference on Learning Representations},
year={2025},
url={https://openreview.net/forum?id=Fty0wTcemV}
}

@inproceedings{
xia2024less,
title={{LESS}: Selecting Influential Data for Targeted Instruction Tuning},
author={Mengzhou Xia and Sadhika Malladi and Suchin Gururangan and Sanjeev Arora and Danqi Chen},
booktitle={Forty-first International Conference on Machine Learning},
year={2024},
url={https://openreview.net/forum?id=PG5fV50maR}
}

@article{wainwright2008graphical,
  title   = {Graphical Models, Exponential Families, and Variational Inference},
  author  = {Wainwright, Martin J. and Jordan, Michael I.},
  journal = {Foundations and Trends in Machine Learning},
  volume  = {1},
  number  = {1--2},
  pages   = {1--305},
  year    = {2008},
  doi     = {10.1561/2200000001}
}

@inproceedings{yarotsky2026corner,
  title={Corner Gradient Descent},
  author={Yarotsky, Dmitry},
  booktitle={International Conference on Learning Representations},
  volume={2026},
  pages={156292--156327},
  year={2026}
}

@article{killamsetty2021retrieve,
  title={Retrieve: Coreset selection for efficient and robust semi-supervised learning},
  author={Killamsetty, Krishnateja and Zhao, Xujiang and Chen, Feng and Iyer, Rishabh},
  journal={Advances in neural information processing systems},
  volume={34},
  pages={14488--14501},
  year={2021}
}

@inproceedings{
sener2018active,
title={Active Learning for Convolutional Neural Networks: A Core-Set Approach},
author={Ozan Sener and Silvio Savarese},
booktitle={International Conference on Learning Representations},
year={2018},
url={https://openreview.net/forum?id=H1aIuk-RW},
}

@inproceedings{zheng2023coveragecentric,
  title={Coverage-centric Coreset Selection for High Pruning Rates},
  author={Zheng, Haizhong and Liu, Rui and Lai, Fan and Prakash, Atul},
  booktitle={11th International Conference on Learning Representations, ICLR 2023},
  year={2023}
}

@article{maharana2024mathbbd,
  title={D2 pruning: Message passing for balancing diversity and difficulty in data pruning},
  author={Maharana, Adyasha and Yadav, Prateek and Bansal, Mohit},
  journal={arXiv preprint arXiv:2310.07931},
  year={2023}
}

@article{dai2025improving,
  title={Improving influence-based instruction tuning data selection for balanced learning of diverse capabilities},
  author={Dai, Qirun and Zhang, Dylan and Ma, Jiaqi W and Peng, Hao},
  journal={arXiv preprint arXiv:2501.12147},
  year={2025}
}

@inproceedings{
paul2021deep,
title={Deep Learning on a Data Diet: Finding Important Examples Early in Training},
author={Mansheej Paul and Surya Ganguli and Gintare Karolina Dziugaite},
booktitle={Advances in Neural Information Processing Systems},
editor={A. Beygelzimer and Y. Dauphin and P. Liang and J. Wortman Vaughan},
year={2021},
url={https://openreview.net/forum?id=Uj7pF-D-YvT}
}

@inproceedings{
toneva2018an,
title={An Empirical Study of Example Forgetting during Deep Neural Network Learning},
author={Mariya Toneva and Alessandro Sordoni and Remi Tachet des Combes and Adam Trischler and Yoshua Bengio and Geoffrey J. Gordon},
booktitle={International Conference on Learning Representations},
year={2019},
url={https://openreview.net/forum?id=BJlxm30cKm},
}

@inproceedings{koh2017understanding,
  title={Understanding black-box predictions via influence functions},
  author={Koh, Pang Wei and Liang, Percy},
  booktitle={International conference on machine learning},
  pages={1885--1894},
  year={2017},
  organization={PMLR}
}

@inproceedings{achille2019task2vec,
  title={Task2vec: Task embedding for meta-learning},
  author={Achille, Alessandro and Lam, Michael and Tewari, Rahul and Ravichandran, Avinash and Maji, Subhransu and Fowlkes, Charless C and Soatto, Stefano and Perona, Pietro},
  booktitle={Proceedings of the IEEE/CVF international conference on computer vision},
  pages={6430--6439},
  year={2019}
}

@inproceedings{hwang2020data,
  title={Data distribution search to select core-set for machine learning},
  author={Hwang, Myunggwon and Jeong, Yuna and Sung, Wonkyung},
  booktitle={The 9th International conference on smart media and applications},
  pages={172--176},
  year={2020}
}

@article{alvarez2020geometric,
  title={Geometric dataset distances via optimal transport},
  author={Alvarez-Melis, David and Fusi, Nicolo},
  journal={Advances in Neural Information Processing Systems},
  volume={33},
  pages={21428--21439},
  year={2020}
}

@inproceedings{kim-etal-2024-single,
    title = "A Single Linear Layer Yields Task-Adapted Low-Rank Matrices",
    author = "Kim, Hwichan  and
      Sasaki, Shota  and
      Hoshino, Sho  and
      Honda, Ukyo",
    editor = "Calzolari, Nicoletta  and
      Kan, Min-Yen  and
      Hoste, Veronique  and
      Lenci, Alessandro  and
      Sakti, Sakriani  and
      Xue, Nianwen",
    booktitle = "Proceedings of the 2024 Joint International Conference on Computational Linguistics, Language Resources and Evaluation (LREC-COLING 2024)",
    month = may,
    year = "2024",
    address = "Torino, Italia",
    publisher = "ELRA and ICCL",
    url = "https://aclanthology.org/2024.lrec-main.141/",
    pages = "1602--1608"
}

@article{chung2022scaling,
  title={Scaling instruction-finetuned language models},
  author={Chung, Hyung Won and Hou, Le and Longpre, Shayne and Zoph, Barret and Tay, Yi and Fedus, William and Li, Yunxuan and Wang, Xuezhi and Dehghani, Mostafa and Brahma, Siddhartha and others},
  journal={Journal of Machine Learning Research},
  volume={25},
  number={70},
  pages={1--53},
  year={2024}
}

@article{sanh2021multitask,
  title={Multitask prompted training enables zero-shot task generalization},
  author={Sanh, Victor and Webson, Albert and Raffel, Colin and Bach, Stephen H and Sutawika, Lintang and Alyafeai, Zaid and Chaffin, Antoine and Stiegler, Arnaud and Scao, Teven Le and Raja, Arun and others},
  journal={arXiv preprint arXiv:2110.08207},
  year={2021}
}

@article{wei2022chain,
  title={Chain-of-thought prompting elicits reasoning in large language models},
  author={Wei, Jason and Wang, Xuezhi and Schuurmans, Dale and Bosma, Maarten and Xia, Fei and Chi, Ed and Le, Quoc V and Zhou, Denny and others},
  journal={Advances in neural information processing systems},
  volume={35},
  pages={24824--24837},
  year={2022}
}

@article{touvron2023llama,
  title={Llama 2: Open foundation and fine-tuned chat models},
  author={Touvron, Hugo and Martin, Louis and Stone, Kevin and Albert, Peter and Almahairi, Amjad and Babaei, Yasmine and Bashlykov, Nikolay and Batra, Soumya and Bhargava, Prajjwal and Bhosale, Shruti and others},
  journal={arXiv preprint arXiv:2307.09288},
  year={2023}
}

@inproceedings{wang2023super,
  title={Super-naturalinstructions: Generalization via declarative instructions on 1600+ nlp tasks},
  author={Wang, Yizhong and Mishra, Swaroop and Alipoormolabashi, Pegah and Kordi, Yeganeh and Mirzaei, Amirreza and Naik, Atharva and Ashok, Arjun and Dhanasekaran, Arut Selvan and Arunkumar, Anjana and Stap, David and others},
  booktitle={Proceedings of the 2022 conference on empirical methods in natural language processing},
  pages={5085--5109},
  year={2022}
}

@inproceedings{singh2024aya,
  title     = {Aya Dataset: An Open-Access Collection for Multilingual Instruction Tuning},
  author    = {Singh, Shivalika and Vargus, Freddie and D'souza, Daniel and Karlsson, B{\"o}rje F. and Mahendiran, Abinaya and Ko, Wei-Yin and Shandilya, Herumb and Patel, Jay and Mataciunas, Deividas and O'Mahony, Laura and others},
  booktitle = {Proceedings of the 62nd Annual Meeting of the Association for Computational Linguistics (ACL)},
  pages     = {11521--11567},
  year      = {2024}
}

@misc{qwen3.5,
    title  = {{Qwen3.5}: Towards Native Multimodal Agents},
    author = {{Qwen Team}},
    month  = {February},
    year   = {2026},
    url    = {https://qwen.ai/blog?id=qwen3.5}
}

@inproceedings{longpre2023flan,
  title={The flan collection: Designing data and methods for effective instruction tuning},
  author={Longpre, Shayne and Hou, Le and Vu, Tu and Webson, Albert and Chung, Hyung Won and Tay, Yi and Zhou, Denny and Le, Quoc V and Zoph, Barret and Wei, Jason and others},
  booktitle={International conference on machine learning},
  pages={22631--22648},
  year={2023},
  organization={PMLR}
}

@article{lambert2024t,
  title={Tulu 3: Pushing frontiers in open language model post-training},
  author={Lambert, Nathan and Morrison, Jacob and Pyatkin, Valentina and Huang, Shengyi and Ivison, Hamish and Brahman, Faeze and Miranda, Lester James V and Liu, Alisa and Dziri, Nouha and Lyu, Shane and others},
  journal={arXiv preprint arXiv:2411.15124},
  year={2024}
}

@inproceedings{wang2018glue,
  title={GLUE: A multi-task benchmark and analysis platform for natural language understanding},
  author={Wang, Alex and Singh, Amanpreet and Michael, Julian and Hill, Felix and Levy, Omer and Bowman, Samuel},
  booktitle={Proceedings of the 2018 EMNLP workshop BlackboxNLP: Analyzing and interpreting neural networks for NLP},
  pages={353--355},
  year={2018}
}

@article{wang2019superglue,
  title={Superglue: A stickier benchmark for general-purpose language understanding systems},
  author={Wang, Alex and Pruksachatkun, Yada and Nangia, Nikita and Singh, Amanpreet and Michael, Julian and Hill, Felix and Levy, Omer and Bowman, Samuel},
  journal={Advances in neural information processing systems},
  volume={32},
  year={2019}
}

@article{liu2024regmix,
  title={Regmix: Data mixture as regression for language model pre-training},
  author={Liu, Qian and Zheng, Xiaosen and Muennighoff, Niklas and Zeng, Guangtao and Dou, Longxu and Pang, Tianyu and Jiang, Jing and Lin, Min},
  journal={arXiv preprint arXiv:2407.01492},
  year={2024}
}

@article{li2025data,
  title={Data mixing optimization for supervised fine-tuning of large language models},
  author={Li, Yuan and Liu, Zhengzhong and Xing, Eric},
  journal={arXiv preprint arXiv:2508.11953},
  year={2025}
}

@article{ortizjimenez2023taskarithmetic_tangent,
  title = {Task Arithmetic in the Tangent Space: Improved Editing of Pre-Trained Models},
  author = {Ortiz-Jimenez, Guillermo and Favero, Alessandro and Frossard, Pascal},
  journal = {arXiv preprint arXiv:2305.12827},
  year = {2023},
  url = {https://arxiv.org/abs/2305.12827},
  doi = {10.48550/arXiv.2305.12827}
}

@misc{yang2024qwen2technicalreport,
      title={Qwen2 Technical Report}, 
      author={An Yang and Baosong Yang and Binyuan Hui and Bo Zheng and Bowen Yu and Chang Zhou and Chengpeng Li and Chengyuan Li and Dayiheng Liu and Fei Huang and Guanting Dong and Haoran Wei and Huan Lin and Jialong Tang and Jialin Wang and Jian Yang and Jianhong Tu and Jianwei Zhang and Jianxin Ma and Jianxin Yang and Jin Xu and Jingren Zhou and Jinze Bai and Jinzheng He and Junyang Lin and Kai Dang and Keming Lu and Keqin Chen and Kexin Yang and Mei Li and Mingfeng Xue and Na Ni and Pei Zhang and Peng Wang and Ru Peng and Rui Men and Ruize Gao and Runji Lin and Shijie Wang and Shuai Bai and Sinan Tan and Tianhang Zhu and Tianhao Li and Tianyu Liu and Wenbin Ge and Xiaodong Deng and Xiaohuan Zhou and Xingzhang Ren and Xinyu Zhang and Xipin Wei and Xuancheng Ren and Xuejing Liu and Yang Fan and Yang Yao and Yichang Zhang and Yu Wan and Yunfei Chu and Yuqiong Liu and Zeyu Cui and Zhenru Zhang and Zhifang Guo and Zhihao Fan},
      year={2024},
      eprint={2407.10671},
      archivePrefix={arXiv},
      primaryClass={cs.CL},
      url={https://arxiv.org/abs/2407.10671}, 
}
